\documentclass{article} % For LaTeX2e
\usepackage{iclr2027_conference_arxiv,times}

\usepackage{amsmath,amsfonts,bm}

\def\eqref#1{equation~\ref{#1}}
\def\1{\bm{1}}

\DeclareMathAlphabet{\mathsfit}{\encodingdefault}{\sfdefault}{m}{sl}
\SetMathAlphabet{\mathsfit}{bold}{\encodingdefault}{\sfdefault}{bx}{n}

\usepackage{hyperref}
\usepackage{url}

\usepackage{booktabs}       % professional-quality tables
\usepackage{amsfonts}       % blackboard math symbols
\usepackage{nicefrac}       % compact symbols for 1/2, etc.
\usepackage{microtype}      % microtypography
\usepackage{xcolor}         % colors
\usepackage{wrapfig}
\usepackage{graphicx}
\usepackage{algorithm}
\usepackage{algorithmic}
\usepackage{amsmath}
\usepackage{amsfonts}
\usepackage{amssymb}
\usepackage{booktabs} % For professional table lines
\usepackage{multirow} % For multi-row cells
\usepackage{amsmath}  % For \pm and math symbols
\usepackage{graphicx}
\usepackage[table]{xcolor}
\usepackage{colortbl}
\usepackage[none]{hyphenat}
\usepackage[table]{xcolor}
\usepackage{booktabs}
\usepackage{multirow}
\usepackage[table]{xcolor}
\usepackage[table]{xcolor}
\usepackage{multirow, booktabs, graphicx}
\usepackage{enumitem}

\definecolor{headergray}{RGB}{240,240,240}
\definecolor{headergraytwo}{RGB}{248,248,248}
\definecolor{lastcolblue}{RGB}{235,242,255}
\definecolor{summarygreen}{RGB}{235,248,235}

\newtheorem{proposition}{Proposition}

\usepackage[most]{tcolorbox}
\usepackage{inconsolata}
\usepackage{float}

\newtcolorbox{coffeebox}[1]{
  enhanced,
  colback=white,
  colframe=brown!60!black,   % Rich coffee brown border
  arc=5pt,
  boxrule=1.2pt,
  fontupper=\small\ttfamily,
  title={\raggedright #1},
  colbacktitle=brown!60!black, % Coffee bean header
  coltitle=white,
  fonttitle=\small\ttfamily\bfseries,
  left=8pt, right=8pt, top=8pt, bottom=8pt,
  drop shadow={black!30!white} % Adds a subtle professional "lift"
}

\usepackage{xspace}
\newcommand{\ours}{\textsc{CofFeQA}\xspace}

\title{\ours: Counterfactual-Guided Few-Shot \\Distillation for LLM Question Answering}

\author{Kavinda Kehelella \\ University of Maryland, College Park \\ \texttt{kkehelel@umd.edu} \And Sanghamitra Dutta \\ University of Maryland, College Park \\ \texttt{sanghamd@umd.edu}}

\iclrpreprintcopy
\begin{document}

\maketitle

\begin{abstract}
Knowledge distillation aims to train a smaller student model to mimic a larger teacher model for resource-constrained environments. However, the process of distillation is typically data-intensive and computationally expensive, particularly for generative tasks. Few-shot distillation offers an alternative by transferring teacher knowledge using limited training samples. However, we observe that few-shot distillation performance is heavily dependent on the choice of these training samples. In fact, we demonstrate that using more uncertain samples near the teacher's decision boundary can significantly deteriorate distillation performance. Motivated by this observation, we propose \textit{\textbf{Co}unter\textbf{f}actual-guided \textbf{Fe}w-shot Distillation in open-book \textbf{Q}uestion \textbf{A}nswering} (\textbf{\ours}), a novel strategy that aptly uses counterfactuals to provide additional supervision around challenging factual samples. In the generative QA setting, we define a counterfactual as a minimally modified context that induces the teacher to change its response to an alternative answer. We provide geometric insights on how factual-counterfactual pairs can constrain the deviation between the teacher's and student's decision boundaries. We further introduce a new metric, \textit{\textbf{C}ontext \textbf{F}ollowing \textbf{D}ifficulty} (\textbf{CFD}) that further validates counterfactuals prior to distillation based on agreement with the intended alternative answer and dependence on the modified context. \ours uses CFD-validated factual-counterfactual pairs for few-shot distillation. Experiments across five in-domain and three out-of-domain QA benchmarks using T5 and Qwen2.5 families show that \ours outperforms factual-only distillation baselines under the same few-shot budget. For instance, on TriviaQA, \ours improves over the factual-only baseline by $8.77$ average EM points and $9.98$ average F1 points under the T5 setting with 32 few-shot pairs.
\end{abstract}

\section{Introduction}
Large language models (LLMs) have achieved remarkable success across a wide range of natural language processing tasks~\citep{achiam2023gpt, chung2024scaling, gou2021knowledge, raffel2020exploring, fang2026knowledge}. However, their increasing size and computational requirements make deployment in resource-constrained environments particularly challenging~\citep{kaplan2020scaling, brown2020language, strubell2019energy, hoffmann2022training}. This motivates the need for compact models with low training and inference costs that preserve task-relevant capabilities of larger models~\citep{liang2023less, saadi2025taskd, liu2024ddk,dissanayake2024quantifying}. Knowledge distillation~\citep{hinton2015distilling} (see surveys~\citet{gou2021knowledge, sanh2019distilbert, xu2024survey}) seeks to address this challenge by transferring knowledge from a large teacher model to a smaller student model. However, most distillation approaches typically rely on large-scale datasets or extensive supervision from the teacher, thus incurring prohibitive training
costs for already resource-intensive language models~\citep{sanh2019distilbert, sun2019patient, jiao2020tinybert, wang2020minilm,dissanayake2025tabdistill,hamman2025few}. This leads to our primary question: \emph{How can we perform data-efficient distillation in LLMs?}

Towards addressing this question, our work studies few-shot distillation for generative tasks, with a focus on open-book question answering. We find that the performance of few-shot distillation is heavily dependent on the choice of the distillation dataset, consistent with the broader importance
of data selection for efficient LLM training~\citep{fu2026t,zhou2023lima}. In particular, samples near the teacher's decision boundary can significantly deteriorate distillation performance (see Fig.~\ref{fig:boundary-performance} in Section~\ref{app:boundary}), also consistent with prior findings on knowledge transfer around decision boundaries~\citep{chen2025medium}. This motivates us to provide additional supervision around these challenging regions while maintaining a limited few-shot budget. To achieve this, we leverage counterfactual examples. Counterfactual-based data augmentation has improved classification performance~\citep{kaushik2019learning, wu2021polyjuice, ross2021explaining, gardner2020evaluating, mcaleese2024comparative, bhattacharjee2024zero,egea2025vision} by encouraging models to capture causal dependencies and avoid shortcut learning~\citep{guo2023counterfactual, chen2023causal}. However, its applicability to few-shot distillation for generative tasks remains largely unexplored. Unlike classification, generative tasks produce free-form outputs, making counterfactuals non-trivial to define. In generative tasks, we define a counterfactual to be a minimally modified input that changes the expected output while preserving overall semantic consistency, such as modifying supporting context in a question to induce a different answer~\citep{guo2023counterfactual, longpre2021entity, paranjape2022retrieval, koksal2023hallucination, yu2023ifqa}.

In this work, we propose \textit{\textbf{Co}unter\textbf{f}actual-guided \textbf{Fe}w-shot distillation in open-book \textbf{Q}uestion \textbf{A}nswering} (\textbf{\ours}), a novel strategy for few-shot distillation that aptly uses counterfactuals to provide additional supervision around challenging factual samples near the teacher's decision boundary. We demonstrate how selecting high-quality counterfactuals can guide knowledge transfer more effectively using extremely limited few-shot
budget. Our main contributions are as follows:

\begin{figure}[t]
    \centering
    \includegraphics[width=\linewidth]{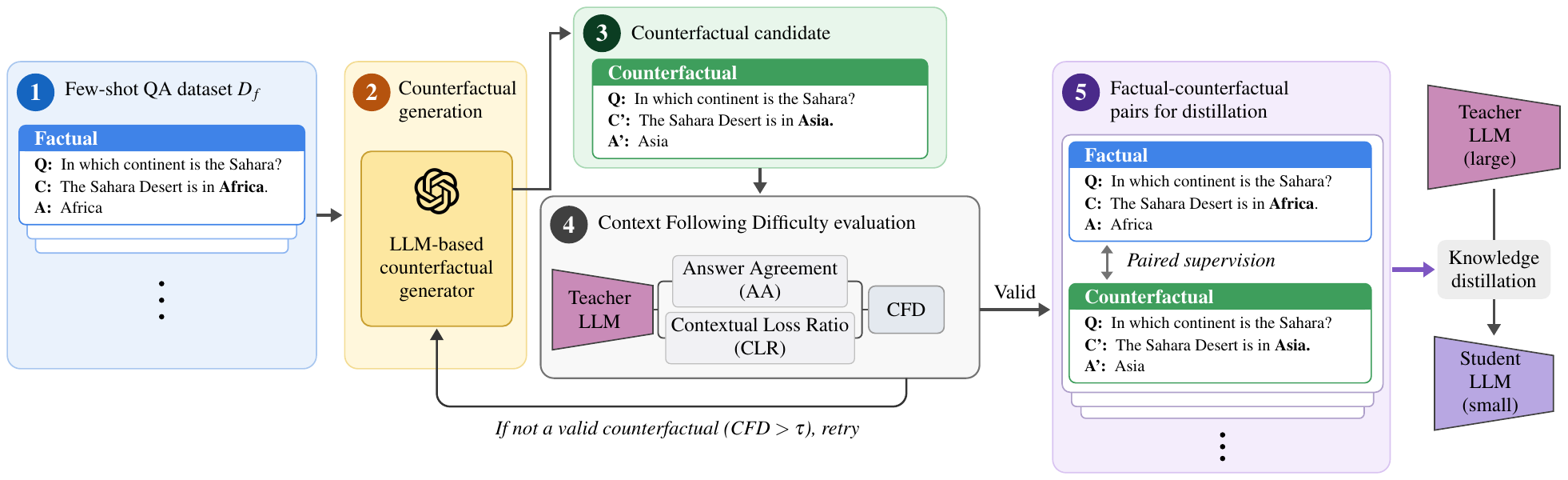}
    \caption{\textbf{Counterfactual-guided few-shot distillation framework.}
    We begin with (1) a few-shot factual QA dataset $\mathcal{D}_f$ and (2) generate a counterfactual candidate for a factual sample. (3) The generated counterfactual is evaluated using the teacher model with (4) our proposed Context Following Difficulty (CFD) metric and accepted if it satisfies the CFD threshold $\tau$. (5) The CFD-validated counterfactuals are then paired with their corresponding factuals and used to distill the student model. }
    \label{fig:framework_diagram}
\end{figure}

     \textbf{Counterfactual-guided few-shot distillation for generative tasks.}
    We show that few-shot distillation becomes particularly challenging for factuals near the teacher's decision boundary, where the teacher has a smaller margin between its preferred answer and competing alternatives. Motivated by this observation, we propose \textbf{\ours}, a novel framework which leverages counterfactuals for few-shot distillation in open-book QA. Rather than relying solely on factual supervision, \ours pairs challenging factuals with their counterfactuals that expose changes in the teacher's response under minimal modifications to the input context, providing additional supervision around the teacher's decision boundary. This reduces dependence on large factual training corpora and improves training efficiency under limited supervision (see Figure~\ref{fig:framework_diagram} for an overview).

\textbf{Geometric analysis of counterfactual-guided knowledge transfer.}
We first introduce a notion of decision boundaries for the teacher and student in the context-embedding space, motivated by the fact that variations in the input context can lead to different answers. Our formulation enables us to analyze distillation in open-book QA from a geometric perspective. Specifically, we show that factual-counterfactual pairs can act as probes for the student to approximate the local geometry of the teacher's decision boundary. To quantify the deviation between the teacher and student, we use the \textit{Hausdorff distance}~\citep{rockafellar1998variational}, which measures the worst-case deviation between their decision boundaries. We show in Proposition~\ref{prop:dev_bound} that this deviation is bounded by the magnitude of the counterfactual perturbations and their coverage of the teacher's decision boundary. Importantly, factuals near the teacher's decision boundary can require smaller perturbations to induce a change in the teacher's response, which can lead to tighter teacher-student boundary alignment. 
    
    \textbf{Context Following Difficulty (CFD) for counterfactual validation.}
    Although we generate counterfactuals by perturbing the original input context to induce a targeted alternative answer, not all generated examples are truly valid with respect to the teacher. The teacher may not produce same target answers using the modified context or may still be able to infer them without actually relying on the context. Such cases weaken the intended causal relationship that the student was intended to learn from the teacher and may encourage shortcut learning. To address this issue, we introduce a new metric \textit{Context Following Difficulty (CFD)} to evaluate generated counterfactuals before distillation. CFD jointly considers whether the teacher produces the intended alternative answer and whether that answer becomes more predictable when conditioned on the modified context. \ours uses CFD-validated counterfactuals, ensuring that the resulting supervision better captures the teacher's context-dependent behavior.
    
    \textbf{Empirical evaluation.}
    We conduct comprehensive experiments across five in-domain and three out-of-domain benchmarks using the T5~\citep{raffel2020exploring} and Qwen2.5~\citep{bai2023qwen} model families. The in-domain benchmarks include NaturalQuestions, SQuAD, NewsQA, TriviaQA, and HotpotQA, while the out-of-domain benchmarks include DROP, DuoRC, and RelationExtraction~\citep{fisch2019mrqa}. We evaluate \ours with standard knowledge distillation (KD), sequence-level knowledge distillation (SeqKD), and layer-wise distillation (LWD) baselines, under the same few-shot training budget. Across these settings, \ours outperforms factual-only baselines. For instance, on TriviaQA, \ours improves over the KD baseline by $8.77$ average EM points and $9.98$ average F1 points under the T5 setting with 32 few-shot pairs.

\subsection{Related Work}

\textbf{Counterfactual Data Augmentation.}
Counterfactual data augmentation has been widely studied for improving robustness in neural networks~\citep{kaushik2020explaining, sachdeva2024catfood, calderon2022docogen,egea2025vision, mishra2021survey}. Early approaches relied on human-generated counterfactuals~\citep{kaushik2019learning, gardner2020evaluating}, while later works automated counterfactual generation through trained generators and editors~\citep{wu2021polyjuice, ross2021explaining, chen2021reinforced, tokpo2024fairflow} and more recently through prompting LLMs~\citep{bhattacharjee2024zero}, alongside efforts to evaluate the quality of generated counterfactuals~\citep{youssef2024llms, mcaleese2024comparative}. For example, VISION~\citep{egea2025vision} uses LLM-generated counterfactuals to train graph neural networks for vulnerability classification. Other works study counterfactual augmentation through active learning~\citep{balashankar2023improving} and fairness-aware pipelines~\citep{tokpo2024fairflow}.  Counterfactual explanations have been studied for model extraction with theoretical guarantees~\citep{dissanayake2024model,dissanayake2026foundations} and for robustness to model changes~\citep{noorani2025counterfactual,hamman2024robust}. Counterfactual applications also extend to generative tasks, including machine translation~\citep{liu2021counterfactual} and dialogue~\citep{ou2022counterfactual}. However, most of these works focus on neural network models. Counterfactual augmentation to fine-tune LLMs has received limited attention.
 CATfOOD~\citep{sachdeva2024catfood} uses LLM-generated counterfactuals to train smaller language models, while \citet{rajagopal2022counterfactual} use counterfactually augmented training data for pretrained summarization models. In contrast, we use factual-counterfactual pairs as structured supervision for few-shot distillation in generative open-book QA, where the teacher produces free-form answers rather than predictions over a predefined labels.

\textbf{Counterfactual Question Answering.}
Counterfactual reasoning has recently gained traction in question answering (QA)~\citep{ross2022tailor, khashabi2020more, huang2025survey}. DisentQA~\citep{neeman2023disentqa} uses counterfactual supervision to separate parametric and contextual knowledge~\citep{xu2024knowledge, wang2023resolving}, while IfQA~\citep{yu2023ifqa} provides a benchmark for QA under hypothetical scenarios that require reasoning beyond factual retrieval. Counterfactual reasoning has also been studied in multi-hop QA~\citep{guo2023counterfactual} and CofCA~\citep{wu2024cofca}, where counterfactual inference is used to mitigate shortcut reasoning. Retrieval-guided approaches further construct counterfactuals through context and answer modifications~\citep{paranjape2022retrieval, dixit2022core}. While these works demonstrate the value of counterfactuals for improving robustness and reasoning in QA, they do not study how they can facilitate knowledge distillation.

\textbf{Distillation in LLMs.}
Knowledge Distillation~\citep{hinton2015distilling} aims to train smaller student models to mimic larger teacher models, with extensions to sequence-level distillation~\citep{xu2024survey, kim2016sequence, gu2024minillm, yang2025survey}, task-specific distillation~\citep{dissanayake2024quantifying,sanh2019distilbert, team2024gemma, wang2025comprehensive}, and layer-wise distillation~\citep{jiao2020tinybert}. Few-shot distillation has been studied for
transferring pretrained transformer knowledge in tabular classification~\citep{dissanayake2025tabdistill} and
multiple-choice question answering~\citep{sutanto2025llm}. In particular, \citet{sutanto2025llm} use an LLM to generate multiple-choice questions and answer options, together with probability scores that are distilled into an encoder-only student model. However, their approach relies on a predefined answer-choice space, making it less directly applicable to generative open-book QA, where the student produces free-form answers. 
While generative and instruction-level distillation often rely on direct supervision or synthetic data~\citep{hsieh2023distilling, wang2023self, gu2024minillm}, only a few recent works have investigated counterfactuals for knowledge transfer. DISCO~\citep{chen2023disco} generates counterfactual examples using an LLM and employs a task-specific teacher to filter the generated examples, which are then used as an augmented dataset to train a smaller student model. Notably, DISCO is different from standard distillation via aligning representations since the teacher primarily serves as a data-filtering mechanism rather than directly transferring its predictive behavior to the student through a distillation loss. Moreover, its empirical evaluation focuses on natural language inference, a classification task with a predefined label space. Another related work \citet{hamman2025few} study counterfactual supervision for few-shot distillation, but also focus on  classification settings. These approaches do not address generative tasks, where the outputs are open-ended. In contrast, we specifically focus on \emph{few-shot} knowledge distillation for generative open-book QA. Our approach is motivated by the difficulty of distilling with factuals near the teacher's decision boundary and uses counterfactuals to provide additional supervision around these challenging regions. We also provide new geometric insights on counterfactual-guided knowledge transfer and a new metric CFD to further check counterfactual validity in open-book QA. 

\section{\ours: Counterfactual-Guided Few-Shot Distillation}
\label{sec:method}

Our strategy is motivated by the observation that factual samples near the
teacher's decision boundary are particularly challenging to distill under
limited supervision. We use margin sampling~\citep{scheffer2001active}
as an operational proxy for boundary proximity, where smaller teacher prediction
margins indicate boundary near samples. Empirically, we observe that
distillation performance degrades as this margin decreases; a detailed analysis
is provided in Section~\ref{app:boundary}. Motivated by this observation, \ours
pairs boundary-near factuals with their counterfactuals that expose
how the teacher's response changes under minimal variations of the input
context. We first define and describe the counterfactual generation process, and then geometrically analyze how factual-counterfactual pairs
facilitate teacher-student boundary alignment. Finally, we introduce Context
Following Difficulty (CFD) to validate generated counterfactuals and present the
overall \ours distillation framework.

\subsection{Counterfactual Definition and Generation}
\label{sec:counterfactual_generation}

We define counterfactual generation for open-book QA as constructing minimally modified training examples that preserve the original question while changing the evidence to support a different answer. Given a triplet $(Q, C, A)$, where $Q$ is the question, $C$ is the context, and $A$ is the answer, the counterfactual generation framework constructs a counterfactual  by modifying the context using an alternative answer. The resulting counterfactual triplet is denoted as $(Q, C', A')$. We generate counterfactuals using GPT-4o-mini~\citep{openai2024gpt4omini} in three stages. First, the answer $A$ is classified into one of five entity types: \textit{person, location, organization, date}, or \textit{numeric}; samples outside these categories are excluded. Second, GPT-4o-mini generates an alternative answer $A'$ of the same entity type that is similar in category  but distinct from $A$. Previously generated alternatives are excluded to encourage diversity across generation attempts. 
Finally, GPT-4o-mini is provided with the original triplet $(Q,C,A)$ and $A'$ to generate the counterfactual context $C'$ by minimally editing $C$. Specifically, occurrences of $A$ are replaced with $A'$, together with any related references that must be changed to keep the passage grammatically consistent. As a result, the modified context $C'$ supports $A'$ as the answer to the same question $Q$, yielding the counterfactual triplet $(Q,C',A')$. The system and user prompts used are shown in Figure~\ref{fig:cf_generation_prompt}. Additional prompt details are provided in Appendix~\ref{app:prompts}.

\begin{figure}[!htbp]
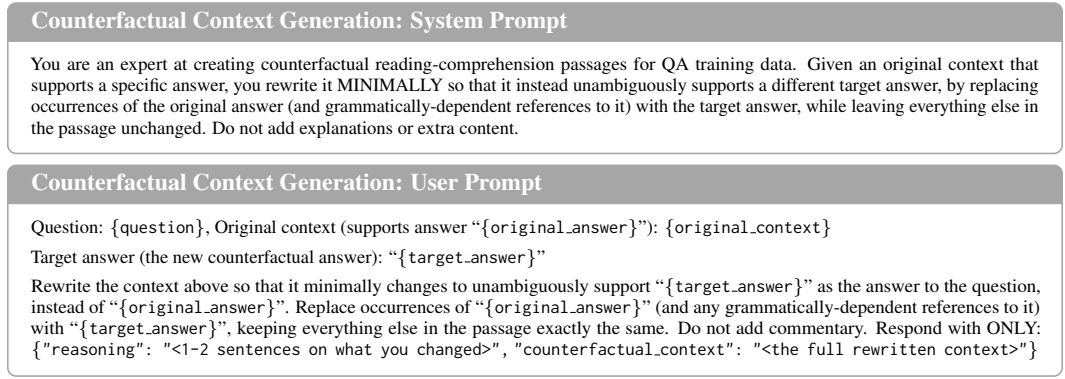

\centering
% \vspace{-1.5mm}
\begin{tcolorbox}[
    title=\textbf{Counterfactual Context Generation: System Prompt},
    colback=white,
    colframe=gray!70,
    colbacktitle=gray!70,
    coltitle=white,
    fonttitle=\bfseries\footnotesize,
    boxrule=0.6pt,
    arc=1mm,
    left=2mm,
    right=2mm,
    top=1mm,
    bottom=1mm,
    boxsep=1mm
]
\scriptsize
You are an expert at creating counterfactual reading-comprehension
passages for QA training data. Given an original context that supports a
specific answer, you rewrite it MINIMALLY so that it instead
unambiguously supports a different target answer, by replacing
occurrences of the original answer (and grammatically-dependent
references to it) with the target answer, while leaving everything else
in the passage unchanged. Do not add explanations or extra content.
\end{tcolorbox}

\vspace{-2.5mm}

\begin{tcolorbox}[
    title=\textbf{Counterfactual Context Generation: User Prompt},
    colback=white,
    colframe=gray!70,
    colbacktitle=gray!70,
    coltitle=white,
    fonttitle=\bfseries\footnotesize,
    boxrule=0.6pt,
    arc=1mm,
    left=2mm,
    right=2mm,
    top=1mm,
    bottom=1mm,
    boxsep=1mm
]
\scriptsize
Question: \texttt{\{question\}}, Original context (supports answer
``\texttt{\{original\_answer\}}''): \texttt{\{original\_context\}}

\smallskip
Target answer (the new counterfactual answer):
``\texttt{\{target\_answer\}}''

\smallskip
Rewrite the context above so that it minimally changes to unambiguously
support ``\texttt{\{target\_answer\}}'' as the answer to the question,
instead of ``\texttt{\{original\_answer\}}''. Replace occurrences of
``\texttt{\{original\_answer\}}'' (and any grammatically-dependent
references to it) with ``\texttt{\{target\_answer\}}'', keeping everything
else in the passage exactly the same. Do not add commentary.
Respond with ONLY:
\texttt{\{"reasoning": "<1-2 sentences on what you changed>",}
\texttt{"counterfactual\_context": "<the full rewritten context>"\}}
\end{tcolorbox}

\caption{\textbf{Counterfactual context generation prompts.}
System and user prompts used to minimally modify the factual context $C$
into a counterfactual context $C'$ that supports the target answer
$A'$. }
\label{fig:cf_generation_prompt}
\end{figure}

\subsection{Geometric analysis of counterfactual-guided knowledge transfer}

We provide a geometric perspective on how factual-counterfactual pairs can facilitate knowledge transfer. We first introduce a notion of a decision boundary for open-book QA. While classification tasks typically consider a \textit{label-flipping boundary}, we define a \textit{context-response boundary} in the context-embedding space. In classification, the boundary separates regions where two classes have equal probabilities. In contrast, in open-book QA, it represents the transition region in context embedding space where the model shifts from generating one answer to another answer.  
In open-book QA, the context space may contain regions corresponding to the factual answer $A$, the intended counterfactual answer $A'$, and other answer ($A''$)  regions, different from both $A$ and $A'$ corresponding to out-of-context or unrelated outputs. Our analysis focuses specifically on the boundary between the $A$ and $A'$ response regions, as this is the boundary relevant to factual-counterfactual supervision. Our CFD-based 
validation helps identify and reject counterfactual candidates that induce responses corresponding to the other answer ($A''$) regions before distillation.
 We therefore restrict the geometric analysis to the boundary between factual answer space ($A$) and intended counterfactual answer ($A'$) space.
We use the Hausdorff distance~\citep{rockafellar1998variational} to quantify the worst-case deviation between
the teacher context-response boundary and the student context-response boundary and derive an upper bound for this deviation in Proposition~\ref{prop:dev_bound} (The proof can be found in Appendix~\ref{app:proof}).

\begin{wrapfigure}[15]{r}{0.225\textwidth}
    \centering
    \includegraphics[width=0.225\textwidth]{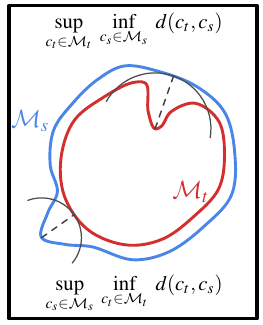}
    \caption{Visualizing the Hausdorff distance between $\mathcal{M}_t$ and $\mathcal{M}_s$.}
    \label{fig:housdorff}   

\end{wrapfigure}

\textbf{Definition of Hausdorff Distance \citep{rockafellar1998variational}.} 
 Let $(\mathcal{M},d)$ be a metric space. For each pair of non-empty subsets $\mathcal{M}_{t}\subset\mathcal{M}$ and $\mathcal{M}_{s}\subset\mathcal{M}$, the Hausdorff distance between $\mathcal{M}_{t}$ and $\mathcal{M}_{s}$ is defined as:

$$    d_{H}(\mathcal{M}_{t},\mathcal{M}_{s}) := \max \left\{ \sup_{c_{t} \in \mathcal{M}_{t}} \inf_{c_{s} \in \mathcal{M}_{s}} d(c_{t},c_{s}), \sup_{c_{s} \in \mathcal{M}_{s}} \inf_{c_{t} \in \mathcal{M}_{t}} d(c_{t},c_{s}) \right\}. $$

This measures the maximum of the minimum distances between two sets, capturing the worst-case geometric deviation. (see Figure~\ref{fig:housdorff}).

\begin{proposition}
\label{prop:dev_bound}
Let $\mathcal{C}$ be the space of context embeddings, $c\in\mathcal{C}$ arbitrary input context, $Q\in\mathcal{Q}$ input question, $A\in\mathcal{A}$ factual answer, and $A^{\prime}\in\mathcal{A^{\prime}}$ counterfactual answer. Consider a teacher and a student with response functions $G_{t}(Q,c)=\arg\max_{A} P_{t}(A|Q,c)$ and $G_{s}(Q,c)=\arg\max_{A} P_{s}(A|Q,c)$. The respective context-flipping boundaries can be defined as the sets:
\[
\mathcal{M}_{t}=\{c \in \mathcal{C} \mid P_{t}(A|Q,c)=P_{t}(A'|Q,c)\}, \quad
\mathcal{M}_{s}=\{c \in \mathcal{C} \mid P_{s}(A|Q,c)=P_{s}(A'|Q,c)\}.
\]

Suppose there exists a factual-counterfactual paired dataset $\mathcal{D}_{f+cf}=\{(c_i,c_i')\}_{i=1}^{k}$ for the above $\mathcal{M}_{t}$ that satisfies three conditions: $(1)$ Minimal context perturbation in context embedding space is bounded by $\alpha$ (i.e., $||c_{i}-c_{i}'||\le\alpha$), $(2)$ Exact answer generation: $G_{t}(Q,c)=G_{s}(Q,c)$ where $c = \{c_i,c'_i\}$, $(3)$ $\epsilon$-spread boundary coverage: Counterfactuals are well-distributed across the context-response boundaries with spread $\epsilon$, i.e., if $t \in \mathcal{M}_{t}$, then $\exists t_i^{*} \; \text{such that} \; \|t - t_i^{*}\| \le \epsilon$. This applies for $\mathcal{M}_{s}$ as well. i.e., if $s \in \mathcal{M}_{s}$, then $\exists s_i^{*} \; \text{such that} \; \|s - s_i^{*}\| \le \epsilon$.

Then, the Hausdorff distance between the teacher boundary and student boundary satisfies: $
d_{H}(\mathcal{M}_{t},\mathcal{M}_{s}) \le \alpha + \epsilon.$ 
\end{proposition}

\textbf{Intuition behind proposition and assumptions.}
Proposition~\ref{prop:dev_bound} highlights two properties of factual-counterfactual supervision that govern teacher-student boundary alignment: the magnitude of the counterfactual perturbations ($\alpha$) and the coverage of the context-response boundary ($\epsilon$). First, each factual-counterfactual pair provides supervision across a local change in the teacher's response. Importantly, when the factual sample is already close to the teacher's decision boundary, a smaller contextual perturbation is sufficient to cross the boundary and induce the alternative response $A'$. Therefore, starting from boundary-near factual samples leads to a smaller upper bound on the perturbation $\alpha$, tightening the bound. Although those samples are difficult under factual-only supervision, their proximity to the boundary makes them useful starting points for constructing local counterfactual supervision. Second, the boundary coverage condition captures how well the induced crossing points cover both the context-response boundaries. Better coverage corresponds to a smaller $\epsilon$, tightening the bound. Thus, we provide
a geometric characterization of how local counterfactual perturbations and boundary coverage relate to teacher-student boundary deviation.

The analysis above also assumes that the student model agrees with the teacher model on the paired factual and counterfactual samples (Assumption (2) Exact answer generation). This idealized condition captures the objective of distillation, where the student is trained to reproduce the teacher's behavior on the available supervision. More importantly, the analysis also relies on the generated counterfactual context $C'$ to induce the intended alternative answer $A'$. In practice, this may not be guaranteed due to inherently stochastic and unpredictable generative behavior of LLMs~\cite{holtzman2019curious}. The teacher may fail to produce $A'$, or conditioning on $C'$ may provide only limited additional predictive benefit
for $A'$. For realization of this objective in our implementation, we therefore introduce a new metric \textit{Context Following Difficulty (CFD)} in the following section to validate generated counterfactuals before they are used for knowledge distillation.

\begin{figure*}[ht]
    \centering
    \includegraphics[width=0.99\textwidth]{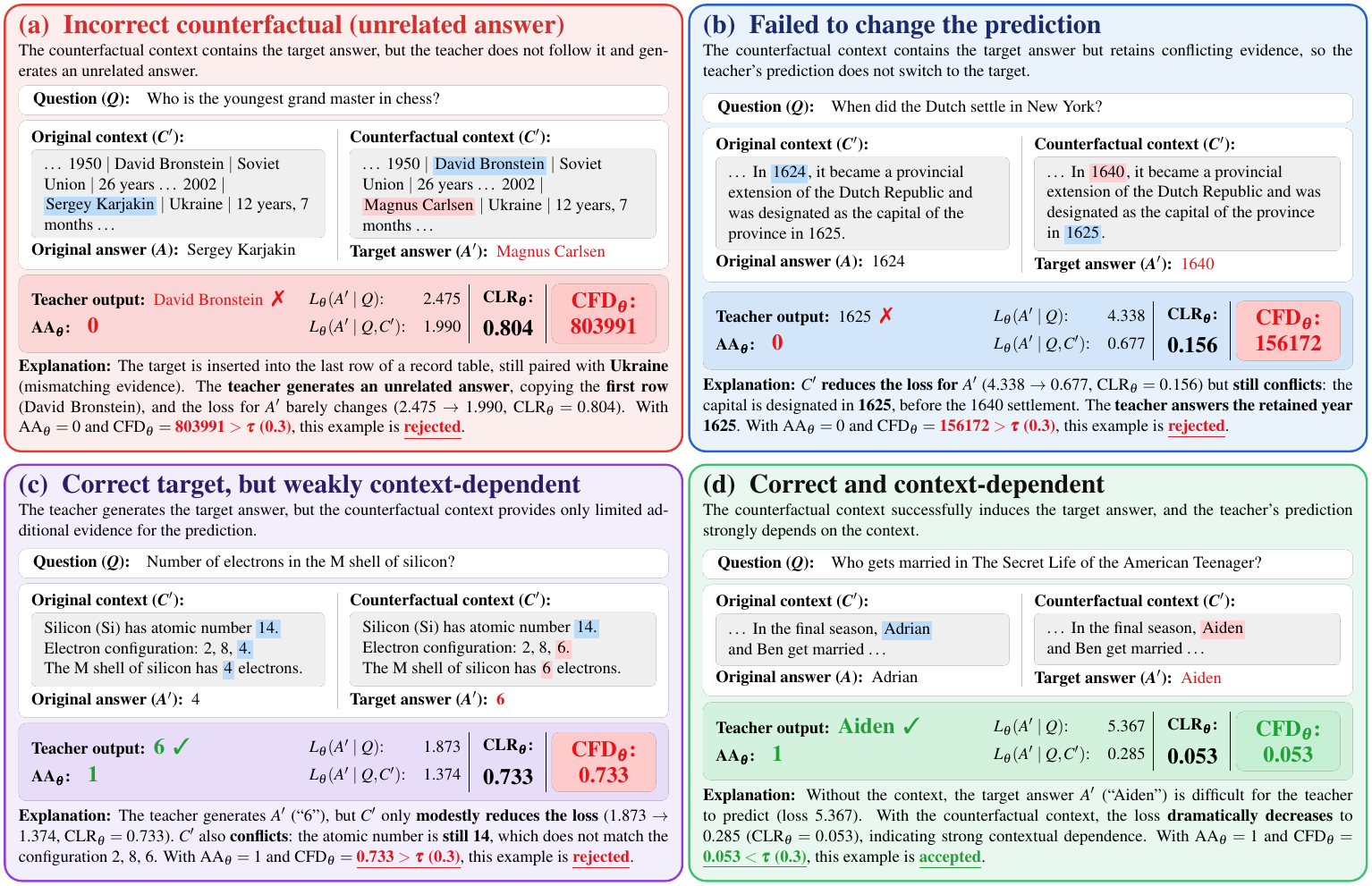}
    \caption{\textbf{Illustrative examples of CFD-based counterfactual evaluation from the NQ dataset.}
    The four cases demonstrate how Answer Agreement (AA) and Contextual Loss Ratio (CLR) complement each other in evaluating generated counterfactual candidates.}
    \label{fig:cfd_examples}
\end{figure*}

\subsection{Context Following Difficulty (CFD) for  counterfactual validation}
\label{sec:cfd}

Generated counterfactual samples should be evaluated using the teacher to determine whether they induce the intended alternative answer $A'$ rather than another response. Since exact
matching can be unreliable for generative tasks~\citep{holtzman2019curious}, we use token-level F1~\citep{rajpurkar2016squad} to measure agreement between the teacher output and $A'$. However, answer agreement alone is insufficient, as generating the intended answer does not necessarily indicate that the $C'$ provides strong evidence for that prediction. A high-quality counterfactual should therefore not only induce the intended answer $A'$, but also make $A'$ substantially more predictable when conditioned on $C'$ than from the  $Q$ alone.
To capture these two properties, we introduce a new metric \textit{Context Following Difficulty (CFD)}, which combines \textit{Answer Agreement (AA)} with the \textit{Contextual Loss Ratio (CLR)}. First, $\mathrm{AA}_{\theta}$ measures whether the teacher generates the intended alternative answer $A'$ under the counterfactual context $C'$:
\begin{equation}
    \mathrm{AA}_{\theta}(Q,C',A')
    =
    \mathrm{F1}\big(\hat{A}_{\theta}(Q,C'), A'\big),
\end{equation}
where $\hat{A}_{\theta}(Q,C')$ is the answer generated by the teacher given $Q$ and $C'$. A higher $\mathrm{AA}_{\theta}$ indicates stronger agreement with the intended answer $A'$.
However, high answer agreement alone does not establish that the counterfactual context plays a role in predicting $A'$. Inspired by the Instruction Following Difficulty (IFD) metric~\citep{li2024quantity}, we define the \textit{Contextual Loss Ratio (CLR)} as:
\begin{equation}
    \mathrm{CLR}_{\theta}(Q,C',A')
    =
    \frac{L_{\theta}(A' \mid Q,C')}
         {L_{\theta}(A' \mid Q)}.
\end{equation}
A low $\mathrm{CLR}_{\theta}$ indicates that conditioning on $C'$ substantially reduces the teacher's loss for $A'$ relative to conditioning on $Q$ alone, i.e., $C'$ plays a strong role in predicting the intended alternative answer. Conversely, a higher $\mathrm{CLR}_{\theta}$ indicates that providing $C'$ yields only a limited reduction in the loss of $A'$, suggesting weaker dependence of the target prediction on $C'$. Thus, $\mathrm{CLR}_{\theta}$ measures how strongly $C'$ contributes to the prediction of the teacher model $A'$.

Combining these two measures, we finally define our \textit{Context Following Difficulty (CFD)} score as:
\begin{equation}
    \mathrm{CFD}_{\theta}(Q,C',A')
    =
    \frac{\mathrm{CLR}_{\theta}(Q,C',A')}
    {\mathrm{AA}_{\theta}(Q,C',A') {+} \delta},
\end{equation}
\begin{wrapfigure}[24]{r}{0.49\textwidth}
\begin{minipage}{0.49\textwidth}
\vspace{-5pt}
 \begin{algorithm}[H]
            \caption{\ours}
            \label{alg:cid}
            \begin{algorithmic}[1]
                \REQUIRE $\mathcal{D}$, teacher $\theta_t$, student $\theta_s$, epochs $E$, learning rate $\eta$, weight $\alpha$, CFD threshold $\tau$, few-shot budget $k$,  $\mathcal{D}_{cf} {=} \phi, \mathcal{D}_{f} {=}\phi$
                \WHILE{$|\mathcal{D}_{cf}| < k/2$}
                    \STATE Pick $(Q, C, A) \in \mathcal{D}$
                        \STATE $(Q, C', A') \leftarrow \text{GenCF}(Q, C, A)$
                        \IF{$ \text{CFD}_{\theta_t}(Q, C', A') \le \tau$}
                            \STATE $\mathcal{D}_{cf} \leftarrow \mathcal{D}_{cf} \cup \{(Q, C', A')\}$
                            \STATE $\mathcal{D}_{f} \leftarrow \mathcal{D}_{f} \cup \{(Q, C, A)\}$
                        \ENDIF
                \ENDWHILE
                    \FOR{$e = 1$ \TO $E$}
                        \FORALL{$(Q, C, A) \in \mathcal{D}_{f} \cup \mathcal{D}_{cf}$}
                            \STATE $\mathcal{L}_{\text{CE}} \leftarrow \text{CE}(\theta_s(Q, C), A)$
                            \STATE $\mathcal{L}_{\text{KD}} \leftarrow \text{KL}(\theta_t(Q, C){\parallel} \theta_s(Q, C))$
                            \STATE $\mathcal{L} \leftarrow \alpha \mathcal{L}_{\text{CE}} + (1 - \alpha) \mathcal{L}_{\text{KD}}$
                            \STATE Update $\theta_s \leftarrow \theta_s - \eta \nabla_{\theta_s} \mathcal{L}$
                        \ENDFOR
                    \ENDFOR
                    \RETURN distilled student $\theta_s$
\end{algorithmic}
\end{algorithm}
\end{minipage}
\end{wrapfigure}

where $\delta > 0$ is a small constant for numerical stability. We prefer low $\mathrm{CLR}_{\theta}$ and high $\mathrm{AA}_{\theta}$. A lower $\mathrm{CFD}_{\theta}$ score favors counterfactuals for which the teacher both generates an answer close to the intended target $A'$ and relies more strongly on the counterfactual context when predicting $A'$. Figure~\ref{fig:cfd_examples} illustrates the complementary roles of $\mathrm{AA}_{\theta}$ and $\mathrm{CLR}_{\theta}$ in four representative cases. Cases (a) and (b) show that a counterfactual may fail to induce the intended answer even when the modified context contains or supports it. Case (c) has high $\mathrm{AA}_{\theta}$ but high $\mathrm{CLR}_{\theta}$, indicating limited additional predictive benefit from $C'$. In contrast, case (c) represents a valid counterfactual with a low $\mathrm{CFD}_{\theta}$. Together, these examples show why both $\mathrm{AA}_{\theta}$ and $\mathrm{CLR}_{\theta}$ are necessary for validation. The two components of $\mathrm{CFD}_{\theta}$ therefore capture complementary properties of a valid and informative counterfactual. $\mathrm{AA}_{\theta}$ verifies the intended answer $A'$, while $\mathrm{CLR}_{\theta}$ measures how strongly the modified context contributes to predicting $A'$. 
Our algorithm is provided in Algorithm~\ref{alg:cid} (also in Figure~\ref{fig:framework_diagram}).

\section{Experiments}
\label{others}

\textbf{Datasets.} 
We evaluate \ours on eight open-domain question answering benchmarks. The five in-domain benchmarks are Natural Questions (NQ)~\citep{kwiatkowski2019natural}, SQuAD~\citep{rajpurkar2016squad}, NewsQA~\citep{trischler2017newsqa}, TriviaQA~\citep{joshi2017triviaqa}, and HotpotQA~\citep{yang2018hotpotqa}. The three out-of-domain benchmarks are DROP~\citep{dua2019drop}, DuoRC~\citep{saha2018duorc}, and RelationExtraction~\citep{levy2017zero}. Additional details about the datasets used in our experimental evaluations are provided in Appendix~\ref{app:benchmarks}

\textbf{Models.} We evaluate our method using two model families with different architectures: the encoder-decoder model T5~\citep{raffel2020exploring} and the decoder-only model Qwen2.5~\citep{bai2023qwen}. In the T5 setting, T5-Large (770M parameters) serves as the teacher, while T5-Base (220M parameters) serves as the student. In the Qwen2.5 setting, Qwen2.5-7B-IT serves as the teacher, while Qwen2.5-1.5B-IT serves as the student. Full-training details are in Appendix~\ref{app:finetuning}.

\begin{table*}[t]
\caption{In-domain F1 score comparison across $k$-shots for T5 and Qwen2.5 models using baselines. Mean F1 scores are reported with standard deviations as subscripts ($Mean_{std}$). Bold values indicate the better-performing method (\ours vs. baseline). Blue shading highlights aggregate results.}
\label{tab:f1_t5_qwen_combined_tiny_subscript}
\centering
\fontsize{7pt}{9pt}\selectfont
\setlength{\tabcolsep}{1.2pt}
\renewcommand{\arraystretch}{0.95}

\definecolor{tableblue}{HTML}{E6F0FF}

\resizebox{\textwidth}{!}{
\begin{tabular}{llcccccccccccccc}
\toprule
 & & \multicolumn{6}{c}{\textbf{T5}} & \multicolumn{6}{c}{\textbf{Qwen2.5}} \\
\cmidrule(lr){3-8} \cmidrule(lr){9-14}
Dataset & Method & 16 & 32 & 64 & 128 & 256 & \cellcolor{tableblue}Avg & 16 & 32 & 64 & 128 & 256 & \cellcolor{tableblue}Avg \\
\midrule

\multirow{6}{*}{NQ} 
& KD & 61.83$_{\scalebox{0.6}{0.49}}$ & 61.44$_{\scalebox{0.6}{2.08}}$ & 60.34$_{\scalebox{0.6}{1.91}}$ & 56.91$_{\scalebox{0.6}{3.20}}$ & 58.07$_{\scalebox{0.6}{1.33}}$ & \cellcolor{tableblue}59.72$_{\scalebox{0.6}{2.15}}$ & \textbf{66.91}$_{\scalebox{0.6}{0.44}}$ & 66.01$_{\scalebox{0.6}{0.59}}$ & 65.71$_{\scalebox{0.6}{0.39}}$ & 64.68$_{\scalebox{0.6}{0.61}}$ & 65.66$_{\scalebox{0.6}{0.69}}$ & \cellcolor{tableblue}65.79$_{\scalebox{0.6}{0.80}}$ \\
& + \ours & \textbf{62.16}$_{\scalebox{0.6}{0.27}}$ & \textbf{62.09}$_{\scalebox{0.6}{1.30}}$ & \textbf{60.85}$_{\scalebox{0.6}{1.37}}$ & \textbf{59.08}$_{\scalebox{0.6}{2.01}}$ & \textbf{58.63}$_{\scalebox{0.6}{2.30}}$ & \cellcolor{tableblue}\textbf{60.56}$_{\scalebox{0.6}{1.65}}$ & 66.27$_{\scalebox{0.6}{0.27}}$ & \textbf{68.10}$_{\scalebox{0.6}{0.20}}$ & \textbf{67.23}$_{\scalebox{0.6}{0.32}}$ & \textbf{67.25}$_{\scalebox{0.6}{0.73}}$ & \textbf{67.19}$_{\scalebox{0.6}{0.99}}$ & \cellcolor{tableblue}\textbf{67.21}$_{\scalebox{0.6}{0.65}}$ \\
\cmidrule(lr){2-14}
& LWD & 61.81$_{\scalebox{0.6}{0.56}}$ & 61.23$_{\scalebox{0.6}{2.49}}$ & \textbf{60.37}$_{\scalebox{0.6}{1.56}}$ & 58.16$_{\scalebox{0.6}{1.81}}$ & 58.72$_{\scalebox{0.6}{2.29}}$ & \cellcolor{tableblue}60.06$_{\scalebox{0.6}{1.57}}$ & \textbf{65.62}$_{\scalebox{0.6}{5.03}}$ & 66.32$_{\scalebox{0.6}{0.94}}$ & 65.90$_{\scalebox{0.6}{1.18}}$ & 65.61$_{\scalebox{0.6}{0.85}}$ & 66.62$_{\scalebox{0.6}{0.86}}$ & \cellcolor{tableblue}66.01$_{\scalebox{0.6}{0.45}}$ \\
& + \ours & \textbf{62.12}$_{\scalebox{0.6}{0.22}}$ & \textbf{61.46}$_{\scalebox{0.6}{2.57}}$ & 60.04$_{\scalebox{0.6}{2.09}}$ & \textbf{60.03}$_{\scalebox{0.6}{1.54}}$ & \textbf{58.93}$_{\scalebox{0.6}{2.71}}$ & \cellcolor{tableblue}\textbf{60.52}$_{\scalebox{0.6}{1.27}}$ & 63.98$_{\scalebox{0.6}{4.96}}$ & \textbf{68.16}$_{\scalebox{0.6}{0.35}}$ & \textbf{67.78}$_{\scalebox{0.6}{0.64}}$ & \textbf{68.07}$_{\scalebox{0.6}{0.56}}$ & \textbf{67.96}$_{\scalebox{0.6}{0.77}}$ & \cellcolor{tableblue}\textbf{67.19}$_{\scalebox{0.6}{1.80}}$ \\
\cmidrule(lr){2-14}
& SeqKD & 60.68$_{\scalebox{0.6}{1.24}}$ & 59.19$_{\scalebox{0.6}{1.07}}$ & 60.80$_{\scalebox{0.6}{1.12}}$ & 59.78$_{\scalebox{0.6}{1.57}}$ & \textbf{62.09}$_{\scalebox{0.6}{1.54}}$ & \cellcolor{tableblue}60.51$_{\scalebox{0.6}{1.11}}$ & 65.89$_{\scalebox{0.6}{0.34}}$ & \textbf{67.55}$_{\scalebox{0.6}{0.55}}$ & 66.84$_{\scalebox{0.6}{0.31}}$ & 67.33$_{\scalebox{0.6}{0.35}}$ & 67.73$_{\scalebox{0.6}{0.76}}$ & \cellcolor{tableblue}67.07$_{\scalebox{0.6}{0.74}}$ \\
& + \ours & \textbf{61.51}$_{\scalebox{0.6}{0.69}}$ & \textbf{61.88}$_{\scalebox{0.6}{0.64}}$ & \textbf{61.52}$_{\scalebox{0.6}{0.92}}$ & \textbf{61.87}$_{\scalebox{0.6}{0.81}}$ & 61.36$_{\scalebox{0.6}{1.41}}$ & \cellcolor{tableblue}\textbf{61.62}$_{\scalebox{0.6}{0.23}}$ & \textbf{66.56}$_{\scalebox{0.6}{0.42}}$ & 66.92$_{\scalebox{0.6}{0.74}}$ & \textbf{68.48}$_{\scalebox{0.6}{0.89}}$ & \textbf{68.04}$_{\scalebox{0.6}{0.58}}$ & \textbf{68.22}$_{\scalebox{0.6}{0.39}}$ & \cellcolor{tableblue}\textbf{67.64}$_{\scalebox{0.6}{0.85}}$ \\
\midrule

\multirow{6}{*}{SQuAD} 
& KD & 86.45$_{\scalebox{0.6}{0.54}}$ & 84.18$_{\scalebox{0.6}{1.39}}$ & 83.63$_{\scalebox{0.6}{1.71}}$ & 84.05$_{\scalebox{0.6}{1.14}}$ & 83.10$_{\scalebox{0.6}{1.19}}$ & \cellcolor{tableblue}84.28$_{\scalebox{0.6}{1.28}}$ & \textbf{77.75}$_{\scalebox{0.6}{0.13}}$ & 76.78$_{\scalebox{0.6}{0.24}}$ & 76.40$_{\scalebox{0.6}{0.47}}$ & 75.87$_{\scalebox{0.6}{0.41}}$ & \textbf{76.45}$_{\scalebox{0.6}{0.39}}$ & \cellcolor{tableblue}76.65$_{\scalebox{0.6}{0.70}}$ \\
& + \ours & \textbf{86.56}$_{\scalebox{0.6}{0.16}}$ & \textbf{86.13}$_{\scalebox{0.6}{0.72}}$ & \textbf{85.36}$_{\scalebox{0.6}{1.03}}$ & \textbf{85.46}$_{\scalebox{0.6}{0.48}}$ & \textbf{84.23}$_{\scalebox{0.6}{1.58}}$ & \cellcolor{tableblue}\textbf{85.55}$_{\scalebox{0.6}{0.89}}$ & 76.95$_{\scalebox{0.6}{0.46}}$ & \textbf{77.28}$_{\scalebox{0.6}{0.59}}$ & \textbf{77.42}$_{\scalebox{0.6}{0.42}}$ & \textbf{76.64}$_{\scalebox{0.6}{0.85}}$ & 75.95$_{\scalebox{0.6}{0.34}}$ & \cellcolor{tableblue}\textbf{76.85}$_{\scalebox{0.6}{0.59}}$ \\
\cmidrule(lr){2-14}
& LWD & 86.47$_{\scalebox{0.6}{0.50}}$ & 84.13$_{\scalebox{0.6}{1.13}}$ & 83.86$_{\scalebox{0.6}{1.78}}$ & 83.45$_{\scalebox{0.6}{0.83}}$ & 82.92$_{\scalebox{0.6}{1.18}}$ & \cellcolor{tableblue}84.17$_{\scalebox{0.6}{1.37}}$ & \textbf{76.70}$_{\scalebox{0.6}{0.65}}$ & 76.96$_{\scalebox{0.6}{0.30}}$ & 76.71$_{\scalebox{0.6}{0.41}}$ & 76.15$_{\scalebox{0.6}{1.00}}$ & 77.20$_{\scalebox{0.6}{1.07}}$ & \cellcolor{tableblue}76.74$_{\scalebox{0.6}{0.39}}$ \\
& + \ours & \textbf{86.56}$_{\scalebox{0.6}{0.23}}$ & \textbf{86.09}$_{\scalebox{0.6}{0.78}}$ & \textbf{84.85}$_{\scalebox{0.6}{0.81}}$ & \textbf{85.57}$_{\scalebox{0.6}{0.49}}$ & \textbf{84.53}$_{\scalebox{0.6}{1.65}}$ & \cellcolor{tableblue}\textbf{85.52}$_{\scalebox{0.6}{0.84}}$ & 76.36$_{\scalebox{0.6}{0.06}}$ & \textbf{77.19}$_{\scalebox{0.6}{0.24}}$ & \textbf{77.75}$_{\scalebox{0.6}{0.41}}$ & \textbf{77.22}$_{\scalebox{0.6}{0.33}}$ & \textbf{77.82}$_{\scalebox{0.6}{1.50}}$ & \cellcolor{tableblue}\textbf{77.27}$_{\scalebox{0.6}{0.58}}$ \\
\cmidrule(lr){2-14}
& SeqKD & 85.43$_{\scalebox{0.6}{0.69}}$ & \textbf{86.25}$_{\scalebox{0.6}{0.67}}$ & 85.57$_{\scalebox{0.6}{0.28}}$ & 85.45$_{\scalebox{0.6}{0.32}}$ & 84.96$_{\scalebox{0.6}{0.52}}$ & \cellcolor{tableblue}85.53$_{\scalebox{0.6}{0.46}}$ & \textbf{78.22}$_{\scalebox{0.6}{0.46}}$ & 77.79$_{\scalebox{0.6}{0.17}}$ & \textbf{77.90}$_{\scalebox{0.6}{0.51}}$ & 77.35$_{\scalebox{0.6}{0.49}}$ & 78.26$_{\scalebox{0.6}{0.96}}$ & \cellcolor{tableblue}77.90$_{\scalebox{0.6}{0.37}}$ \\
& + \ours & \textbf{86.41}$_{\scalebox{0.6}{0.45}}$ & 85.47$_{\scalebox{0.6}{0.51}}$ & \textbf{85.73}$_{\scalebox{0.6}{0.73}}$ & \textbf{85.65}$_{\scalebox{0.6}{0.36}}$ & \textbf{85.02}$_{\scalebox{0.6}{0.37}}$ & \cellcolor{tableblue}\textbf{85.66}$_{\scalebox{0.6}{0.50}}$ & 77.25$_{\scalebox{0.6}{0.37}}$ & \textbf{77.89}$_{\scalebox{0.6}{0.33}}$ & 77.80$_{\scalebox{0.6}{0.65}}$ & \textbf{77.85}$_{\scalebox{0.6}{0.53}}$ & \textbf{78.63}$_{\scalebox{0.6}{0.67}}$ & \cellcolor{tableblue}\textbf{77.92}$_{\scalebox{0.6}{0.49}}$ \\
\midrule

\multirow{6}{*}{NewsQA} 
& KD & 72.72$_{\scalebox{0.6}{2.33}}$ & 73.43$_{\scalebox{0.6}{1.92}}$ & 70.56$_{\scalebox{0.6}{2.16}}$ & 69.95$_{\scalebox{0.6}{2.11}}$ & 65.35$_{\scalebox{0.6}{2.07}}$ & \cellcolor{tableblue}70.40$_{\scalebox{0.6}{3.17}}$ & 57.32$_{\scalebox{0.6}{0.77}}$ & 54.14$_{\scalebox{0.6}{1.48}}$ & 54.32$_{\scalebox{0.6}{1.11}}$ & 55.10$_{\scalebox{0.6}{1.69}}$ & 53.64$_{\scalebox{0.6}{1.71}}$ & \cellcolor{tableblue}54.90$_{\scalebox{0.6}{1.45}}$ \\
& + \ours & \textbf{73.66}$_{\scalebox{0.6}{3.02}}$ & \textbf{74.26}$_{\scalebox{0.6}{1.18}}$ & \textbf{75.43}$_{\scalebox{0.6}{1.71}}$ & \textbf{74.19}$_{\scalebox{0.6}{1.38}}$ & \textbf{72.85}$_{\scalebox{0.6}{2.24}}$ & \cellcolor{tableblue}\textbf{74.08}$_{\scalebox{0.6}{0.94}}$ & \textbf{60.77}$_{\scalebox{0.6}{0.73}}$ & \textbf{57.95}$_{\scalebox{0.6}{0.82}}$ & \textbf{56.88}$_{\scalebox{0.6}{0.60}}$ & \textbf{57.48}$_{\scalebox{0.6}{1.17}}$ & \textbf{59.41}$_{\scalebox{0.6}{2.54}}$ & \cellcolor{tableblue}\textbf{58.50}$_{\scalebox{0.6}{1.58}}$ \\
\cmidrule(lr){2-14}
& LWD & 72.67$_{\scalebox{0.6}{2.30}}$ & 73.00$_{\scalebox{0.6}{2.05}}$ & 70.42$_{\scalebox{0.6}{2.59}}$ & 69.29$_{\scalebox{0.6}{2.49}}$ & 65.25$_{\scalebox{0.6}{0.79}}$ & \cellcolor{tableblue}70.12$_{\scalebox{0.6}{3.14}}$ & 58.76$_{\scalebox{0.6}{1.01}}$ & \textbf{58.62}$_{\scalebox{0.6}{0.68}}$ & 56.90$_{\scalebox{0.6}{0.93}}$ & 57.51$_{\scalebox{0.6}{1.55}}$ & 55.72$_{\scalebox{0.6}{2.27}}$ & \cellcolor{tableblue}57.50$_{\scalebox{0.6}{1.26}}$ \\
& + \ours & \textbf{73.43}$_{\scalebox{0.6}{3.03}}$ & \textbf{74.98}$_{\scalebox{0.6}{0.66}}$ & \textbf{74.85}$_{\scalebox{0.6}{1.56}}$ & \textbf{73.77}$_{\scalebox{0.6}{1.18}}$ & \textbf{73.06}$_{\scalebox{0.6}{2.36}}$ & \cellcolor{tableblue}\textbf{74.02}$_{\scalebox{0.6}{0.86}}$ & \textbf{60.49}$_{\scalebox{0.6}{0.88}}$ & 58.18$_{\scalebox{0.6}{1.05}}$ & \textbf{58.05}$_{\scalebox{0.6}{1.13}}$ & \textbf{59.89}$_{\scalebox{0.6}{0.48}}$ & \textbf{61.23}$_{\scalebox{0.6}{4.39}}$ & \cellcolor{tableblue}\textbf{59.57}$_{\scalebox{0.6}{1.41}}$ \\
\cmidrule(lr){2-14}
& SeqKD & 66.77$_{\scalebox{0.6}{1.63}}$ & 70.09$_{\scalebox{0.6}{1.72}}$ & 70.21$_{\scalebox{0.6}{3.11}}$ & 69.45$_{\scalebox{0.6}{1.83}}$ & 72.16$_{\scalebox{0.6}{0.81}}$ & \cellcolor{tableblue}69.74$_{\scalebox{0.6}{1.94}}$ & 49.10$_{\scalebox{0.6}{1.20}}$ & 51.49$_{\scalebox{0.6}{1.04}}$ & 57.60$_{\scalebox{0.6}{0.49}}$ & 58.97$_{\scalebox{0.6}{1.25}}$ & 61.61$_{\scalebox{0.6}{1.31}}$ & \cellcolor{tableblue}55.75$_{\scalebox{0.6}{5.25}}$ \\
& + \ours & \textbf{72.72}$_{\scalebox{0.6}{1.51}}$ & \textbf{74.65}$_{\scalebox{0.6}{1.26}}$ & \textbf{75.99}$_{\scalebox{0.6}{1.48}}$ & \textbf{76.54}$_{\scalebox{0.6}{1.88}}$ & \textbf{75.77}$_{\scalebox{0.6}{1.46}}$ & \cellcolor{tableblue}\textbf{75.13}$_{\scalebox{0.6}{1.51}}$ & \textbf{51.22}$_{\scalebox{0.6}{1.13}}$ & \textbf{55.65}$_{\scalebox{0.6}{0.88}}$ & \textbf{56.90}$_{\scalebox{0.6}{0.95}}$ & \textbf{60.14}$_{\scalebox{0.6}{1.62}}$ & \textbf{63.38}$_{\scalebox{0.6}{1.95}}$ & \cellcolor{tableblue}\textbf{57.46}$_{\scalebox{0.6}{4.60}}$ \\
\midrule

\multirow{6}{*}{TriviaQA} 
& KD & 53.70$_{\scalebox{0.6}{4.24}}$ & 50.40$_{\scalebox{0.6}{3.27}}$ & 46.49$_{\scalebox{0.6}{2.23}}$ & 45.80$_{\scalebox{0.6}{2.86}}$ & 46.93$_{\scalebox{0.6}{1.44}}$ & \cellcolor{tableblue}48.66$_{\scalebox{0.6}{3.33}}$ & 61.57$_{\scalebox{0.6}{1.09}}$ & \textbf{61.98}$_{\scalebox{0.6}{0.79}}$ & 59.92$_{\scalebox{0.6}{1.06}}$ & 57.50$_{\scalebox{0.6}{1.14}}$ & \textbf{59.26}$_{\scalebox{0.6}{1.01}}$ & \cellcolor{tableblue}60.05$_{\scalebox{0.6}{1.82}}$ \\
& + \ours & \textbf{60.06}$_{\scalebox{0.6}{1.45}}$ & \textbf{56.32}$_{\scalebox{0.6}{2.19}}$ & \textbf{56.47}$_{\scalebox{0.6}{7.04}}$ & \textbf{50.54}$_{\scalebox{0.6}{1.92}}$ & \textbf{49.52}$_{\scalebox{0.6}{1.77}}$ & \cellcolor{tableblue}\textbf{54.58}$_{\scalebox{0.6}{4.43}}$ & \textbf{62.49}$_{\scalebox{0.6}{0.42}}$ & 60.58$_{\scalebox{0.6}{0.69}}$ & \textbf{62.32}$_{\scalebox{0.6}{0.67}}$ & \textbf{60.53}$_{\scalebox{0.6}{0.74}}$ & 57.96$_{\scalebox{0.6}{1.49}}$ & \cellcolor{tableblue}\textbf{60.77}$_{\scalebox{0.6}{1.83}}$ \\
\cmidrule(lr){2-14}
& LWD & 53.73$_{\scalebox{0.6}{3.89}}$ & 51.86$_{\scalebox{0.6}{2.61}}$ & 44.61$_{\scalebox{0.6}{2.48}}$ & 45.08$_{\scalebox{0.6}{1.90}}$ & 47.19$_{\scalebox{0.6}{1.42}}$ & \cellcolor{tableblue}48.50$_{\scalebox{0.6}{4.10}}$ & 61.10$_{\scalebox{0.6}{0.41}}$ & 61.20$_{\scalebox{0.6}{1.11}}$ & 59.26$_{\scalebox{0.6}{0.40}}$ & 58.07$_{\scalebox{0.6}{0.42}}$ & \textbf{58.86}$_{\scalebox{0.6}{1.29}}$ & \cellcolor{tableblue}59.70$_{\scalebox{0.6}{1.40}}$ \\
& + \ours & \textbf{59.70}$_{\scalebox{0.6}{1.51}}$ & \textbf{56.49}$_{\scalebox{0.6}{2.51}}$ & \textbf{56.35}$_{\scalebox{0.6}{7.32}}$ & \textbf{50.46}$_{\scalebox{0.6}{1.88}}$ & \textbf{49.67}$_{\scalebox{0.6}{1.09}}$ & \cellcolor{tableblue}\textbf{54.53}$_{\scalebox{0.6}{4.30}}$ & \textbf{62.25}$_{\scalebox{0.6}{0.68}}$ & \textbf{61.49}$_{\scalebox{0.6}{0.79}}$ & \textbf{61.76}$_{\scalebox{0.6}{0.50}}$ & \textbf{60.21}$_{\scalebox{0.6}{1.16}}$ & 57.87$_{\scalebox{0.6}{1.39}}$ & \cellcolor{tableblue}\textbf{60.72}$_{\scalebox{0.6}{1.76}}$ \\
\cmidrule(lr){2-14}
& SeqKD & 59.01$_{\scalebox{0.6}{3.46}}$ & 52.99$_{\scalebox{0.6}{1.83}}$ & 50.19$_{\scalebox{0.6}{2.62}}$ & 50.53$_{\scalebox{0.6}{2.14}}$ & 48.61$_{\scalebox{0.6}{1.73}}$ & \cellcolor{tableblue}52.27$_{\scalebox{0.6}{4.08}}$ & 61.36$_{\scalebox{0.6}{1.02}}$ & 59.99$_{\scalebox{0.6}{0.54}}$ & 60.52$_{\scalebox{0.6}{0.81}}$ & 59.85$_{\scalebox{0.6}{1.07}}$ & \textbf{60.58}$_{\scalebox{0.6}{1.30}}$ & \cellcolor{tableblue}60.46$_{\scalebox{0.6}{0.60}}$ \\
& + \ours & \textbf{60.70}$_{\scalebox{0.6}{2.14}}$ & \textbf{55.11}$_{\scalebox{0.6}{1.75}}$ & \textbf{54.79}$_{\scalebox{0.6}{1.59}}$ & \textbf{51.05}$_{\scalebox{0.6}{3.29}}$ & \textbf{53.40}$_{\scalebox{0.6}{1.24}}$ & \cellcolor{tableblue}\textbf{55.01}$_{\scalebox{0.6}{3.56}}$ & \textbf{63.13}$_{\scalebox{0.6}{0.84}}$ & \textbf{62.05}$_{\scalebox{0.6}{1.05}}$ & \textbf{61.76}$_{\scalebox{0.6}{0.76}}$ & \textbf{60.17}$_{\scalebox{0.6}{0.64}}$ & 59.86$_{\scalebox{0.6}{0.66}}$ & \cellcolor{tableblue}\textbf{61.39}$_{\scalebox{0.6}{1.36}}$ \\
\midrule

\multirow{6}{*}{HotpotQA} 
& KD & \textbf{70.37}$_{\scalebox{0.6}{2.43}}$ & 70.96$_{\scalebox{0.6}{3.32}}$ & 64.27$_{\scalebox{0.6}{3.26}}$ & 70.23$_{\scalebox{0.6}{1.10}}$ & 69.42$_{\scalebox{0.6}{3.49}}$ & \cellcolor{tableblue}69.05$_{\scalebox{0.6}{2.73}}$ & \textbf{74.09}$_{\scalebox{0.6}{0.48}}$ & \textbf{75.35}$_{\scalebox{0.6}{0.53}}$ & 72.73$_{\scalebox{0.6}{1.19}}$ & 73.15$_{\scalebox{0.6}{0.57}}$ & 73.10$_{\scalebox{0.6}{0.97}}$ & \cellcolor{tableblue}73.68$_{\scalebox{0.6}{1.06}}$ \\
& + \ours & 69.87$_{\scalebox{0.6}{3.08}}$ & \textbf{73.39}$_{\scalebox{0.6}{2.01}}$ & \textbf{71.31}$_{\scalebox{0.6}{1.26}}$ & \textbf{70.65}$_{\scalebox{0.6}{1.90}}$ & \textbf{72.61}$_{\scalebox{0.6}{2.43}}$ & \cellcolor{tableblue}\textbf{71.56}$_{\scalebox{0.6}{1.43}}$ & 73.38$_{\scalebox{0.6}{0.92}}$ & 74.79$_{\scalebox{0.6}{0.50}}$ & \textbf{76.44}$_{\scalebox{0.6}{0.65}}$ & \textbf{75.69}$_{\scalebox{0.6}{0.61}}$ & \textbf{76.17}$_{\scalebox{0.6}{1.07}}$ & \cellcolor{tableblue}\textbf{75.29}$_{\scalebox{0.6}{1.24}}$ \\
\cmidrule(lr){2-14}
& LWD & \textbf{69.89}$_{\scalebox{0.6}{2.41}}$ & 70.77$_{\scalebox{0.6}{3.76}}$ & 64.11$_{\scalebox{0.6}{3.08}}$ & 69.88$_{\scalebox{0.6}{1.68}}$ & 70.44$_{\scalebox{0.6}{2.77}}$ & \cellcolor{tableblue}69.02$_{\scalebox{0.6}{2.77}}$ & \textbf{74.20}$_{\scalebox{0.6}{0.49}}$ & 74.45$_{\scalebox{0.6}{0.78}}$ & 72.40$_{\scalebox{0.6}{0.86}}$ & 74.36$_{\scalebox{0.6}{0.54}}$ & 74.07$_{\scalebox{0.6}{1.84}}$ & \cellcolor{tableblue}73.90$_{\scalebox{0.6}{0.85}}$ \\
& + \ours & 69.21$_{\scalebox{0.6}{3.32}}$ & \textbf{74.09}$_{\scalebox{0.6}{1.82}}$ & \textbf{72.04}$_{\scalebox{0.6}{1.56}}$ & \textbf{70.50}$_{\scalebox{0.6}{2.17}}$ & \textbf{73.28}$_{\scalebox{0.6}{2.23}}$ & \cellcolor{tableblue}\textbf{71.82}$_{\scalebox{0.6}{1.99}}$ & 73.87$_{\scalebox{0.6}{0.91}}$ & \textbf{74.77}$_{\scalebox{0.6}{0.41}}$ & \textbf{74.70}$_{\scalebox{0.6}{0.53}}$ & \textbf{75.75}$_{\scalebox{0.6}{0.89}}$ & \textbf{77.32}$_{\scalebox{0.6}{0.82}}$ & \cellcolor{tableblue}\textbf{75.28}$_{\scalebox{0.6}{1.32}}$ \\
\cmidrule(lr){2-14}
& SeqKD & \textbf{75.81}$_{\scalebox{0.6}{0.84}}$ & 75.65$_{\scalebox{0.6}{0.88}}$ & 75.79$_{\scalebox{0.6}{1.16}}$ & 74.16$_{\scalebox{0.6}{2.96}}$ & \textbf{74.22}$_{\scalebox{0.6}{3.15}}$ & \cellcolor{tableblue}75.13$_{\scalebox{0.6}{0.86}}$ & \textbf{75.37}$_{\scalebox{0.6}{0.66}}$ & \textbf{76.49}$_{\scalebox{0.6}{0.78}}$ & 75.26$_{\scalebox{0.6}{0.50}}$ & 72.29$_{\scalebox{0.6}{0.71}}$ & 73.59$_{\scalebox{0.6}{1.65}}$ & \cellcolor{tableblue}74.60$_{\scalebox{0.6}{1.66}}$ \\
& + \ours & 74.84$_{\scalebox{0.6}{0.98}}$ & \textbf{76.99}$_{\scalebox{0.6}{0.54}}$ & \textbf{75.84}$_{\scalebox{0.6}{1.63}}$ & \textbf{75.03}$_{\scalebox{0.6}{2.36}}$ & 73.14$_{\scalebox{0.6}{3.32}}$ & \cellcolor{tableblue}\textbf{75.17}$_{\scalebox{0.6}{1.41}}$ & 73.67$_{\scalebox{0.6}{0.66}}$ & 74.07$_{\scalebox{0.6}{1.30}}$ & \textbf{76.23}$_{\scalebox{0.6}{0.70}}$ & \textbf{76.04}$_{\scalebox{0.6}{0.56}}$ & \textbf{74.90}$_{\scalebox{0.6}{1.46}}$ & \cellcolor{tableblue}\textbf{74.98}$_{\scalebox{0.6}{1.14}}$ \\
\midrule

\rowcolor{tableblue} 
& KD & 69.01$_{\scalebox{0.6}{12.31}}$ & 68.08$_{\scalebox{0.6}{12.77}}$ & 65.06$_{\scalebox{0.6}{13.63}}$ & 65.39$_{\scalebox{0.6}{14.56}}$ & 64.58$_{\scalebox{0.6}{13.42}}$ & \cellcolor{tableblue}66.42$_{\scalebox{0.6}{13.25}}$ & 67.53$_{\scalebox{0.6}{8.48}}$ & 66.85$_{\scalebox{0.6}{9.44}}$ & 65.81$_{\scalebox{0.6}{9.04}}$ & 65.26$_{\scalebox{0.6}{9.20}}$ & 65.62$_{\scalebox{0.6}{9.45}}$ & \cellcolor{tableblue}66.21$_{\scalebox{0.6}{9.09}}$ \\
\rowcolor{tableblue} 
& + \ours & \textbf{70.46}$_{\scalebox{0.6}{10.57}}$ & \textbf{70.44}$_{\scalebox{0.6}{11.60}}$ & \textbf{69.88}$_{\scalebox{0.6}{11.55}}$ & \textbf{67.98}$_{\scalebox{0.6}{13.55}}$ & \textbf{67.57}$_{\scalebox{0.6}{13.57}}$ & \cellcolor{tableblue}\textbf{69.27}$_{\scalebox{0.6}{12.09}}$ & \textbf{67.97}$_{\scalebox{0.6}{6.97}}$ & \textbf{67.74}$_{\scalebox{0.6}{8.48}}$ & \textbf{68.06}$_{\scalebox{0.6}{8.89}}$ & \textbf{67.52}$_{\scalebox{0.6}{8.66}}$ & \textbf{67.34}$_{\scalebox{0.6}{8.70}}$ & \cellcolor{tableblue}\textbf{67.72}$_{\scalebox{0.6}{8.28}}$ \\
\cmidrule(lr){2-14}
\rowcolor{tableblue} 
& LWD & 68.91$_{\scalebox{0.6}{12.29}}$ & 68.20$_{\scalebox{0.6}{12.24}}$ & 64.67$_{\scalebox{0.6}{14.34}}$ & 65.17$_{\scalebox{0.6}{14.37}}$ & 64.90$_{\scalebox{0.6}{13.30}}$ & \cellcolor{tableblue}66.37$_{\scalebox{0.6}{13.20}}$ & 67.28$_{\scalebox{0.6}{7.91}}$ & 67.51$_{\scalebox{0.6}{8.03}}$ & 66.24$_{\scalebox{0.6}{8.42}}$ & 66.34$_{\scalebox{0.6}{8.77}}$ & 66.49$_{\scalebox{0.6}{9.31}}$ & \cellcolor{tableblue}66.77$_{\scalebox{0.6}{8.47}}$ \\
\rowcolor{tableblue} 
\textbf{AVG} & + \ours & \textbf{70.21}$_{\scalebox{0.6}{10.66}}$ & \textbf{70.62}$_{\scalebox{0.6}{11.77}}$ & \textbf{69.63}$_{\scalebox{0.6}{11.55}}$ & \textbf{68.07}$_{\scalebox{0.6}{13.41}}$ & \textbf{67.89}$_{\scalebox{0.6}{13.65}}$ & \cellcolor{tableblue}\textbf{69.28}$_{\scalebox{0.6}{12.11}}$ & \textbf{67.39}$_{\scalebox{0.6}{7.21}}$ & \textbf{67.96}$_{\scalebox{0.6}{8.20}}$ & \textbf{68.01}$_{\scalebox{0.6}{8.34}}$ & \textbf{68.23}$_{\scalebox{0.6}{8.24}}$ & \textbf{68.44}$_{\scalebox{0.6}{9.09}}$ & \cellcolor{tableblue}\textbf{68.00}$_{\scalebox{0.6}{8.12}}$ \\
\cmidrule(lr){2-14}
\rowcolor{tableblue} 
& SeqKD & 69.54$_{\scalebox{0.6}{11.05}}$ & 68.84$_{\scalebox{0.6}{13.18}}$ & 68.51$_{\scalebox{0.6}{13.62}}$ & 67.87$_{\scalebox{0.6}{13.39}}$ & 68.41$_{\scalebox{0.6}{13.73}}$ & \cellcolor{tableblue}68.63$_{\scalebox{0.6}{12.87}}$ & 65.99$_{\scalebox{0.6}{11.67}}$ & 66.66$_{\scalebox{0.6}{11.13}}$ & 67.62$_{\scalebox{0.6}{8.88}}$ & 67.16$_{\scalebox{0.6}{7.92}}$ & 68.35$_{\scalebox{0.6}{7.61}}$ & \cellcolor{tableblue}67.16$_{\scalebox{0.6}{9.30}}$ \\
\rowcolor{tableblue} 
& + \ours & \textbf{71.24}$_{\scalebox{0.6}{10.62}}$ & \textbf{70.82}$_{\scalebox{0.6}{12.19}}$ & \textbf{70.77}$_{\scalebox{0.6}{12.42}}$ & \textbf{70.03}$_{\scalebox{0.6}{13.59}}$ & \textbf{69.74}$_{\scalebox{0.6}{12.44}}$ & \cellcolor{tableblue}\textbf{70.52}$_{\scalebox{0.6}{12.16}}$ & \textbf{66.37}$_{\scalebox{0.6}{10.14}}$ & \textbf{67.32}$_{\scalebox{0.6}{8.96}}$ & \textbf{68.27}$_{\scalebox{0.6}{9.08}}$ & \textbf{68.45}$_{\scalebox{0.6}{8.42}}$ & \textbf{69.00}$_{\scalebox{0.6}{7.80}}$ & \cellcolor{tableblue}\textbf{67.88}$_{\scalebox{0.6}{8.69}}$ \\
\bottomrule
\end{tabular}
}
\end{table*}

\textbf{Baselines.}
We compare \ours against three distillation baselines: (i) Standard Knowledge Distillation (KD)~\citep{hinton2015distilling}, (ii) Sequence-Level Knowledge Distillation (seqKD)~\citep{kim2016sequence}, and (iii) Layer-Wise Distillation (LWD)~\citep{jiao2020tinybert}. All methods use the same few-shot splits for fair comparison. Additional details of the baselines and implementation details are provided in Appendix~\ref{app:baselines}.

\textbf{Setup.}
Following prior work~\citep{hamman2025few}, we first fine-tune the teacher model on the full training dataset. For each few-shot budget $k \in \{16,32,64,128,256\}$, we rank factual samples by their teacher prediction margins, using margin sampling~\citep{scheffer2001active}. The factual-only distillation baseline uses the $k$ samples with the smallest margins. \ours instead uses the $k/2$ factual samples with the smallest margins together with their corresponding $k/2$ counterfactuals, maintaining the same total few-shot budget $k$. We evaluate accuracy using EM and F1 against the gold answers and teacher-student agreement using EM and F1 between their generated answers. All results are reported as the mean and standard deviation over five random seeds.

\subsection{Main Experimental Results}

\textbf{\ours outperforms the baselines.}
As shown in Table~\ref{tab:f1_t5_qwen_combined_tiny_subscript}, \ours improves performance across all three distillation baselines under the T5 setting. When averaged across all datasets and $k$-shot settings, \ours improves the F1 score from $66.42$ to $69.27$ with standard KD, from $66.37$ to $69.28$ with LWD, and from $68.63$ to $70.52$ with SeqKD. Moreover, \ours exhibits lower standard deviations than the corresponding baselines in many settings, indicating more stable performance across seeds. These results demonstrate that counterfactual supervision provides consistent benefits over the baselines. Detailed EM scores are provided in Table~\ref{tab:indomain_em_all_baselines} in Appendix~\ref{app:emindomain}.

\textbf{\ours generalizes across different model families.}
\ours improves standard KD across both the encoder-decoder T5 and decoder-only Qwen2.5 model families, where the teachers are approximately $3.5\times$ and $4.7\times$ larger than their respective students. Under Qwen2.5, \ours improves the average F1 score from $66.21$ to $67.72$ with standard KD, from $66.77$ to $68.00$ with LWD, and from $\text{67.16}$ to $\text{67.88}$ with SeqKD. These results indicate that the benefits of \ours extend across different model architectures and scales.

\begin{figure*}[t]
    \centering
    \includegraphics[width=\textwidth]{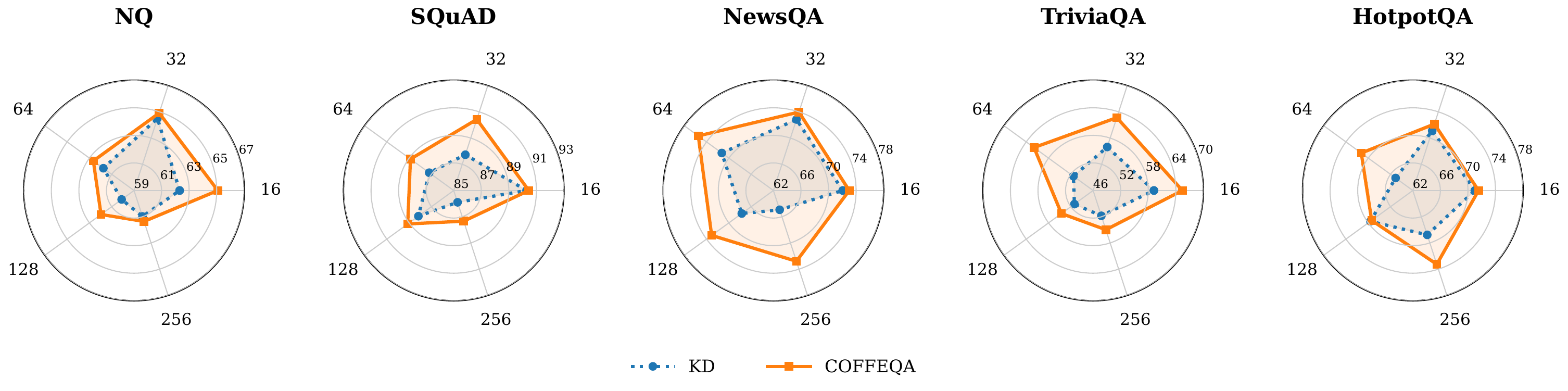}
    \caption{Teacher-student F1 agreement of standard KD and \ours across datasets and $k$-shot settings using T5. \ours generally achieves higher agreement than the baseline. Spokes denote $k$-shot settings, and radial axis F1 score ranges are adjusted independently by dataset for visualization.}
    \label{fig:agreement_f1_radar}
\end{figure*}

\textbf{Students mimic the teacher more closely under \ours.}
To examine how closely the student follows the teacher, we measure teacher-student agreement using EM and F1. As shown in Figure~\ref{fig:agreement_f1_radar}, \ours generally achieves higher teacher-student agreement than standard KD across datasets and $k$-shot settings. Averaged across all benchmarks and settings, \ours improves F1 agreement under standard KD from $68.51$ to $71.50$ under T5, with a similar trend for Qwen2.5. These results suggest that counterfactual supervision helps the student more closely reproduce the teacher's output behavior. Detailed results and additional radar plots are provided in Appendix~\ref{app:agr}.

\textbf{\ours generalizes well to out-of-domain datasets.}
To evaluate out-of-domain generalization, we use Natural Questions for distillation and evaluate the resulting students on three out-of-domain test sets: DROP, DuoRC, and RelationExtraction. As shown in Table~\ref{tab:ood_f1_t5_qwen}, \ours improves over standard KD when averaged across datasets and $k$-shot settings for both model families. For T5 and Qwen2.5, \ours improves the overall F1 score over standard KD from $53.35$ to $54.52$ and from $57.09$ to $57.77$, respectively. These results indicate that the benefits of counterfactual supervision extend beyond the in-domain datasets used for distillation. Detailed EM scores are in Appendix~\ref{app:ood}.

\begin{table*}[t]
\caption{Out-of-domain F1 score comparison across $k$-shots for T5 and Qwen2.5 models using KD. Mean F1 scores are reported with standard deviations as subscripts ($Mean_{std}$). Bold values indicate the better-performing method (\ours vs. KD). Blue shading highlights aggregate results.}
\label{tab:ood_f1_t5_qwen}
\centering
\fontsize{7pt}{9pt}\selectfont
\setlength{\tabcolsep}{1.2pt}
\renewcommand{\arraystretch}{0.95}

\definecolor{tableblue}{HTML}{E6F0FF}

\resizebox{\textwidth}{!}{
\begin{tabular}{llcccccccccccccc}
\toprule
 & & \multicolumn{6}{c}{\textbf{T5}} & \multicolumn{6}{c}{\textbf{Qwen2.5}} \\
\cmidrule(lr){3-8} \cmidrule(lr){9-14}
Dataset & Method & 16 & 32 & 64 & 128 & 256 & \cellcolor{tableblue}Avg & 16 & 32 & 64 & 128 & 256 & \cellcolor{tableblue}Avg \\
\midrule

\multirow{2}{*}{DROP} 
& KD & 38.56$_{\scalebox{0.6}{1.72}}$ & 35.81$_{\scalebox{0.6}{3.76}}$ & \textbf{32.48}$_{\scalebox{0.6}{2.47}}$ & 28.06$_{\scalebox{0.6}{3.65}}$ & 27.78$_{\scalebox{0.6}{1.41}}$ & \cellcolor{tableblue}32.54$_{\scalebox{0.6}{4.73}}$ & \textbf{49.58}$_{\scalebox{0.6}{1.13}}$ & 44.92$_{\scalebox{0.6}{1.31}}$ & 42.07$_{\scalebox{0.6}{0.45}}$ & 41.28$_{\scalebox{0.6}{0.67}}$ & \textbf{42.57}$_{\scalebox{0.6}{0.94}}$ & \cellcolor{tableblue}44.08$_{\scalebox{0.6}{3.36}}$ \\
& + \ours & \textbf{39.24}$_{\scalebox{0.6}{1.14}}$ & \textbf{37.54}$_{\scalebox{0.6}{1.20}}$ & 32.41$_{\scalebox{0.6}{2.71}}$ & \textbf{32.29}$_{\scalebox{0.6}{2.00}}$ & \textbf{29.44}$_{\scalebox{0.6}{1.53}}$ & \cellcolor{tableblue}\textbf{34.19}$_{\scalebox{0.6}{4.06}}$ & 48.58$_{\scalebox{0.6}{0.79}}$ & \textbf{49.53}$_{\scalebox{0.6}{1.23}}$ & \textbf{45.20}$_{\scalebox{0.6}{1.31}}$ & \textbf{42.65}$_{\scalebox{0.6}{1.43}}$ & 41.36$_{\scalebox{0.6}{0.60}}$ & \cellcolor{tableblue}\textbf{45.46}$_{\scalebox{0.6}{3.58}}$ \\
\cmidrule(lr){1-14}

\multirow{2}{*}{DuoRC} 
& KD & 50.38$_{\scalebox{0.6}{1.23}}$ & 48.49$_{\scalebox{0.6}{3.56}}$ & \textbf{46.98}$_{\scalebox{0.6}{2.07}}$ & 42.52$_{\scalebox{0.6}{1.73}}$ & 44.46$_{\scalebox{0.6}{1.80}}$ & \cellcolor{tableblue}46.57$_{\scalebox{0.6}{3.13}}$ & \textbf{41.73}$_{\scalebox{0.6}{0.77}}$ & 44.84$_{\scalebox{0.6}{1.29}}$ & \textbf{43.38}$_{\scalebox{0.6}{1.30}}$ & 42.77$_{\scalebox{0.6}{1.16}}$ & 42.71$_{\scalebox{0.6}{1.49}}$ & \cellcolor{tableblue}43.08$_{\scalebox{0.6}{1.15}}$ \\
& + \ours & \textbf{51.66}$_{\scalebox{0.6}{0.56}}$ & \textbf{50.58}$_{\scalebox{0.6}{1.02}}$ & 44.55$_{\scalebox{0.6}{2.64}}$ & \textbf{46.14}$_{\scalebox{0.6}{1.23}}$ & \textbf{44.72}$_{\scalebox{0.6}{1.51}}$ & \cellcolor{tableblue}\textbf{47.53}$_{\scalebox{0.6}{3.36}}$ & 41.04$_{\scalebox{0.6}{1.11}}$ & \textbf{45.01}$_{\scalebox{0.6}{0.92}}$ & 43.11$_{\scalebox{0.6}{0.98}}$ & \textbf{43.25}$_{\scalebox{0.6}{1.00}}$ & \textbf{44.39}$_{\scalebox{0.6}{1.42}}$ & \cellcolor{tableblue}\textbf{43.36}$_{\scalebox{0.6}{1.52}}$ \\
\cmidrule(lr){1-14}

\multirow{2}{*}{RelationExtraction} 
& KD & 82.96$_{\scalebox{0.6}{0.99}}$ & \textbf{83.30}$_{\scalebox{0.6}{0.46}}$ & 82.65$_{\scalebox{0.6}{0.61}}$ & 78.30$_{\scalebox{0.6}{2.66}}$ & 77.53$_{\scalebox{0.6}{1.47}}$ & \cellcolor{tableblue}80.95$_{\scalebox{0.6}{2.79}}$ & \textbf{85.32}$_{\scalebox{0.6}{0.45}}$ & 85.34$_{\scalebox{0.6}{0.49}}$ & 84.36$_{\scalebox{0.6}{0.25}}$ & 82.77$_{\scalebox{0.6}{0.71}}$ & 82.74$_{\scalebox{0.6}{0.19}}$ & \cellcolor{tableblue}84.11$_{\scalebox{0.6}{1.30}}$ \\
& + \ours & \textbf{82.98}$_{\scalebox{0.6}{0.39}}$ & 82.99$_{\scalebox{0.6}{0.65}}$ & \textbf{83.22}$_{\scalebox{0.6}{1.22}}$ & \textbf{80.45}$_{\scalebox{0.6}{1.51}}$ & \textbf{79.55}$_{\scalebox{0.6}{1.50}}$ & \cellcolor{tableblue}\textbf{81.84}$_{\scalebox{0.6}{1.71}}$ & 84.71$_{\scalebox{0.6}{0.55}}$ & \textbf{85.66}$_{\scalebox{0.6}{0.33}}$ & \textbf{85.62}$_{\scalebox{0.6}{0.18}}$ & \textbf{83.61}$_{\scalebox{0.6}{0.77}}$ & \textbf{82.88}$_{\scalebox{0.6}{0.80}}$ & \cellcolor{tableblue}\textbf{84.50}$_{\scalebox{0.6}{1.23}}$ \\
\midrule

% OVERALL AVERAGE ROWS
\rowcolor{tableblue} 
& KD & 57.30$_{\scalebox{0.6}{23.00}}$ & 55.87$_{\scalebox{0.6}{24.59}}$ & \textbf{54.04}$_{\scalebox{0.6}{25.82}}$ & 49.63$_{\scalebox{0.6}{25.86}}$ & 49.92$_{\scalebox{0.6}{25.32}}$ & \cellcolor{tableblue}53.35$_{\scalebox{0.6}{24.91}}$ & \textbf{58.88}$_{\scalebox{0.6}{23.24}}$ & 58.37$_{\scalebox{0.6}{23.36}}$ & 56.60$_{\scalebox{0.6}{24.05}}$ & 55.61$_{\scalebox{0.6}{23.54}}$ & 56.00$_{\scalebox{0.6}{23.15}}$ & \cellcolor{tableblue}57.09$_{\scalebox{0.6}{23.40}}$ \\
\rowcolor{tableblue} 
\textbf{AVG} & + \ours & \textbf{57.96}$_{\scalebox{0.6}{22.54}}$ & \textbf{57.03}$_{\scalebox{0.6}{23.40}}$ & 53.39$_{\scalebox{0.6}{26.53}}$ & \textbf{52.96}$_{\scalebox{0.6}{24.79}}$ & \textbf{51.24}$_{\scalebox{0.6}{25.69}}$ & \cellcolor{tableblue}\textbf{54.52}$_{\scalebox{0.6}{24.58}}$ & 58.11$_{\scalebox{0.6}{23.34}}$ & \textbf{60.07}$_{\scalebox{0.6}{22.28}}$ & \textbf{57.97}$_{\scalebox{0.6}{23.96}}$ & \textbf{56.50}$_{\scalebox{0.6}{23.48}}$ & \textbf{56.21}$_{\scalebox{0.6}{23.15}}$ & \cellcolor{tableblue}\textbf{57.77}$_{\scalebox{0.6}{23.17}}$ \\
\bottomrule
\end{tabular}%
}
\end{table*}

\textbf{\ours improves performance under modified contextual evidence.}
Following~\citet{longpre2021entity}, we replace each test answer in the context with a randomly selected answer of the same entity type and evaluate against the modified target. As shown in Table~\ref{tab:counterfactual_f1}, \ours improves average F1 agreement over standard KD from $66.72$ to $69.55$ for T5 and from $49.23$ to $53.55$ for Qwen2.5. This suggests that \ours better adapts predictions to modified contextual evidence when it differs from pre-trained knowledge. Corresponding EM results are provided in Appendix~\ref{app:cfperf}.

\subsection{Ablation Studies}
We conduct three ablation studies to examine different components of \ours: (i) We compare \ours with a variant that uses generated counterfactuals without CFD verification to isolate the contribution of CFD; (ii) We vary the CFD threshold $\tau$ to analyze its effect on downstream performance and teacher-student agreement; and (iii) We measure counterfactual generation costs, training time and energy consumption across baselines to analyze the computational cost of \ours. Full ablation results and experimental details are in Appendix~\ref{app:ablation}.

\section{Conclusion} 
In this paper, we present a counterfactual-guided knowledge distillation method for few-shot open-book question answering. Motivated by the difficulty of distilling with factual samples near the teacher's decision boundary, our approach uses factual-counterfactual pairs to provide additional supervision around these challenging regions. We further introduce CFD to validate counterfactuals before distillation. Experiments across multiple datasets and LLM architectures demonstrate improved knowledge transfer under limited supervision. Limitations and future directions are discussed in Appendix~\ref{app:limitations}. Overall, our findings highlight the potential of counterfactual supervision for improving knowledge transfer in generative QA under limited supervision in the few-shot setting.

\subsection*{Reproducibility statement}
To support reproducibility, we provide detailed descriptions of the datasets,
training procedures, evaluation settings, and baseline implementations in
Appendix~\ref{app:setup}. Specifically, Appendix~\ref{app:benchmarks} describes
the datasets and experimental splits, Appendix~\ref{app:finetuning} provides
details of teacher model fine-tuning and evaluation, and
Appendix~\ref{app:baselines} describes the baseline formulations and
implementation details. We also provide the assumptions and complete proof of
our geometric analysis in Appendix~\ref{app:proof}. 
\newpage
\bibliography{iclr2027_conference}
\bibliographystyle{iclr2027_conference}

\newpage
\appendix
\section{Appendix}
\subsection*{Contents}
\begin{itemize}
    \item \textbf{Section \hyperref[app:limitations]{B}} Limitations and Future Work
    
    \item \textbf{Section \hyperref[app:boundary]{C}} KD Performance on Boundary-Near Factual Samples
    
    \item \textbf{Section \hyperref[app:prompts]{D}} Counterfactual Generation Prompts
    
    \item \textbf{Section \hyperref[app:proof]{E}} Proof of Proposition~\ref{prop:dev_bound}
    
    \item \textbf{Section \hyperref[app:setup]{F}} Experimental Setup Details
    \begin{itemize}
        \item \textbf{Section \hyperref[app:benchmarks]{F.1}} Datasets
        \item \textbf{Section \hyperref[app:finetuning]{F.2}} Teacher Model Fine-Tuning and Evaluation
        \item \textbf{Section \hyperref[app:baselines]{F.3}} Additional Details of the Baselines
    \end{itemize}

    \item \textbf{Section \hyperref[app:ablation]{G}} Ablation Studies
    \begin{itemize}
        \item \textbf{Section \hyperref[app:nocfd]{G.1}} Effect of CFD Validation
        \item \textbf{Section \hyperref[app:cfdthresh]{G.2}} CFD Threshold Ablation
        \item \textbf{Section \hyperref[app:cost]{G.3}} Cost Analysis
    \end{itemize}
    
    \item \textbf{Section \hyperref[app:exp]{H}} Additional Experimental Results
    \begin{itemize}
        \item \textbf{Section \hyperref[app:emindomain]{H.1}} In-Domain EM Results
        \item \textbf{Section \hyperref[app:agr]{H.2}} In-Domain Agreement Results
        \item \textbf{Section \hyperref[app:ood]{H.3}} Out-of-Domain EM Results
        \item \textbf{Section \hyperref[app:cfperf]{H.4}} Performance under Modified Contextual Evidence
    \end{itemize}

\end{itemize}

\newpage
\section{Limitations and Future Work}
\label{app:limitations}

Our current counterfactual generation framework is limited to QA samples with entity-based answers. Extending counterfactual generation beyond these could broaden the applicability of \ours. Future work could explore more general counterfactual generation strategies that can accommodate diverse answer types while preserving the intended contextual changes.

The current formulation of \ours is also designed specifically for open-book question answering, where counterfactuals can be constructed by modifying the supporting context while preserving the question. Extending the definition, generation, and validation of counterfactuals to other generative tasks is an important direction for investigating whether counterfactual-guided few-shot distillation generalizes beyond open-book QA.

\section{KD Performance on Boundary-Near Factual Samples}
\label{app:boundary}

To investigate how proximity to the teacher's decision boundary affects knowledge distillation, we group factual samples according to their prediction margins under the teacher model. Intuitively, a small margin indicates that the teacher assigns similar probabilities to its two most likely token predictions, reflecting greater uncertainty in its output. We use margin sampling~\citep{scheffer2001active} as a proxy for proximity to the teacher's decision boundary.

\textbf{Measuring proximity to the decision boundary.}
For a factual sample $x=(Q,C,A)$ with target answer sequence $y=(y_1,\ldots,y_T)$, the teacher produces a probability distribution over the vocabulary at each answer position $t$. Let $v_{(1)}^t$ and $v_{(2)}^t$ denote the teacher's most likely and next most likely tokens, respectively, at position $t$, conditioned on the input $x$ and the preceding target tokens $y_{<t}$. We define the token-level prediction margin as
\begin{equation}
    m_t(x,y)
    =
    \log P_{\theta}\!\left(v_{(1)}^t \mid x,y_{<t}\right)
    -
    \log P_{\theta}\!\left(v_{(2)}^t \mid x,y_{<t}\right).
    \label{eq:token_margin}
\end{equation}
The prediction margin for the complete answer sequence is then computed as the mean of the token-level margins:
\begin{equation}
    \operatorname{margin}(x,y)
    =
    \frac{1}{T}\sum_{t=1}^{T}m_t(x,y).
    \label{eq:margin_sampling}
\end{equation}
Here, $v_{(1)}^t$ and $v_{(2)}^t$ are the teacher's own top-1 and top-2 predictions at each position and are not defined according to whether they match the ground-truth token. A smaller average margin indicates greater ambiguity between the teacher's two most likely token predictions. We therefore use this quantity as a proxy for proximity to the teacher's decision boundary.

For each dataset, we sort the factual samples in ascending order of their teacher prediction margins and divide them into three margin bins, where Bin~0 contains samples with the smallest margins, Bin~1 contains samples with intermediate margins, and Bin~2 contains samples with the largest margins. To evaluate the effect of teacher prediction margin on knowledge distillation under a controlled few-shot budget, we sample 100 factual examples from each bin and perform knowledge distillation separately using the samples from each bin. Thus, each setting uses the same distillation budget of 100 factual samples, with the primary difference being their teacher prediction margins.

\begin{figure*}[ht]
    \centering

    % Aggregated results
    \begin{minipage}[t]{0.47\textwidth}
        \centering
        \includegraphics[width=\linewidth]{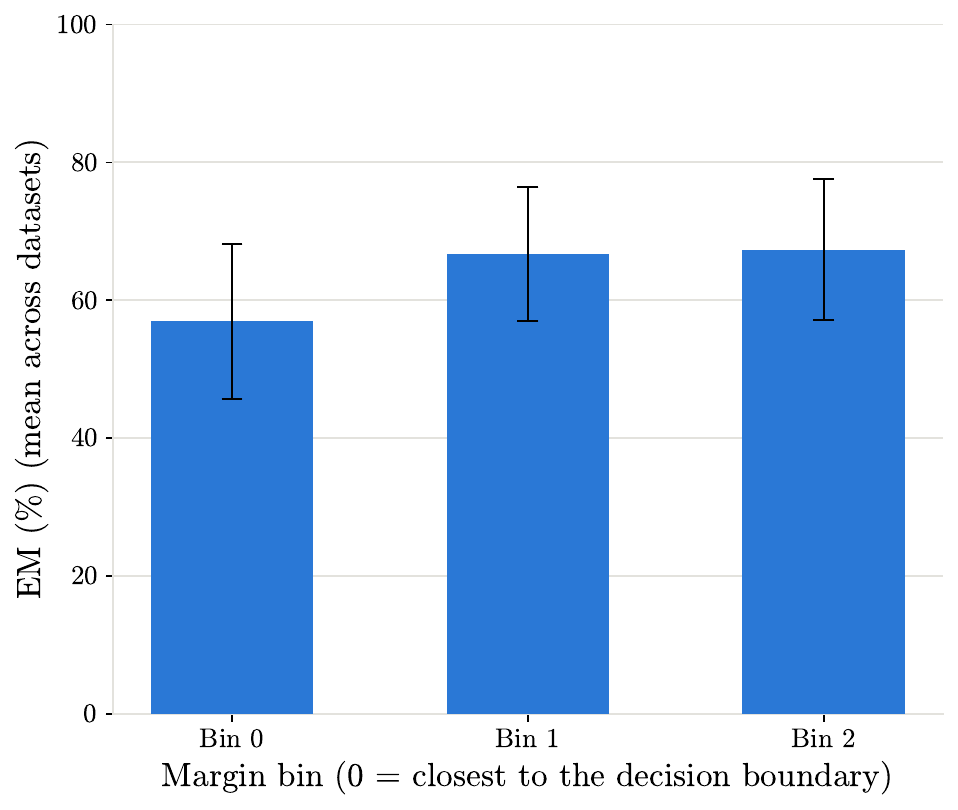}
        \vspace{-2mm}
        \centerline{\small (a) Aggregated EM}
    \end{minipage}
    \hfill
    \begin{minipage}[t]{0.47\textwidth}
        \centering
        \includegraphics[width=\linewidth]{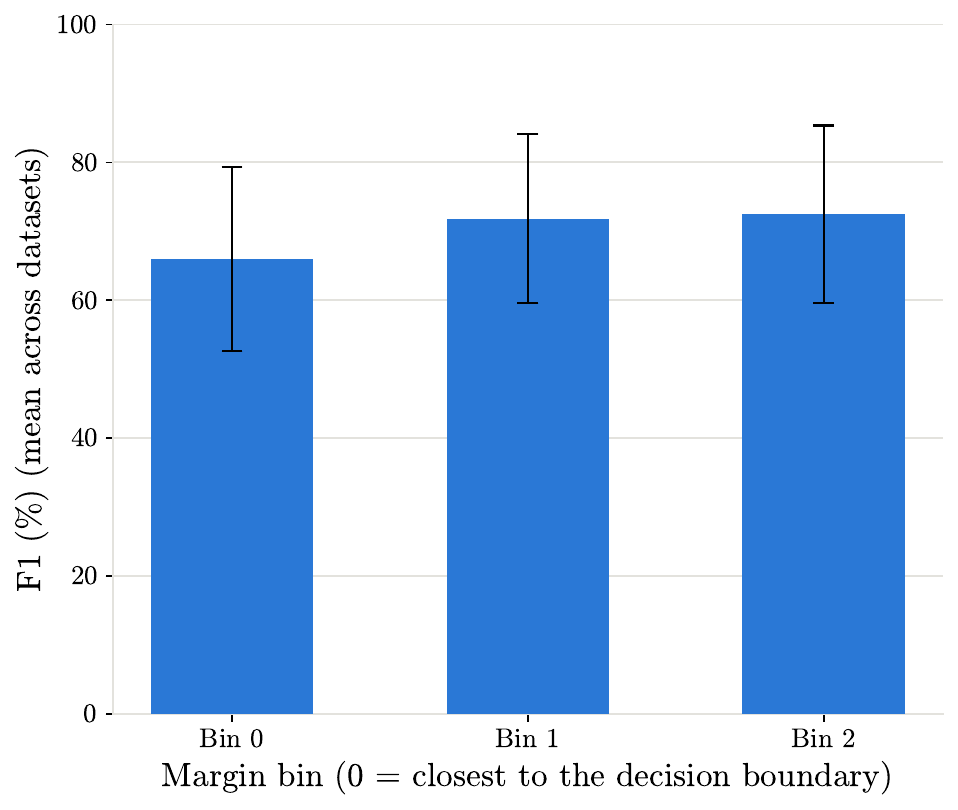}
        \vspace{-2mm}
        \centerline{\small (b) Aggregated F1}
    \end{minipage}

    \vspace{3mm}

    % Dataset-wise results
    \begin{minipage}[t]{0.47\textwidth}
        \centering
        \includegraphics[width=\linewidth]{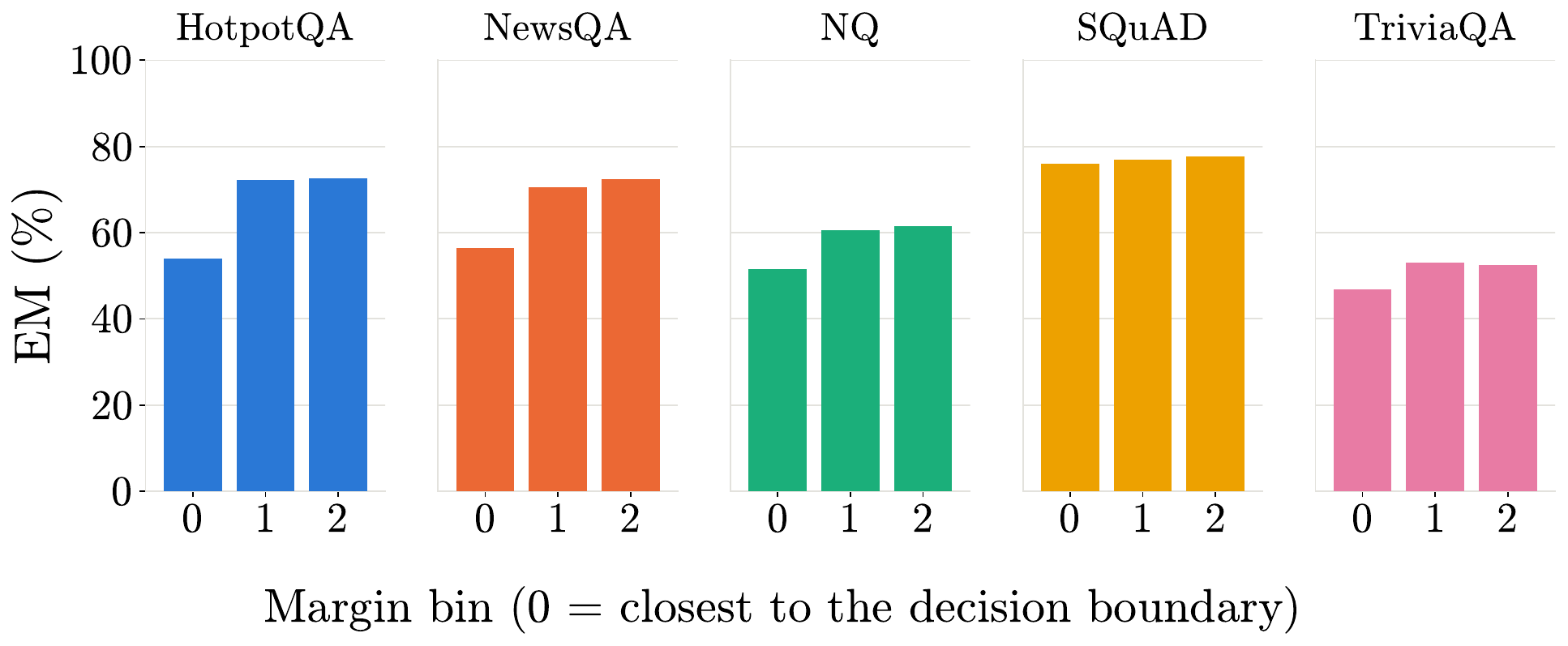}
        \vspace{-2mm}
        \centerline{\small (c) Dataset-wise EM}
    \end{minipage}
    \hfill
    \begin{minipage}[t]{0.47\textwidth}
        \centering
        \includegraphics[width=\linewidth]{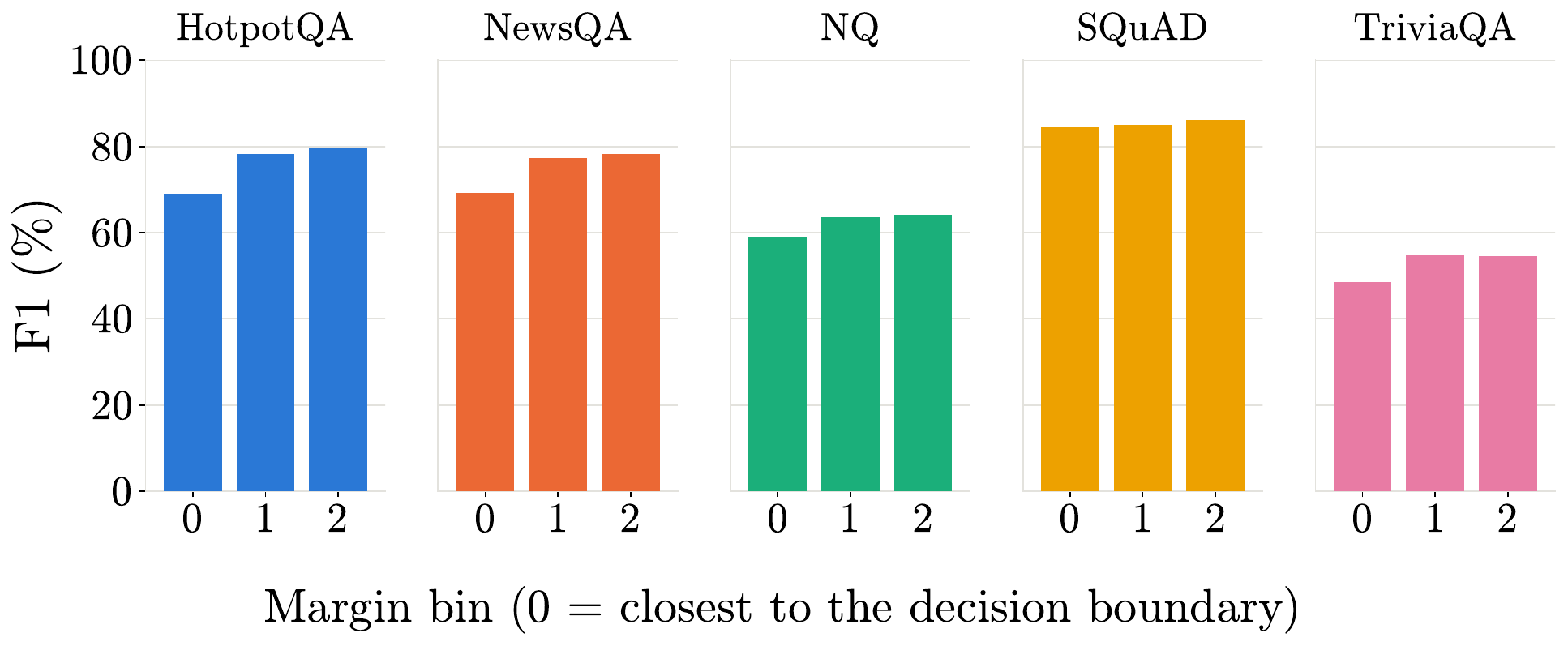}
        \vspace{-2mm}
        \centerline{\small (d) Dataset-wise F1}
    \end{minipage}
    \vspace{10pt}
    \caption{\textbf{KD performance as a function of the teacher prediction margin.}
    Factual samples are grouped into three bins according to their average token-level prediction margins, with Bin~0 containing the smallest margin samples and Bin~2 containing the largest margin samples. Smaller margins are used as a proxy for greater proximity to the teacher's decision boundary. Knowledge distillation is performed separately using 100 factual samples from each bin, maintaining the same distillation budget across all three settings. Panels (a) and (b) report EM and F1 averaged across the five in-domain datasets, respectively, while panels (c) and (d) report the corresponding dataset-wise results. Distillation using samples from the smallest margin bin generally results in lower performance than using samples from the larger margin bins.}
    \label{fig:boundary-performance}
\end{figure*}

Figure~\ref{fig:boundary-performance} reports the resulting KD performance when distillation is performed using 100 factual samples from each of the three margin bins. Figures~\ref{fig:boundary-performance}(a) and~\ref{fig:boundary-performance}(b) show EM and F1 averaged across the five in-domain datasets, respectively. For both metrics, distillation using the smallest-margin samples in Bin~0 results in substantially lower performance than using samples from Bins~1 and~2. The mean EM increases from $56.94\%$ in Bin 0 to $67\%$ in Bins 1–2, while the mean F1 increases from $65.99\%$ to $72–73\%$ in Bins 1–2. In contrast, the difference between Bins~1 and~2 is relatively small for both metrics. This suggests that the main degradation in distillation performance occurs when the few-shot factual supervision consists of samples with the smallest teacher prediction margins.

Figures~\ref{fig:boundary-performance}(c) and~\ref{fig:boundary-performance}(d) show the corresponding dataset-wise EM and F1 results for HotpotQA, NewsQA, Natural Questions (NQ), SQuAD, and TriviaQA. The same general pattern is observed across the five datasets: distillation using samples from Bin~0 results in lower EM and F1 than using samples from the larger-margin bins. The difference is particularly pronounced for HotpotQA, NewsQA, TriviaQA and NQ, whereas SQuAD exhibits a smaller difference across the three bins. Thus, the lower distillation performance associated with the smallest-margin samples is observed across datasets rather than being driven by a single benchmark.

Overall, these results support our motivating observation that factual samples with small teacher prediction margins are more challenging to distill under limited factual supervision. Since a small margin reflects greater ambiguity between the teacher's most likely token predictions, we use these samples as an operational approximation of boundary-near factual examples. Importantly, this degradation is observed while maintaining the same 100 sample distillation budget across all three margin bins. This observation motivates \ours to complement boundary-near factual samples with counterfactual supervision, providing additional information about how the teacher's response changes under modified contextual evidence.

\section{Counterfactual Generation Prompts}
\label{app:prompts}

We use GPT-4o-mini~\citep{openai2024gpt4omini} for counterfactual
generation. The same prompts are used for both the T5 and Qwen2.5
experiments. Counterfactual generation consists of three stages:
(i) identifying the entity type of the original answer,
(ii) generating a plausible alternative entity of the same type, and
(iii) minimally modifying the original context to support the alternative
answer. The prompts used in these stages are shown in
Figure~\ref{fig:cf_generation_prompts}.

\begin{figure*}[ht]
\centering

% ---------- Prompt 1 ----------
\begin{tcolorbox}[
    title=\textbf{Entity Type Classification},
    colback=white,
    colframe=gray!70,
    colbacktitle=gray!70,
    coltitle=white,
    fonttitle=\bfseries\footnotesize,
    boxrule=0.8pt,
    arc=2mm,
    left=3mm,
    right=3mm,
    top=2mm,
    bottom=2mm
]
\footnotesize
What type of named entity is this answer:
``\texttt{\{answer\_text\}}''?

Classify as PERSON, LOCATION, ORGANIZATION, DATE, NUMERIC, or OTHER if
none apply.
\end{tcolorbox}

\vspace{-1mm}

% ---------- Prompt 2 ----------
\begin{tcolorbox}[
    title=\textbf{Alternative Entity Generation},
    colback=white,
    colframe=gray!70,
    colbacktitle=gray!70,
    coltitle=white,
    fonttitle=\bfseries\footnotesize,
    boxrule=0.8pt,
    arc=2mm,
    left=3mm,
    right=3mm,
    top=2mm,
    bottom=2mm
]
\footnotesize
The original answer is ``\texttt{\{original\_answer\}}'', which is a
\texttt{\{entity\_type\}}.

Propose ONE plausible alternative \texttt{\{entity\_type\}} that is CLOSE
to ``\texttt{\{original\_answer\}}'' (similar in scale, category, or
context) but clearly DIFFERENT from it.

Do NOT propose any of these already-tried alternatives:
\texttt{\{avoid\_list\}}.

Respond with ONLY a JSON object:

\texttt{\{"alternative\_entity": "<your proposed alternative>"\}}
\end{tcolorbox}

\vspace{-1mm}

% ---------- Prompt 3 ----------
\begin{tcolorbox}[
    title=\textbf{Counterfactual Context Generation: System Prompt},
    colback=white,
    colframe=gray!70,
    colbacktitle=gray!70,
    coltitle=white,
    fonttitle=\bfseries\footnotesize,
    boxrule=0.8pt,
    arc=2mm,
    left=3mm,
    right=3mm,
    top=2mm,
    bottom=2mm
]
\footnotesize
You are an expert at creating counterfactual reading-comprehension
passages for QA training data. Given an original context that supports a
specific answer, you rewrite it MINIMALLY so that it instead
unambiguously supports a different target answer, by replacing
occurrences of the original answer (and grammatically-dependent
references to it) with the target answer, while leaving everything else
in the passage unchanged. Do not add explanations or extra content.
\end{tcolorbox}

\vspace{-1mm}

% ---------- Prompt 4 ----------
\begin{tcolorbox}[
    title=\textbf{Counterfactual Context Generation: User Prompt},
    colback=white,
    colframe=gray!70,
    colbacktitle=gray!70,
    coltitle=white,
    fonttitle=\bfseries\footnotesize,
    boxrule=0.8pt,
    arc=2mm,
    left=3mm,
    right=3mm,
    top=2mm,
    bottom=2mm
]
\footnotesize
Question: \texttt{\{question\}}

\smallskip
Original context (supports answer
``\texttt{\{original\_answer\}}''):
\texttt{\{original\_context\}}

\smallskip
Target answer (the new counterfactual answer):
``\texttt{\{target\_answer\}}''

\smallskip
Rewrite the context above so that it minimally changes to unambiguously
support ``\texttt{\{target\_answer\}}'' as the answer to the question,
instead of ``\texttt{\{original\_answer\}}''. Replace occurrences of
``\texttt{\{original\_answer\}}'' (and any grammatically-dependent
references to it) with ``\texttt{\{target\_answer\}}'', keeping everything
else in the passage exactly the same. Do not add commentary.

\smallskip
Respond with ONLY a JSON object:

\texttt{\{"reasoning": "<1-2 sentences on what you changed>",}\\
\texttt{"counterfactual\_context": "<the full rewritten context>"\}}
\end{tcolorbox}

\caption{\textbf{Prompts used for counterfactual generation.}
The generation process consists of entity type classification, alternative
entity generation, and counterfactual context generation. The final stage
uses a system prompt and a user prompt to minimally modify the original
context to support the alternative answer.}
\label{fig:cf_generation_prompts}
\end{figure*}

Generated counterfactuals are subsequently evaluated using CFD
before being used for distillation.

\section{Proof of Proposition~\ref{prop:dev_bound}}
\label{app:proof}

In this section, we provide the proof of the upper bound on the Hausdorff distance between the teacher's context-response boundary $\mathcal{M}_t$ and the student's context-response boundary $\mathcal{M}_s$.

By definition, the Hausdorff distance between $\mathcal{M}_t$ and $\mathcal{M}_s$ is
\begin{equation}
    d_H(\mathcal{M}_t,\mathcal{M}_s)
    =
    \max
    \left\{
    \sup_{t\in\mathcal{M}_t}
    \inf_{s\in\mathcal{M}_s}
    \|t-s\|,
    \;
    \sup_{s\in\mathcal{M}_s}
    \inf_{t\in\mathcal{M}_t}
    \|s-t\|
    \right\}.
\end{equation}
We bound the two directed distances separately~\citep{hamman2025few}.

\subsection*{Part 1: Bounding the deviation from $\mathcal{M}_t$ to $\mathcal{M}_s$}

\begin{wrapfigure}{l}{0.28\textwidth}
    \centering
    \vspace{-30pt}
    \includegraphics[width=0.28\textwidth]{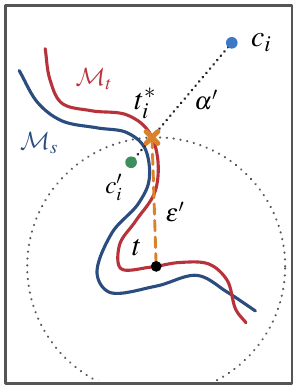}
    \caption{Visualization of the distances between $\mathcal{M}_t$ and $\mathcal{M}_s$. $c_i$ and $c'_i$ are factual and counterfactual pairs with $\|c_i - c_i'\| = \alpha' \le \alpha$). $t$ is an arbitrary point in $\mathcal{M}_t$ that finds some $t_i^*$ cut point in $\mathcal{M}_t$ within $\epsilon'$ distance ($\epsilon' \le \epsilon$).}
    \label{fig:tube}
\end{wrapfigure}

Let $t\in\mathcal{M}_t$ be an arbitrary point on the teacher's boundary. By the boundary-coverage assumption (Assumption~3), there exists a teacher boundary crossing point $t_i^*\in\mathcal{M}_t$ associated with a factual-counterfactual pair $(c_i,c_i')$ such that (see  Figure~\ref{fig:tube}):

\begin{equation}
    \|t-t_i^*\|\leq\epsilon.
    \label{eq:boundary_coverage}
\end{equation}

Under Assumption~1, the factual-counterfactual perturbation is bounded as
\begin{equation}
    \|c_i-c_i'\|\leq\alpha.
\end{equation}
Under the teacher-student agreement assumption (Assumption~2), the teacher and student agree on the responses associated with both endpoints $c_i$ and $c_i'$. Since the factual and counterfactual contexts correspond to different responses, both the teacher and student boundaries cross the segment $[c_i,c_i']$. Let $t_i^*\in\mathcal{M}_t$ and $s_i^*\in\mathcal{M}_s$ denote their respective boundary crossing points on this segment.

Because both $t_i^*$ and $s_i^*$ lie on the same segment $[c_i,c_i']$,
\begin{equation}
    \|t_i^*-s_i^*\|
    \leq
    \|c_i-c_i'\|
    \leq
    \alpha.
    \label{eq:crossing_distance}
\end{equation}

Combining Equations~\eqref{eq:boundary_coverage} and~\eqref{eq:crossing_distance} and applying the triangle inequality, we obtain
\begin{equation}
    \|t-s_i^*\|
    \leq
    \|t-t_i^*\|+\|t_i^*-s_i^*\|
    \leq
    \epsilon+\alpha.
\end{equation}

Since there exists at least one point $s_i^* \in \mathcal{M}_s$ within distance $\epsilon + \alpha$ of $t$, the infimum distance satisfies:
\begin{equation}
    \inf_{s\in\mathcal{M}_s}\|t-s\|
    \leq
    \epsilon+\alpha.
\end{equation}
Taking the supremum over all points $t \in \mathcal{M}_t$ yields:
\begin{equation}
    \sup_{t\in\mathcal{M}_t}
    \inf_{s\in\mathcal{M}_s}
    \|t-s\|
    \leq
    \epsilon+\alpha.
    \label{eq:teacher_to_student}
\end{equation}

\subsection*{Part 2: Bounding the deviation from $\mathcal{M}_s$ to $\mathcal{M}_t$}

We follow a symmetric logic for the student boundary. For any point $s \in \mathcal{M}_s$, Assumption 3 guarantees the existence of a student boundary cutpoint $s_i^*$ such that $\|s - s_i^*\| \leq \epsilon$. 

From the relationship established in Part 1, the distance between this student cutpoint and the corresponding teacher cutpoint $t_i^*$ is $\|s_i^* - t_i^*\| \leq \alpha$. Applying the triangle inequality:
\begin{equation}
    \|s - t_i^*\| \leq \|s - s_i^*\| + \|s_i^* - t_i^*\| \leq \epsilon + \alpha
\end{equation}

Since $t_i^* \in \mathcal{M}_t$, it follows that $\inf_{t \in \mathcal{M}_t} \|s - t\| \leq \epsilon + \alpha$. Taking the supremum over all $s \in \mathcal{M}_s$:
\begin{equation}
    \sup_{s \in \mathcal{M}_s} \inf_{t \in \mathcal{M}_t} \|s - t\| \leq \epsilon + \alpha
    \label{eq:student_to_teacher}
\end{equation}

\subsection*{Final Result}

Combining Equations~\eqref{eq:teacher_to_student} and~\eqref{eq:student_to_teacher}, we obtain
\begin{equation}
    d_H(\mathcal{M}_t,\mathcal{M}_s)
    \leq
    \max\{\epsilon+\alpha,\epsilon+\alpha\}
    =
    \alpha+\epsilon.
\end{equation}

This confirms that the student boundary remains within an $(\alpha + \epsilon)$-tube of the teacher boundary.

\section{Experimental Setup Details} 
\label{app:setup}

\subsection{Datasets} \label{app:benchmarks}

We evaluate \ours on eight open-domain question answering benchmarks, comprising five in-domain and three out-of-domain datasets. The in-domain benchmarks include Natural Questions (NQ), SQuAD, NewsQA, TriviaQA, and HotpotQA. NQ consists of real user queries collected from Google search logs~\citep{kwiatkowski2019natural}. SQuAD (Stanford Question Answering Dataset) contains questions authored by crowdworkers based on Wikipedia articles~\citep{rajpurkar2016squad}. NewsQA is constructed from CNN news articles with questions generated crowdworkers~\citep{trischler2017newsqa}. TriviaQA consists QA pairs sourced from trivia and quiz-league websites~\citep{joshi2017triviaqa}. HotpotQA focuses on multi-hop reasoning, requiring models to combine information from multiple Wikipedia articles~\citep{yang2018hotpotqa}. The out-of-domain benchmarks include DROP, DuoRC, and RelationExtraction. DROP contains quantitative QA pairs which has numerical and text answers~\citep{dua2019drop}. DuoRC contains QA pairs derived from movie plots, constructed by crowdworkers using IMDb and Wikipedia~\citep{saha2018duorc}. RelationExtraction reformulates relation extraction as a reading comprehension task using Wikipedia text~\citep{levy2017zero}. 

The datasets are designed for generalization in open-book question answering, where samples are drawn from diverse distributions. The datasets have following properties in common.

\begin{itemize}
    \item \textbf{Question distribution:} The test sets contain different styles of questions, such as entity-based and relational questions, collected from diverse sources, including crowdworkers, domain experts, and exam writers.
    \item \textbf{Context distribution:} Passages are drawn from heterogeneous sources such as science articles, novels, medical abstracts, news articles.
    \item \textbf{Answerability:} The answer can be generated using the provided context, and the answer span appears within the first 800 tokens of the passage.
\end{itemize}

The training and validation sets for the teacher are drawn from the original training splits after removing duplicate entries. Dataset statistics are reported in Table~\ref{tab:data_stats}. We generate counterfactuals only for the samples with entity-based answers. Statistics of the knowledge distillation datasets are also shown in Table~\ref{tab:data_stats}, where student \textit{train} and \textit{val} denote available factual training samples with entity-based answers for distillation. The in-domain test sets for knowledge distillation are taken from the original development splits after filtering for entity-based answer samples. The out-of-domain test set counts are as follows: DROP (496), and DuoRC (450), RelationExtraction (912).

\begin{table}[t]
\caption{Statistics of teacher and student datasets used in our experiments. Teacher datasets contain the original training/validation splits, while student datasets contain available factual instances for distillation and evaluation.}
\centering
\small
\setlength{\tabcolsep}{5pt}
\renewcommand{\arraystretch}{1.1}

\begin{tabular}{lccccc}
\toprule
& \multicolumn{2}{c}{Teacher Dataset} & \multicolumn{3}{c}{Student Dataset} \\
\cmidrule(lr){2-3} \cmidrule(lr){4-6}
Dataset & Train & Val. & Train & Val. & Test \\
\midrule
NQ        & 93,663 & 10,408 & 34,383 & 3,821 & 2,283 \\
SQuAD     & 77,847 & 8,651  & 16,341 & 1,816 & 2,790 \\
NewsQA    & 66,655 & 7,407  & 10,956 & 1,218 & 690   \\
TriviaQA  & 55,519 & 6,170  & 4,415  & 491   & 607   \\
HotpotQA  & 65,630 & 7,294  & 28,956 & 3,218 & 2,052 \\
\bottomrule
\end{tabular}

\label{tab:data_stats}
\end{table}

\subsection{Teacher Model Fine-Tuning and Evaluation} \label{app:finetuning}

All fine-tuning experiments were conducted on a machine equipped with four NVIDIA RTX 6000 Ada GPUs. For T5-Large fine-tuning experiments, we used a batch size of 4, a learning rate of $1 \times 10^{-4}$, and a maximum of 5 training epochs. Training was stopped early based on validation loss using an early stopping criterion with a patience of 3 validation checks. For Qwen2.5-7B-IT fine-tuning experiments, we used a batch size of 1, a learning rate of $2 \times 10^{-5}$, and a maximum of 5 training epochs. Training was stopped early based on validation loss using an early stopping criterion with a patience of 3 validation checks. On average, all models converged within 1 to 2 epochs.

We evaluated the fine-tuned teacher models using the same test sets for each dataset. The results can be found in Table~\ref{tab:teacher_accuracy}

\begin{table}[t]
\caption{Teacher model performance (EM and F1) on factual and counterfactual test sets across five in-domain QA benchmarks for T5 and Qwen2.5 models.}
\label{tab:teacher_accuracy}
\centering
\scriptsize
\renewcommand{\arraystretch}{0.95}
\setlength{\tabcolsep}{6pt}

\begin{tabular}{lllccccc}
\toprule
Test Set Type & Model & Score & NQ & SQuAD & NewsQA & TriviaQA & HotpotQA \\
\midrule

\multirow{4}{*}{Factual} 
& \multirow{2}{*}{T5} & EM & 74.90 & 79.64 & 73.19 & 67.38 & 77.19 \\
& & F1 & 77.76 & 87.78 & 78.97 & 69.27 & 85.49 \\
\cmidrule(lr){2-8}
& \multirow{2}{*}{Qwen2.5} & EM & 79.46 & 83.91 & 75.65 & 73.48 & 79.34 \\
& & F1 & 82.83 & 91.04 & 82.50 & 76.37 & 88.52 \\
\midrule

\multirow{4}{*}{Counterfactual} 
& \multirow{2}{*}{T5} & EM & 67.94 & 79.75 & 66.38 & 58.32 & 65.84 \\
& & F1 & 70.79 & 87.04 & 71.35 & 60.58 & 71.37 \\
\cmidrule(lr){2-8}
& \multirow{2}{*}{Qwen2.5} & EM & 69.78 & 79.35 & 66.23 & 31.47 & 64.13 \\
& & F1 & 73.61 & 86.46 & 70.55 & 32.22 & 70.74 \\
\bottomrule
\end{tabular}
\end{table}

\subsection{Additional details of the baselines} \label{app:baselines}

We compare \ours against three knowledge distillation baselines. To ensure a fair comparison, all methods use the same $k$-shot data splits, and evaluation settings.

\begin{itemize}
    \item \textbf{Standard Knowledge Distillation (KD).}
Standard knowledge distillation (KD)~\citep{hinton2015distilling} transfers
knowledge from a larger teacher model to a smaller student model by aligning
their output distributions. Let $x$ denote the input, $y=(y_1,\dots,y_T)$
the ground-truth output sequence, and $p_s$ and $p_t$ the student and teacher
model distributions, respectively. The student is first trained using the
standard cross-entropy objective
\begin{equation}
    \mathcal{L}_{\mathrm{CE}}
    =
    -\sum_{t=1}^{T}
    \log p_s(y_t \mid x,y_{<t}).
\end{equation}

In addition, KD encourages the student to match the teacher's token-level
output distribution at each decoding step using
\begin{equation}
    \mathcal{L}_{\mathrm{KD}}
    =
    \sum_{t=1}^{T}
    \mathrm{KL}\left(
        p_t(\cdot \mid x,y_{<t})
        \,\|\, 
        p_s(\cdot \mid x,y_{<t})
    \right).
\end{equation}

The final training objective combines the supervised and distillation losses:
\begin{equation}
    \mathcal{L}
    =
    \alpha\mathcal{L}_{\mathrm{CE}}
    +
    (1-\alpha)\mathcal{L}_{\mathrm{KD}},
\end{equation}
where $\alpha\in[0,1]$ controls the trade-off between fitting the ground-truth
sequence and matching the teacher's output distribution. Thus, standard KD
transfers knowledge by aligning the teacher and student token-level
distributions while retaining supervision from the ground-truth sequence.

\item \textbf{Layer-wise Distillation (LWD).}
    Layer-wise distillation (LWD)~\citep{jiao2020tinybert} extends output-level knowledge distillation by additionally transferring information from the intermediate representations of the teacher. Representations from selected teacher layers are aligned with the corresponding representations of the student through an auxiliary layer-wise distillation objective, providing supervision at both the intermediate representation and output levels. Let $\mathbf{h}_t^{(l_t)}$ and $\mathbf{h}_s^{(l_s)}$ denote the hidden representations from a paired teacher layer $l_t$ and student layer $l_s$, respectively. Since the teacher and student may have different hidden dimensions, a learnable projection $W_{l_s}$ is applied to the student representation before alignment. The layer-wise distillation loss is defined as
    \begin{equation}
        \mathcal{L}_{\mathrm{LWD}}
        =
        \frac{1}{|\mathcal{P}|}
        \sum_{(l_s,l_t)\in\mathcal{P}}
        \left\|
        W_{l_s}\mathbf{h}_s^{(l_s)}
        -
        \mathbf{h}_t^{(l_t)}
        \right\|_2^2,
    \end{equation}
    where $\mathcal{P}$ denotes the set of paired student and teacher layers. The final objective combines the standard KD objective with the layer-wise representation loss:
    \begin{equation}
        \mathcal{L}
        =
        \alpha \mathcal{L}_{\mathrm{CE}}
        +
        (1-\alpha)\mathcal{L}_{\mathrm{KD}}
        +
        \beta\mathcal{L}_{\mathrm{LWD}},
    \end{equation}
    where $\beta$ controls the contribution of intermediate representation alignment. Thus, unlike standard KD, which transfers knowledge through the output distributions, LWD additionally encourages the student to reproduce the teacher's intermediate representations.

    \item \textbf{Sequence-Level Knowledge Distillation (SeqKD).}
    Unlike standard KD, which aligns the teacher and student token-level distributions at each decoding step, sequence-level knowledge distillation (seqKD) transfers knowledge through complete sequences generated by the teacher~\cite{kim2016sequence}. Let $x$ denote the input and $y=(y_1,\dots,y_T)$ denote the output sequence, and let $p_s$ and $p_t$ denote the student and teacher model distributions, respectively. The sequence-level distillation objective is defined as
    
    \begin{equation}
        \mathcal{L}_{\mathrm{SeqKD}}
        =
        -\sum_{y \in \mathcal{Y}}
        p_t(y \mid x)
        \log p_s(y \mid x),
    \end{equation}
    
    where $\mathcal{Y}$ denotes the set of all possible output sequences. Since directly optimizing over all possible output sequences is intractable, seqKD approximates the teacher's sequence-level distribution using its highest-probability output sequence. Specifically, the teacher-generated sequence is obtained as
    
    \begin{equation}
        \hat{y}
        \approx
        \arg\max_{y \in \mathcal{Y}}
        p_t(y \mid x),
    \end{equation}
    
    where beam search is used to approximate the highest probability sequence under the teacher distribution. The resulting sequence-level distillation objective is then
    
    \begin{equation}
        \mathcal{L}_{\mathrm{SeqKD}}
        \approx
        -\log p_s(\hat{y} \mid x)
        =
        -\sum_{t=1}^{|\hat{y}|}
        \log p_s
        \left(
            \hat{y}_t \mid x,\hat{y}_{<t}
        \right).
    \end{equation}

    Thus, while standard KD transfers knowledge by matching the teacher's token-level probability distributions, seqKD uses the complete sequence generated by the teacher as the training target for the student.  
    
\end{itemize}

For standard KD, the student was trained using a weighted combination of
the MLE loss and the token-level distillation loss. The temperature was
set to 1, and the $\alpha$ was set to
$0.5$.

For LWD, student decoder layers were uniformly mapped to corresponding
teacher decoder layers across the model depth. Layer-wise knowledge
transfer was performed by aligning the decoder self-attention
distributions of the mapped teacher and student layers using KL
divergence. Attention distributions were averaged across attention heads
before computing the layer-wise loss. The layer-wise distillation loss
was added to the standard KD objective with a weight of $5.0$.

For seqKD, the teacher generated a complete output sequence for each
training input using beam search with a beam size of 4. The original
ground-truth answer was replaced by the corresponding teacher-generated
sequence, which was then used as the training target for the student.
Following the original seqKD formulation~\citep{kim2016sequence}, the
student was trained using cross-entropy loss on the teacher-generated
sequences without an additional token-level distillation loss.

For all three baselines, we used a batch size of 4 and a learning rate of
$1 \times 10^{-4}$. The maximum number of training epochs was set to 100,
with early stopping based on validation loss using a patience of 3
validation checks.

\newpage

\section{Ablation Study Results} \label{app:ablation}

\subsection{Effect of CFD Validation} \label{app:nocfd}

\begin{figure*}[ht]
    \centering

    \begin{minipage}[t]{0.48\textwidth}
        \centering
        \includegraphics[width=\linewidth]{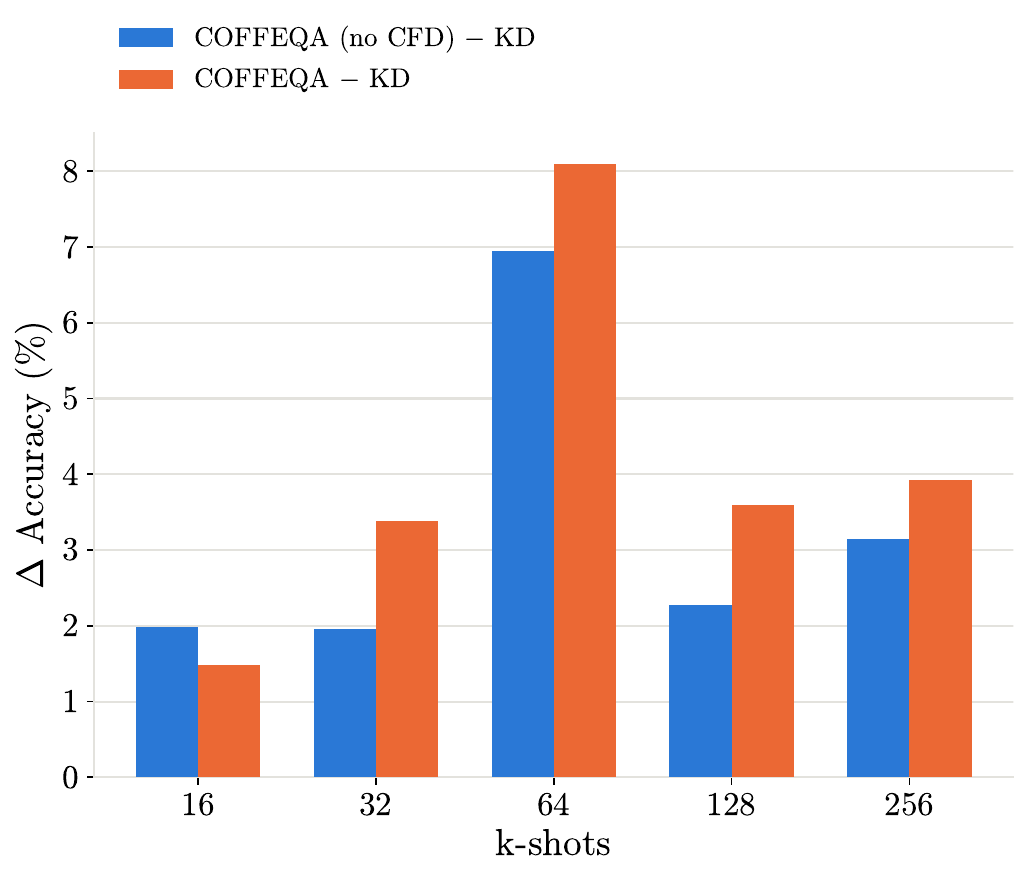}
        \vspace{-0.5em}
        \small (a) Accuracy EM
    \end{minipage}
    \hfill
    \begin{minipage}[t]{0.48\textwidth}
        \centering
        \includegraphics[width=\linewidth]{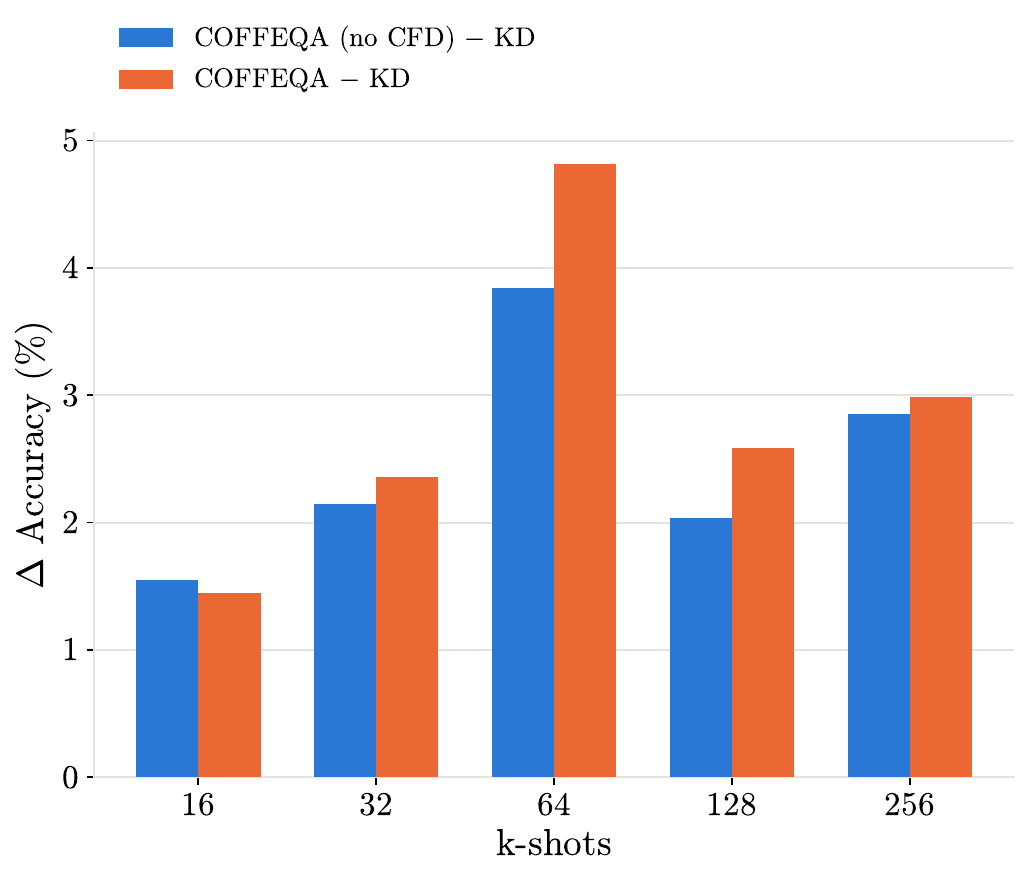}
        \vspace{-0.5em}
        \small (b) Accuracy F1
    \end{minipage}

    \vspace{0.5em}

    \begin{minipage}[t]{0.48\textwidth}
        \centering
        \includegraphics[width=\linewidth]{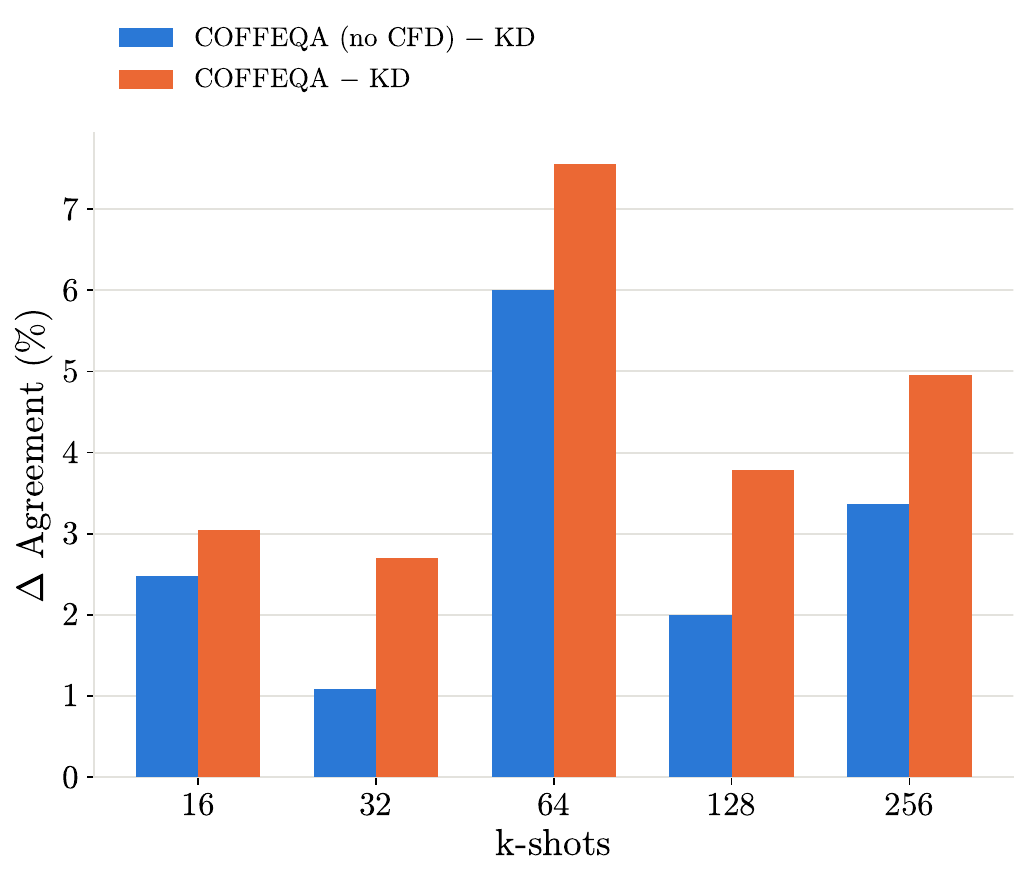}
        \vspace{-0.5em}
        \small (c) Agreement EM
    \end{minipage}
    \hfill
    \begin{minipage}[t]{0.48\textwidth}
        \centering
        \includegraphics[width=\linewidth]{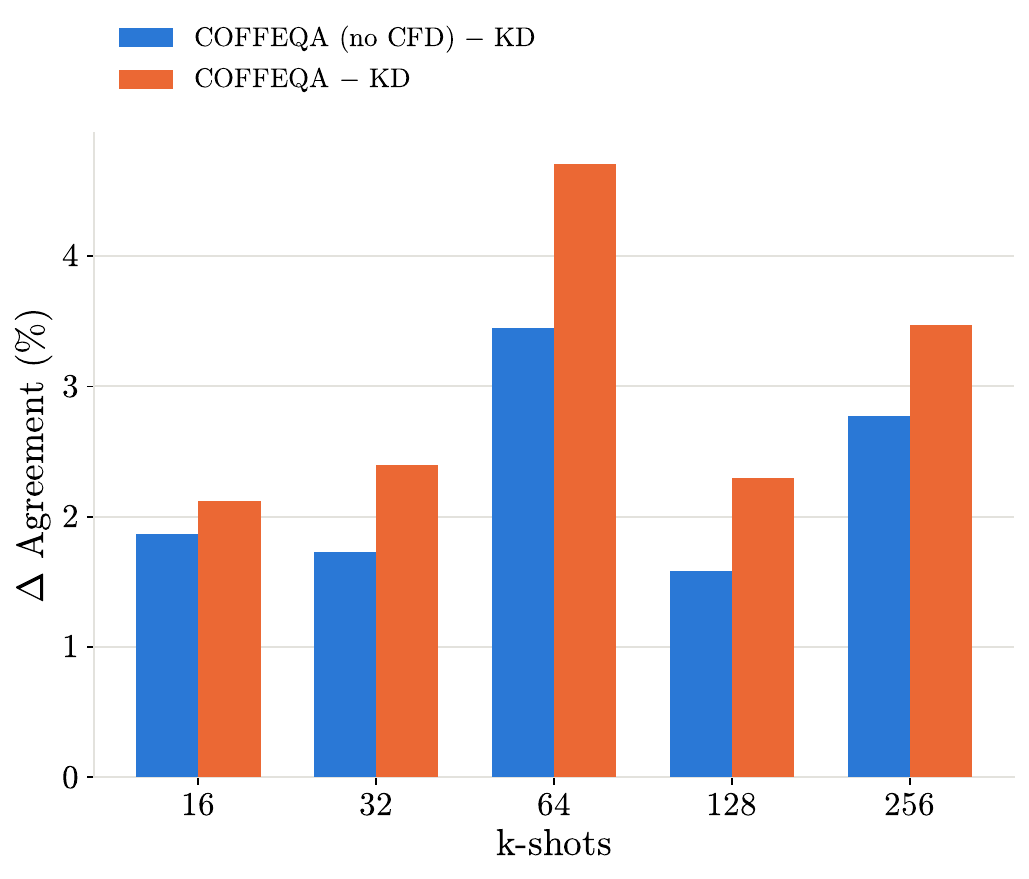}
        \vspace{-0.5em}
        \small (d) Agreement F1
    \end{minipage}
    \vspace{1em}

    \caption{\textbf{Effect of CFD verification on \ours.}
    Performance gains over standard KD ($\Delta=\mathrm{\ours}-\mathrm{KD}$) are shown across different few-shot budgets $k$. We compare \ours without CFD verification against the full \ours framework with CFD verification ($\tau=0.3$). Panels (a) and (b) report accuracy EM and F1 gains, while panels (c) and (d) report teacher-student agreement EM and F1 gains. CFD verification generally yields larger improvements, demonstrating the benefit of validating generated counterfactuals before distillation.}
    \label{fig:cfd-verification}
\end{figure*}

To isolate the contribution of CFD verification, we compare \ours with a variant that generates counterfactuals without CFD verification. Both variants use identical factual data splits and training settings, with the only difference being whether the generated counterfactuals are verified using CFD. For \ours with CFD verification, we use the selected threshold $\tau=0.3$. Experiments are conducted using T5 across all datasets with five random seeds for each $k$-shot setting, and the results are averaged across datasets. Figure~\ref{fig:cfd-verification} reports the improvement of each variant over standard KD. CFD verification generally provides larger gains in both downstream QA performance and teacher-student agreement, with particularly clear improvements at $k=32$, $64$, and $128$. These results indicate that the improvements of \ours are not solely due to the inclusion of counterfactual examples; verifying the generated counterfactuals using CFD provides additional benefits for knowledge transfer.

\subsection{CFD Threshold Ablation} \label{app:cfdthresh}

\begin{figure}[ht]
    \centering
    \begin{minipage}[t]{0.49\linewidth}
        \centering
        \includegraphics[width=\linewidth]{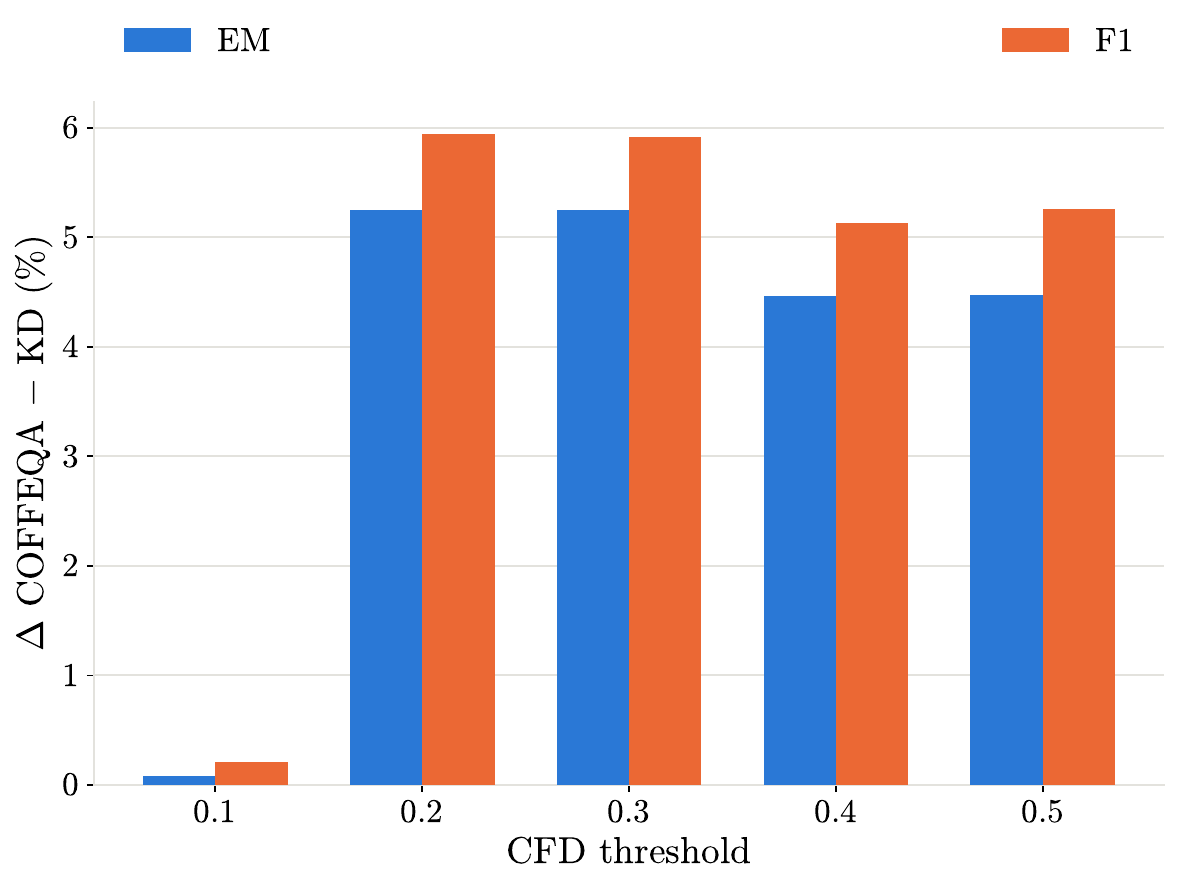}
    \end{minipage}
    \hfill
    \begin{minipage}[t]{0.49\linewidth}
        \centering
        \includegraphics[width=\linewidth]{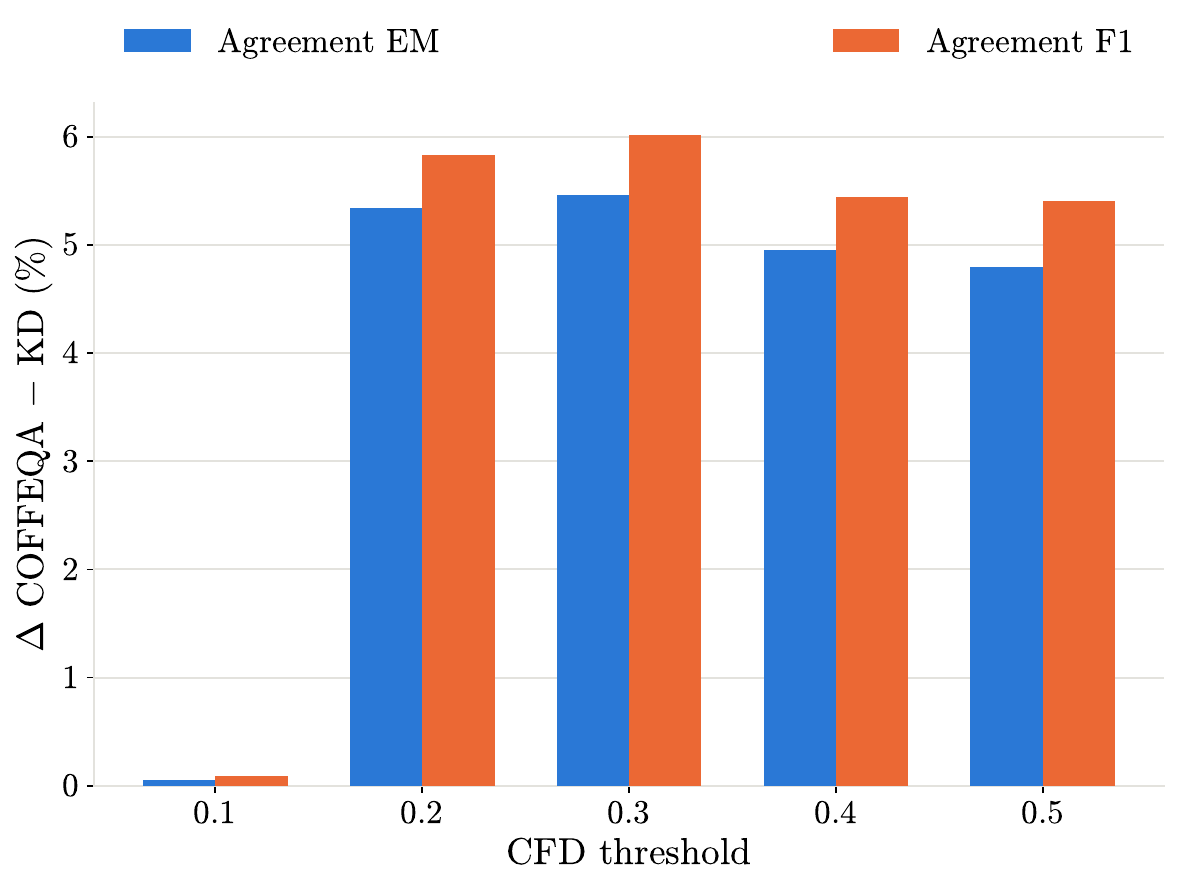}
    \end{minipage}
    \caption{\textbf{CFD threshold ablation.} Effect of the CFD threshold $\tau$ on \ours using T5 on TriviaQA. The bars show the improvement of \ours over standard KD ($\Delta=\mathrm{\ours}-\mathrm{KD}$), averaged across $k \in \{16,32,64,128,256\}$ and five random seeds. The left plot shows the improvement in accuracy measured by EM and F1, while the right plot shows the improvement in teacher-student agreement measured by EM and F1.}
    \label{fig:cfd-threshold-ablation}
\end{figure}

We analyze the sensitivity of \ours to the CFD threshold $\tau$ using T5 on TriviaQA. We evaluate $\tau \in \{0.1,0.2,0.3,0.4,0.5\}$ across $k$, with five random seeds for each $k$-shot setting. For each threshold, we report the averaged improvement of \ours over standard KD ($\Delta=\mathrm{\ours}-\mathrm{KD}$). As shown in Figure~\ref{fig:cfd-threshold-ablation}, performance is sensitive to the CFD threshold. The threshold $\tau=0.1$ yields negligible improvements, whereas moderate thresholds produce substantially larger gains in both accuracy EM/F1 and teacher-student agreement. The gains peak around $\tau=0.2$--$0.3$ and gradually decrease as the threshold becomes more permissive. Overall, these results demonstrate the importance of controlling counterfactual quality through the CFD threshold.

\subsection{Cost Analysis}
\label{app:cost}
We analyze the computational cost of \ours during both counterfactual generation and distillation. Table~\ref{tab:generation_cost} reports the per-sample cost of counterfactual generation, averaged over all generation attempts. Counterfactual generation requires approximately $2.88$ seconds per candidate, with relatively small CPU and GPU energy consumption for both model settings.

Table~\ref{tab:compute_energy_full} and Table~\ref{tab:compute_energy_full_qwen} reports the training duration and energy consumption during distillation under T5 and Qwen2.5, respectively, measured using CodeCarbon. In both models, for standard KD and LWD, \ours achieves higher accuracy while requiring less training time and energy across all $k$-shot settings. For SeqKD, \ours also improves accuracy across all settings, but incurs higher computational costs at the smaller budgets ($k=16$ and $32$). From $k=64$ onward, its computational cost becomes comparable to the baseline. Overall, these results show that \ours introduces an additional counterfactual-generation cost, while maintaining favorable computational efficiency during the subsequent distillation stage.

\begin{table}[ht]
\centering
\caption{Average duration and CPU/GPU energy consumption per counterfactual generation candidate under the T5 and Qwen2.5 settings. Results are averaged across all generation attempts.}
\label{tab:generation_cost}
\begin{tabular}{l r r r}
\toprule
\textbf{Model} & \textbf{Duration/candidate (s)} & \textbf{CPU/candidate (kWh)} & \textbf{GPU/candidate (kWh)} \\
\midrule
T5 & 2.884 & 0.000083 & 0.000055 \\
Qwen2.5 & 2.877 & 0.000071 & 0.000067 \\
\bottomrule
\end{tabular}
\end{table}

\begin{table}[htbp]
\centering
\caption{Comparison of EM score, distillation duration, and CPU/GPU energy consumption between baseline methods and COFFEQA across different few-shot budgets under the T5 setting.}
\label{tab:compute_energy_full}
\begin{tabular}{l l r r r r r}
\toprule
\textbf{Objective} & \textbf{Method} & \textbf{$k$} & \textbf{EM (\%)} & \textbf{Duration (s)} & \textbf{CPU (kWh)} & \textbf{GPU (kWh)} \\
\midrule
\multirow{10}{*}{KD} & \multirow{5}{*}{Baseline} & 16 & 61.88 & 62.86 & 0.00201 & 0.01342 \\
 &  & 32 & 60.82 & 50.34 & 0.00153 & 0.00980 \\
 &  & 64 & 54.83 & 56.15 & 0.00170 & 0.01259 \\
 &  & 128 & 56.07 & 75.90 & 0.00224 & 0.01662 \\
 &  & 256 & 55.56 & 128.67 & 0.00378 & 0.02777 \\
\cmidrule(lr){2-7}
 & \multirow{5}{*}{\ours} & 16 & 63.36 & 45.19 & 0.00143 & 0.01016 \\
 &  & 32 & 64.20 & 42.58 & 0.00133 & 0.00879 \\
 &  & 64 & 62.93 & 49.46 & 0.00149 & 0.01027 \\
 &  & 128 & 59.67 & 68.09 & 0.00203 & 0.01428 \\
 &  & 256 & 59.49 & 114.55 & 0.00336 & 0.02487 \\
\midrule
\multirow{10}{*}{LWD} & \multirow{5}{*}{Baseline} & 16 & 61.75 & 150.44 & 0.00474 & 0.02235 \\
 &  & 32 & 60.94 & 127.87 & 0.00405 & 0.01862 \\
 &  & 64 & 54.52 & 137.14 & 0.00425 & 0.02128 \\
 &  & 128 & 55.49 & 161.15 & 0.00488 & 0.02327 \\
 &  & 256 & 55.88 & 221.42 & 0.00694 & 0.03796 \\
\cmidrule(lr){2-7}
 & \multirow{5}{*}{\ours} & 16 & 62.94 & 121.86 & 0.00383 & 0.01784 \\
 &  & 32 & 64.38 & 109.26 & 0.00344 & 0.01799 \\
 &  & 64 & 62.72 & 127.24 & 0.00389 & 0.01743 \\
 &  & 128 & 59.94 & 131.83 & 0.00400 & 0.01907 \\
 &  & 256 & 59.78 & 180.71 & 0.00569 & 0.03242 \\
\midrule
\multirow{10}{*}{SeqKD} & \multirow{5}{*}{Baseline} & 16 & 61.77 & 59.14 & 0.00186 & 0.01021 \\
 &  & 32 & 61.19 & 39.74 & 0.00127 & 0.00801 \\
 &  & 64 & 61.37 & 36.16 & 0.00115 & 0.00774 \\
 &  & 128 & 61.15 & 44.13 & 0.00136 & 0.00988 \\
 &  & 256 & 62.06 & 71.46 & 0.00215 & 0.01550 \\
\cmidrule(lr){2-7}
 & \multirow{5}{*}{\ours} & 16 & 64.81 & 91.77 & 0.00287 & 0.01409 \\
 &  & 32 & 64.70 & 53.12 & 0.00169 & 0.01120 \\
 &  & 64 & 64.50 & 37.99 & 0.00121 & 0.00832 \\
 &  & 128 & 64.54 & 47.76 & 0.00149 & 0.00999 \\
 &  & 256 & 64.06 & 82.79 & 0.00248 & 0.01780 \\
\bottomrule
\end{tabular}
\end{table}

\begin{table}[ht]
\centering
\caption{Comparison of EM score, distillation duration, and CPU/GPU energy consumption between baseline methods and COFFEQA across different few-shot budgets under the Qwen2.5.}
\label{tab:compute_energy_full_qwen}
\begin{tabular}{l l r r r r r}
\toprule
\textbf{Objective} & \textbf{Method} & \textbf{$k$} & \textbf{EM (\%)} & \textbf{Duration (s)} & \textbf{CPU (kWh)} & \textbf{GPU (kWh)} \\
\midrule
\multirow{10}{*}{KD} & \multirow{5}{*}{Baseline} & 16 & 61.22 & 748.41 & 0.02378 & 0.11453 \\
 &  & 32 & 60.24 & 636.82 & 0.01930 & 0.11645 \\
 &  & 64 & 59.13 & 774.16 & 0.02278 & 0.15832 \\
 &  & 128 & 58.51 & 1202.28 & 0.03660 & 0.27748 \\
 &  & 256 & 59.07 & 1793.38 & 0.05119 & 0.35929 \\
\cmidrule(lr){2-7}
 & \multirow{5}{*}{\ours} & 16 & 61.73 & 704.12 & 0.02244 & 0.11386 \\
 &  & 32 & 61.59 & 698.58 & 0.02124 & 0.13367 \\
 &  & 64 & 61.78 & 693.77 & 0.02043 & 0.14167 \\
 &  & 128 & 60.93 & 1066.53 & 0.03231 & 0.24643 \\
 &  & 256 & 60.59 & 1728.09 & 0.04949 & 0.33793 \\
\midrule
\multirow{10}{*}{LWD} & \multirow{5}{*}{Baseline} & 16 & 60.77 & 892.88 & 0.02964 & 0.14427 \\
 &  & 32 & 60.94 & 614.12 & 0.01998 & 0.11107 \\
 &  & 64 & 59.71 & 560.93 & 0.01830 & 0.10488 \\
 &  & 128 & 59.52 & 641.79 & 0.02027 & 0.11110 \\
 &  & 256 & 59.70 & 840.55 & 0.02688 & 0.16956 \\
\cmidrule(lr){2-7}
 & \multirow{5}{*}{\ours} & 16 & 61.10 & 852.61 & 0.02920 & 0.13976 \\
 &  & 32 & 61.51 & 727.77 & 0.02350 & 0.12773 \\
 &  & 64 & 61.49 & 477.78 & 0.01546 & 0.08719 \\
 &  & 128 & 61.80 & 526.55 & 0.01641 & 0.09615 \\
 &  & 256 & 62.03 & 759.89 & 0.02396 & 0.14949 \\
\midrule
\multirow{10}{*}{SeqKD} & \multirow{5}{*}{Baseline} & 16 & 59.72 & 584.15 & 0.01826 & 0.11342 \\
 &  & 32 & 60.46 & 559.89 & 0.01701 & 0.10684 \\
 &  & 64 & 60.96 & 642.25 & 0.01894 & 0.14049 \\
 &  & 128 & 60.79 & 970.14 & 0.02825 & 0.21411 \\
 &  & 256 & 62.23 & 1683.50 & 0.04842 & 0.38788 \\
\cmidrule(lr){2-7}
 & \multirow{5}{*}{\ours} & 16 & 59.98 & 648.94 & 0.01986 & 0.12576 \\
 &  & 32 & 60.72 & 557.76 & 0.01663 & 0.11214 \\
 &  & 64 & 62.16 & 641.84 & 0.01882 & 0.13954 \\
 &  & 128 & 62.21 & 925.38 & 0.02703 & 0.18591 \\
 &  & 256 & 62.88 & 1527.88 & 0.04422 & 0.34185 \\
\bottomrule
\end{tabular}
\end{table}

\newpage

\section{Additional Experimental Results}
\label{app:exp} 

\subsection{In-Domain EM Results} \label{app:emindomain}

Table~\ref{tab:indomain_em_all_baselines} reports the in-domain EM results across all the baselines.

\begin{table*}[htbp]
\caption{In-Domain EM score comparison across $k$-shots for T5 and Qwen2.5 models using baselines. Mean EM scores are reported with standard deviations as subscripts ($Mean_{std}$). Bold values indicate the better-performing method (\ours vs. baseline). Blue shading highlights aggregate results.}
\label{tab:indomain_em_all_baselines}
\centering
\fontsize{6.5pt}{8.5pt}\selectfont
\setlength{\tabcolsep}{1.0pt}
\renewcommand{\arraystretch}{0.95}

\definecolor{tableblue}{HTML}{E6F0FF}

\resizebox{\textwidth}{!}{%
\begin{tabular}{llcccccccccccccc}
\toprule
 & & \multicolumn{6}{c}{\textbf{T5}} & \multicolumn{6}{c}{\textbf{Qwen2.5}} \\
\cmidrule(lr){3-8} \cmidrule(lr){9-14}
Dataset & Method & 16 & 32 & 64 & 128 & 256 & \cellcolor{tableblue}Avg & 16 & 32 & 64 & 128 & 256 & \cellcolor{tableblue}Avg \\
\midrule

\multirow{6}{*}{NQ} 
& KD & 58.10$_{\scalebox{0.6}{0.44}}$ & 57.73$_{\scalebox{0.6}{3.52}}$ & 54.73$_{\scalebox{0.6}{3.79}}$ & 47.84$_{\scalebox{0.6}{6.31}}$ & 50.32$_{\scalebox{0.6}{3.61}}$ & \cellcolor{tableblue}53.75$_{\scalebox{0.6}{4.54}}$ & \textbf{61.95}$_{\scalebox{0.6}{0.36}}$ & 61.24$_{\scalebox{0.6}{0.61}}$ & 60.89$_{\scalebox{0.6}{0.44}}$ & 60.18$_{\scalebox{0.6}{0.53}}$ & 62.07$_{\scalebox{0.6}{0.47}}$ & \cellcolor{tableblue}61.27$_{\scalebox{0.6}{0.78}}$ \\
& + \ours & \textbf{58.89}$_{\scalebox{0.6}{0.33}}$ & \textbf{58.49}$_{\scalebox{0.6}{1.50}}$ & \textbf{56.92}$_{\scalebox{0.6}{2.34}}$ & \textbf{49.86}$_{\scalebox{0.6}{6.34}}$ & \textbf{51.20}$_{\scalebox{0.6}{5.30}}$ & \cellcolor{tableblue}\textbf{55.07}$_{\scalebox{0.6}{4.24}}$ & 61.79$_{\scalebox{0.6}{0.27}}$ & \textbf{63.12}$_{\scalebox{0.6}{0.35}}$ & \textbf{63.06}$_{\scalebox{0.6}{0.47}}$ & \textbf{62.66}$_{\scalebox{0.6}{0.82}}$ & \textbf{62.84}$_{\scalebox{0.6}{1.32}}$ & \cellcolor{tableblue}\textbf{62.69}$_{\scalebox{0.6}{0.54}}$ \\
\cmidrule(lr){2-14}
& LWD & 58.20$_{\scalebox{0.6}{0.48}}$ & \textbf{57.47}$_{\scalebox{0.6}{3.89}}$ & 54.80$_{\scalebox{0.6}{3.04}}$ & 50.15$_{\scalebox{0.6}{3.73}}$ & 50.90$_{\scalebox{0.6}{5.43}}$ & \cellcolor{tableblue}54.30$_{\scalebox{0.6}{3.68}}$ & \textbf{60.19}$_{\scalebox{0.6}{6.36}}$ & 61.91$_{\scalebox{0.6}{0.79}}$ & 61.10$_{\scalebox{0.6}{1.33}}$ & 60.71$_{\scalebox{0.6}{0.44}}$ & 61.82$_{\scalebox{0.6}{1.03}}$ & \cellcolor{tableblue}61.15$_{\scalebox{0.6}{0.73}}$ \\
& + \ours & \textbf{58.84}$_{\scalebox{0.6}{0.27}}$ & 57.45$_{\scalebox{0.6}{3.65}}$ & \textbf{55.00}$_{\scalebox{0.6}{4.16}}$ & \textbf{52.58}$_{\scalebox{0.6}{4.70}}$ & \textbf{51.16}$_{\scalebox{0.6}{6.81}}$ & \cellcolor{tableblue}\textbf{55.01}$_{\scalebox{0.6}{3.22}}$ & 58.95$_{\scalebox{0.6}{6.05}}$ & \textbf{62.53}$_{\scalebox{0.6}{0.63}}$ & \textbf{63.49}$_{\scalebox{0.6}{0.56}}$ & \textbf{63.48}$_{\scalebox{0.6}{0.69}}$ & \textbf{63.39}$_{\scalebox{0.6}{1.46}}$ & \cellcolor{tableblue}\textbf{62.37}$_{\scalebox{0.6}{1.95}}$ \\
\cmidrule(lr){2-14}
& SeqKD & 54.75$_{\scalebox{0.6}{1.75}}$ & 54.47$_{\scalebox{0.6}{1.07}}$ & 56.44$_{\scalebox{0.6}{1.11}}$ & 55.85$_{\scalebox{0.6}{1.57}}$ & \textbf{58.59}$_{\scalebox{0.6}{1.65}}$ & \cellcolor{tableblue}56.02$_{\scalebox{0.6}{1.64}}$ & 59.85$_{\scalebox{0.6}{0.41}}$ & \textbf{61.60}$_{\scalebox{0.6}{0.61}}$ & 61.96$_{\scalebox{0.6}{0.40}}$ & 62.37$_{\scalebox{0.6}{0.58}}$ & 62.94$_{\scalebox{0.6}{1.21}}$ & \cellcolor{tableblue}61.74$_{\scalebox{0.6}{1.17}}$ \\
& + \ours & \textbf{56.97}$_{\scalebox{0.6}{0.92}}$ & \textbf{57.66}$_{\scalebox{0.6}{1.14}}$ & \textbf{57.78}$_{\scalebox{0.6}{1.20}}$ & \textbf{58.74}$_{\scalebox{0.6}{0.76}}$ & 58.29$_{\scalebox{0.6}{1.49}}$ & \cellcolor{tableblue}\textbf{57.89}$_{\scalebox{0.6}{0.67}}$ & \textbf{61.72}$_{\scalebox{0.6}{0.49}}$ & 61.23$_{\scalebox{0.6}{0.77}}$ & \textbf{63.98}$_{\scalebox{0.6}{0.80}}$ & \textbf{63.72}$_{\scalebox{0.6}{0.76}}$ & \textbf{64.06}$_{\scalebox{0.6}{0.83}}$ & \cellcolor{tableblue}\textbf{62.94}$_{\scalebox{0.6}{1.36}}$ \\
\cmidrule(lr){1-14}

\multirow{6}{*}{SQuAD} 
& KD & 78.36$_{\scalebox{0.6}{0.77}}$ & 75.46$_{\scalebox{0.6}{2.06}}$ & 73.68$_{\scalebox{0.6}{3.48}}$ & 74.97$_{\scalebox{0.6}{1.75}}$ & 74.37$_{\scalebox{0.6}{1.95}}$ & \cellcolor{tableblue}75.37$_{\scalebox{0.6}{1.80}}$ & \textbf{70.27}$_{\scalebox{0.6}{0.18}}$ & 69.40$_{\scalebox{0.6}{0.28}}$ & 69.00$_{\scalebox{0.6}{0.40}}$ & 66.88$_{\scalebox{0.6}{0.58}}$ & \textbf{67.91}$_{\scalebox{0.6}{0.25}}$ & \cellcolor{tableblue}68.69$_{\scalebox{0.6}{1.32}}$ \\
& + \ours & \textbf{78.47}$_{\scalebox{0.6}{0.26}}$ & \textbf{77.86}$_{\scalebox{0.6}{1.15}}$ & \textbf{77.23}$_{\scalebox{0.6}{1.18}}$ & \textbf{77.11}$_{\scalebox{0.6}{0.46}}$ & \textbf{75.84}$_{\scalebox{0.6}{1.68}}$ & \cellcolor{tableblue}\textbf{77.30}$_{\scalebox{0.6}{0.99}}$ & 69.39$_{\scalebox{0.6}{0.57}}$ & \textbf{70.13}$_{\scalebox{0.6}{0.69}}$ & \textbf{69.93}$_{\scalebox{0.6}{0.55}}$ & \textbf{68.92}$_{\scalebox{0.6}{0.92}}$ & 67.21$_{\scalebox{0.6}{0.67}}$ & \cellcolor{tableblue}\textbf{69.12}$_{\scalebox{0.6}{1.17}}$ \\
\cmidrule(lr){2-14}
& LWD & 78.36$_{\scalebox{0.6}{0.68}}$ & 75.43$_{\scalebox{0.6}{1.58}}$ & 74.06$_{\scalebox{0.6}{3.53}}$ & 73.84$_{\scalebox{0.6}{2.17}}$ & 74.44$_{\scalebox{0.6}{1.54}}$ & \cellcolor{tableblue}75.23$_{\scalebox{0.6}{1.85}}$ & \textbf{69.40}$_{\scalebox{0.6}{0.82}}$ & 69.51$_{\scalebox{0.6}{0.30}}$ & 68.95$_{\scalebox{0.6}{0.31}}$ & 67.55$_{\scalebox{0.6}{0.90}}$ & 68.92$_{\scalebox{0.6}{1.15}}$ & \cellcolor{tableblue}68.87$_{\scalebox{0.6}{0.78}}$ \\
& + \ours & \textbf{78.37}$_{\scalebox{0.6}{0.33}}$ & \textbf{77.83}$_{\scalebox{0.6}{1.14}}$ & \textbf{76.54}$_{\scalebox{0.6}{1.06}}$ & \textbf{77.23}$_{\scalebox{0.6}{0.60}}$ & \textbf{76.13}$_{\scalebox{0.6}{1.79}}$ & \cellcolor{tableblue}\textbf{77.22}$_{\scalebox{0.6}{0.92}}$ & 68.95$_{\scalebox{0.6}{0.17}}$ & \textbf{69.85}$_{\scalebox{0.6}{0.24}}$ & \textbf{70.07}$_{\scalebox{0.6}{0.75}}$ & \textbf{69.38}$_{\scalebox{0.6}{0.31}}$ & \textbf{69.74}$_{\scalebox{0.6}{1.37}}$ & \cellcolor{tableblue}\textbf{69.60}$_{\scalebox{0.6}{0.44}}$ \\
\cmidrule(lr){2-14}
& SeqKD & 76.45$_{\scalebox{0.6}{0.95}}$ & \textbf{77.94}$_{\scalebox{0.6}{0.93}}$ & 76.87$_{\scalebox{0.6}{0.67}}$ & 76.95$_{\scalebox{0.6}{0.18}}$ & 76.38$_{\scalebox{0.6}{0.64}}$ & \cellcolor{tableblue}76.92$_{\scalebox{0.6}{0.62}}$ & \textbf{70.77}$_{\scalebox{0.6}{0.46}}$ & 70.96$_{\scalebox{0.6}{0.28}}$ & 70.78$_{\scalebox{0.6}{0.47}}$ & 70.32$_{\scalebox{0.6}{0.39}}$ & 71.38$_{\scalebox{0.6}{0.74}}$ & \cellcolor{tableblue}70.84$_{\scalebox{0.6}{0.38}}$ \\
& + \ours & \textbf{78.15}$_{\scalebox{0.6}{0.48}}$ & 77.30$_{\scalebox{0.6}{0.46}}$ & \textbf{77.69}$_{\scalebox{0.6}{0.71}}$ & \textbf{77.27}$_{\scalebox{0.6}{0.68}}$ & \textbf{76.56}$_{\scalebox{0.6}{0.15}}$ & \cellcolor{tableblue}\textbf{77.39}$_{\scalebox{0.6}{0.59}}$ & 70.04$_{\scalebox{0.6}{0.44}}$ & \textbf{71.00}$_{\scalebox{0.6}{0.43}}$ & \textbf{71.12}$_{\scalebox{0.6}{0.88}}$ & \textbf{70.78}$_{\scalebox{0.6}{0.62}}$ & \textbf{71.46}$_{\scalebox{0.6}{1.00}}$ & \cellcolor{tableblue}\textbf{70.88}$_{\scalebox{0.6}{0.53}}$ \\
\cmidrule(lr){1-14}

\multirow{6}{*}{NewsQA} 
& KD & \textbf{64.99}$_{\scalebox{0.6}{2.59}}$ & 64.20$_{\scalebox{0.6}{2.93}}$ & 59.01$_{\scalebox{0.6}{4.29}}$ & 55.83$_{\scalebox{0.6}{2.73}}$ & 50.14$_{\scalebox{0.6}{1.27}}$ & \cellcolor{tableblue}58.83$_{\scalebox{0.6}{6.15}}$ & 48.20$_{\scalebox{0.6}{1.05}}$ & 43.74$_{\scalebox{0.6}{1.28}}$ & 42.99$_{\scalebox{0.6}{1.22}}$ & 45.19$_{\scalebox{0.6}{1.65}}$ & 42.93$_{\scalebox{0.6}{1.48}}$ & \cellcolor{tableblue}44.61$_{\scalebox{0.6}{2.21}}$ \\
& + \ours & 64.78$_{\scalebox{0.6}{3.94}}$ & \textbf{67.59}$_{\scalebox{0.6}{1.30}}$ & \textbf{67.39}$_{\scalebox{0.6}{2.56}}$ & \textbf{64.35}$_{\scalebox{0.6}{2.37}}$ & \textbf{59.80}$_{\scalebox{0.6}{3.90}}$ & \cellcolor{tableblue}\textbf{64.78}$_{\scalebox{0.6}{3.15}}$ & \textbf{52.29}$_{\scalebox{0.6}{0.50}}$ & \textbf{49.33}$_{\scalebox{0.6}{1.11}}$ & \textbf{47.28}$_{\scalebox{0.6}{1.13}}$ & \textbf{46.67}$_{\scalebox{0.6}{1.56}}$ & \textbf{48.93}$_{\scalebox{0.6}{2.76}}$ & \cellcolor{tableblue}\textbf{48.90}$_{\scalebox{0.6}{2.20}}$ \\
\cmidrule(lr){2-14}
& LWD & \textbf{64.75}$_{\scalebox{0.6}{2.54}}$ & 63.80$_{\scalebox{0.6}{3.28}}$ & 58.87$_{\scalebox{0.6}{4.49}}$ & 54.14$_{\scalebox{0.6}{4.04}}$ & 50.14$_{\scalebox{0.6}{0.74}}$ & \cellcolor{tableblue}58.34$_{\scalebox{0.6}{6.24}}$ & 49.16$_{\scalebox{0.6}{0.55}}$ & 48.29$_{\scalebox{0.6}{1.11}}$ & 46.81$_{\scalebox{0.6}{1.47}}$ & 47.33$_{\scalebox{0.6}{1.21}}$ & 44.93$_{\scalebox{0.6}{2.77}}$ & \cellcolor{tableblue}47.30$_{\scalebox{0.6}{1.60}}$ \\
& + \ours & 64.43$_{\scalebox{0.6}{4.04}}$ & \textbf{68.06}$_{\scalebox{0.6}{0.73}}$ & \textbf{66.93}$_{\scalebox{0.6}{1.52}}$ & \textbf{63.62}$_{\scalebox{0.6}{2.63}}$ & \textbf{60.41}$_{\scalebox{0.6}{4.62}}$ & \cellcolor{tableblue}\textbf{64.69}$_{\scalebox{0.6}{3.00}}$ & \textbf{52.26}$_{\scalebox{0.6}{0.82}}$ & \textbf{49.19}$_{\scalebox{0.6}{1.21}}$ & \textbf{48.12}$_{\scalebox{0.6}{1.39}}$ & \textbf{50.23}$_{\scalebox{0.6}{0.52}}$ & \textbf{51.68}$_{\scalebox{0.6}{4.43}}$ & \cellcolor{tableblue}\textbf{50.30}$_{\scalebox{0.6}{1.71}}$ \\
\cmidrule(lr){2-14}
& SeqKD & 54.58$_{\scalebox{0.6}{2.52}}$ & 55.68$_{\scalebox{0.6}{2.89}}$ & 58.23$_{\scalebox{0.6}{3.72}}$ & 60.67$_{\scalebox{0.6}{3.24}}$ & 62.23$_{\scalebox{0.6}{1.37}}$ & \cellcolor{tableblue}58.28$_{\scalebox{0.6}{3.23}}$ & 41.54$_{\scalebox{0.6}{1.12}}$ & 43.74$_{\scalebox{0.6}{0.88}}$ & 45.77$_{\scalebox{0.6}{0.55}}$ & 48.41$_{\scalebox{0.6}{1.32}}$ & 52.00$_{\scalebox{0.6}{1.08}}$ & \cellcolor{tableblue}46.29$_{\scalebox{0.6}{4.08}}$ \\
& + \ours & \textbf{63.80}$_{\scalebox{0.6}{2.04}}$ & \textbf{66.35}$_{\scalebox{0.6}{2.23}}$ & \textbf{66.20}$_{\scalebox{0.6}{1.23}}$ & \textbf{70.09}$_{\scalebox{0.6}{1.82}}$ & \textbf{68.17}$_{\scalebox{0.6}{2.03}}$ & \cellcolor{tableblue}\textbf{66.92}$_{\scalebox{0.6}{2.36}}$ & \textbf{42.58}$_{\scalebox{0.6}{1.04}}$ & \textbf{45.71}$_{\scalebox{0.6}{1.17}}$ & \textbf{48.17}$_{\scalebox{0.6}{0.85}}$ & \textbf{50.00}$_{\scalebox{0.6}{1.99}}$ & \textbf{53.62}$_{\scalebox{0.6}{2.15}}$ & \cellcolor{tableblue}\textbf{48.02}$_{\scalebox{0.6}{4.19}}$ \\
\cmidrule(lr){1-14}

\multirow{6}{*}{TriviaQA} 
& KD & 50.54$_{\scalebox{0.6}{3.88}}$ & 47.71$_{\scalebox{0.6}{3.13}}$ & 44.38$_{\scalebox{0.6}{2.08}}$ & 44.12$_{\scalebox{0.6}{2.82}}$ & 45.30$_{\scalebox{0.6}{1.17}}$ & \cellcolor{tableblue}46.41$_{\scalebox{0.6}{2.71}}$ & 58.39$_{\scalebox{0.6}{1.08}}$ & \textbf{58.68}$_{\scalebox{0.6}{0.90}}$ & 56.90$_{\scalebox{0.6}{1.23}}$ & 54.43$_{\scalebox{0.6}{1.23}}$ & \textbf{56.05}$_{\scalebox{0.6}{0.79}}$ & \cellcolor{tableblue}56.89$_{\scalebox{0.6}{1.75}}$ \\
& + \ours & \textbf{56.47}$_{\scalebox{0.6}{1.45}}$ & \textbf{53.05}$_{\scalebox{0.6}{2.10}}$ & \textbf{53.15}$_{\scalebox{0.6}{6.24}}$ & \textbf{48.20}$_{\scalebox{0.6}{1.94}}$ & \textbf{47.45}$_{\scalebox{0.6}{1.73}}$ & \cellcolor{tableblue}\textbf{51.66}$_{\scalebox{0.6}{3.78}}$ & \textbf{59.04}$_{\scalebox{0.6}{0.48}}$ & 57.46$_{\scalebox{0.6}{0.65}}$ & \textbf{59.24}$_{\scalebox{0.6}{0.69}}$ & \textbf{57.66}$_{\scalebox{0.6}{0.92}}$ & 55.19$_{\scalebox{0.6}{1.49}}$ & \cellcolor{tableblue}\textbf{57.72}$_{\scalebox{0.6}{1.62}}$ \\
\cmidrule(lr){2-14}
& LWD & 50.51$_{\scalebox{0.6}{3.54}}$ & 49.00$_{\scalebox{0.6}{2.51}}$ & 42.70$_{\scalebox{0.6}{2.28}}$ & 43.26$_{\scalebox{0.6}{1.75}}$ & 45.50$_{\scalebox{0.6}{1.07}}$ & \cellcolor{tableblue}46.19$_{\scalebox{0.6}{3.46}}$ & 57.79$_{\scalebox{0.6}{0.32}}$ & 57.96$_{\scalebox{0.6}{0.83}}$ & 56.24$_{\scalebox{0.6}{0.30}}$ & 54.99$_{\scalebox{0.6}{0.59}}$ & \textbf{55.75}$_{\scalebox{0.6}{1.27}}$ & \cellcolor{tableblue}56.55$_{\scalebox{0.6}{1.29}}$ \\
& + \ours & \textbf{56.18}$_{\scalebox{0.6}{1.51}}$ & \textbf{52.98}$_{\scalebox{0.6}{2.33}}$ & \textbf{53.11}$_{\scalebox{0.6}{6.50}}$ & \textbf{48.01}$_{\scalebox{0.6}{1.59}}$ & \textbf{47.64}$_{\scalebox{0.6}{0.98}}$ & \cellcolor{tableblue}\textbf{51.58}$_{\scalebox{0.6}{3.66}}$ & \textbf{58.95}$_{\scalebox{0.6}{0.59}}$ & \textbf{58.42}$_{\scalebox{0.6}{0.99}}$ & \textbf{58.48}$_{\scalebox{0.6}{0.35}}$ & \textbf{57.27}$_{\scalebox{0.6}{1.17}}$ & 54.83$_{\scalebox{0.6}{1.27}}$ & \cellcolor{tableblue}\textbf{57.59}$_{\scalebox{0.6}{1.66}}$ \\
\cmidrule(lr){2-14}
& SeqKD & 55.35$_{\scalebox{0.6}{3.42}}$ & 50.05$_{\scalebox{0.6}{1.85}}$ & 47.51$_{\scalebox{0.6}{2.53}}$ & 48.14$_{\scalebox{0.6}{1.96}}$ & 46.56$_{\scalebox{0.6}{1.92}}$ & \cellcolor{tableblue}49.52$_{\scalebox{0.6}{3.50}}$ & 58.29$_{\scalebox{0.6}{0.90}}$ & 56.97$_{\scalebox{0.6}{0.38}}$ & 57.66$_{\scalebox{0.6}{0.53}}$ & 56.84$_{\scalebox{0.6}{1.21}}$ & \textbf{57.53}$_{\scalebox{0.6}{1.28}}$ & \cellcolor{tableblue}57.46$_{\scalebox{0.6}{0.58}}$ \\
& + \ours & \textbf{56.80}$_{\scalebox{0.6}{1.98}}$ & \textbf{51.86}$_{\scalebox{0.6}{1.82}}$ & \textbf{51.83}$_{\scalebox{0.6}{1.36}}$ & \textbf{48.53}$_{\scalebox{0.6}{2.97}}$ & \textbf{51.14}$_{\scalebox{0.6}{1.35}}$ & \cellcolor{tableblue}\textbf{52.03}$_{\scalebox{0.6}{2.99}}$ & \textbf{59.70}$_{\scalebox{0.6}{0.85}}$ & \textbf{59.11}$_{\scalebox{0.6}{1.01}}$ & \textbf{58.85}$_{\scalebox{0.6}{0.75}}$ & \textbf{57.23}$_{\scalebox{0.6}{0.69}}$ & 56.97$_{\scalebox{0.6}{0.76}}$ & \cellcolor{tableblue}\textbf{58.37}$_{\scalebox{0.6}{1.21}}$ \\
\cmidrule(lr){1-14}

\multirow{6}{*}{HotpotQA} 
& KD & 57.42$_{\scalebox{0.6}{5.77}}$ & 58.98$_{\scalebox{0.6}{7.55}}$ & 42.35$_{\scalebox{0.6}{6.57}}$ & 57.57$_{\scalebox{0.6}{2.64}}$ & 57.64$_{\scalebox{0.6}{5.67}}$ & \cellcolor{tableblue}54.79$_{\scalebox{0.6}{6.98}}$ & \textbf{67.31}$_{\scalebox{0.6}{0.57}}$ & \textbf{68.16}$_{\scalebox{0.6}{0.42}}$ & 65.86$_{\scalebox{0.6}{1.18}}$ & 65.88$_{\scalebox{0.6}{0.37}}$ & 66.40$_{\scalebox{0.6}{1.26}}$ & \cellcolor{tableblue}66.72$_{\scalebox{0.6}{1.00}}$ \\
& + \ours & \textbf{58.16}$_{\scalebox{0.6}{8.44}}$ & \textbf{64.03}$_{\scalebox{0.6}{4.40}}$ & \textbf{59.96}$_{\scalebox{0.6}{2.24}}$ & \textbf{58.81}$_{\scalebox{0.6}{4.43}}$ & \textbf{63.19}$_{\scalebox{0.6}{3.02}}$ & \cellcolor{tableblue}\textbf{60.83}$_{\scalebox{0.6}{2.63}}$ & 66.12$_{\scalebox{0.6}{1.10}}$ & 67.91$_{\scalebox{0.6}{0.48}}$ & \textbf{69.40}$_{\scalebox{0.6}{0.74}}$ & \textbf{68.75}$_{\scalebox{0.6}{0.75}}$ & \textbf{68.79}$_{\scalebox{0.6}{1.00}}$ & \cellcolor{tableblue}\textbf{68.19}$_{\scalebox{0.6}{1.27}}$ \\
\cmidrule(lr){2-14}
& LWD & \textbf{56.93}$_{\scalebox{0.6}{5.65}}$ & 59.03$_{\scalebox{0.6}{8.17}}$ & 42.16$_{\scalebox{0.6}{6.37}}$ & 56.03$_{\scalebox{0.6}{5.04}}$ & 58.39$_{\scalebox{0.6}{5.61}}$ & \cellcolor{tableblue}54.51$_{\scalebox{0.6}{7.00}}$ & \textbf{67.32}$_{\scalebox{0.6}{0.58}}$ & 67.03$_{\scalebox{0.6}{0.72}}$ & 65.46$_{\scalebox{0.6}{0.93}}$ & 67.03$_{\scalebox{0.6}{0.45}}$ & 67.10$_{\scalebox{0.6}{2.22}}$ & \cellcolor{tableblue}66.79$_{\scalebox{0.6}{0.75}}$ \\
& + \ours & 56.87$_{\scalebox{0.6}{9.31}}$ & \textbf{65.58}$_{\scalebox{0.6}{3.89}}$ & \textbf{62.01}$_{\scalebox{0.6}{1.86}}$ & \textbf{58.25}$_{\scalebox{0.6}{4.73}}$ & \textbf{63.55}$_{\scalebox{0.6}{3.92}}$ & \cellcolor{tableblue}\textbf{61.25}$_{\scalebox{0.6}{3.63}}$ & 66.41$_{\scalebox{0.6}{1.07}}$ & \textbf{67.55}$_{\scalebox{0.6}{0.54}}$ & \textbf{67.30}$_{\scalebox{0.6}{0.79}}$ & \textbf{68.63}$_{\scalebox{0.6}{0.97}}$ & \textbf{70.52}$_{\scalebox{0.6}{1.08}}$ & \cellcolor{tableblue}\textbf{68.08}$_{\scalebox{0.6}{1.57}}$ \\
\cmidrule(lr){2-14}
& SeqKD & 67.69$_{\scalebox{0.6}{1.02}}$ & 67.82$_{\scalebox{0.6}{0.94}}$ & 67.81$_{\scalebox{0.6}{2.05}}$ & 64.14$_{\scalebox{0.6}{5.13}}$ & \textbf{66.56}$_{\scalebox{0.6}{3.21}}$ & \cellcolor{tableblue}66.80$_{\scalebox{0.6}{1.58}}$ & \textbf{68.14}$_{\scalebox{0.6}{0.77}}$ & \textbf{69.02}$_{\scalebox{0.6}{0.94}}$ & 68.65$_{\scalebox{0.6}{0.43}}$ & 66.00$_{\scalebox{0.6}{0.72}}$ & 67.30$_{\scalebox{0.6}{1.53}}$ & \cellcolor{tableblue}\textbf{67.82}$_{\scalebox{0.6}{1.20}}$ \\
& + \ours & \textbf{68.32}$_{\scalebox{0.6}{1.38}}$ & \textbf{70.32}$_{\scalebox{0.6}{0.59}}$ & \textbf{69.00}$_{\scalebox{0.6}{2.03}}$ & \textbf{68.08}$_{\scalebox{0.6}{2.52}}$ & 66.13$_{\scalebox{0.6}{4.35}}$ & \cellcolor{tableblue}\textbf{68.37}$_{\scalebox{0.6}{1.52}}$ & 65.88$_{\scalebox{0.6}{0.68}}$ & 66.57$_{\scalebox{0.6}{1.26}}$ & \textbf{68.68}$_{\scalebox{0.6}{0.60}}$ & \textbf{69.32}$_{\scalebox{0.6}{0.64}}$ & \textbf{68.28}$_{\scalebox{0.6}{1.59}}$ & \cellcolor{tableblue}67.74$_{\scalebox{0.6}{1.46}}$ \\
\midrule

% OVERALL AVERAGE ROWS
\rowcolor{tableblue} 
& KD & 61.88$_{\scalebox{0.6}{10.53}}$ & 60.82$_{\scalebox{0.6}{10.13}}$ & 54.83$_{\scalebox{0.6}{12.63}}$ & 56.07$_{\scalebox{0.6}{11.94}}$ & 55.56$_{\scalebox{0.6}{11.40}}$ & \cellcolor{tableblue}57.83$_{\scalebox{0.6}{10.78}}$ & 61.22$_{\scalebox{0.6}{8.62}}$ & 60.24$_{\scalebox{0.6}{10.27}}$ & 59.13$_{\scalebox{0.6}{10.14}}$ & 58.51$_{\scalebox{0.6}{8.96}}$ & 59.07$_{\scalebox{0.6}{10.13}}$ & \cellcolor{tableblue}59.64$_{\scalebox{0.6}{9.59}}$ \\
\rowcolor{tableblue} 
& + \ours & \textbf{63.36}$_{\scalebox{0.6}{9.01}}$ & \textbf{64.20}$_{\scalebox{0.6}{9.42}}$ & \textbf{62.93}$_{\scalebox{0.6}{9.55}}$ & \textbf{59.67}$_{\scalebox{0.6}{11.77}}$ & \textbf{59.49}$_{\scalebox{0.6}{11.12}}$ & \cellcolor{tableblue}\textbf{61.93}$_{\scalebox{0.6}{9.98}}$ & \textbf{61.73}$_{\scalebox{0.6}{6.60}}$ & \textbf{61.59}$_{\scalebox{0.6}{8.40}}$ & \textbf{61.78}$_{\scalebox{0.6}{9.26}}$ & \textbf{60.93}$_{\scalebox{0.6}{9.25}}$ & \textbf{60.59}$_{\scalebox{0.6}{8.39}}$ & \cellcolor{tableblue}\textbf{61.32}$_{\scalebox{0.6}{8.32}}$ \\
\cmidrule(lr){2-14}
\rowcolor{tableblue} 
& LWD & 61.75$_{\scalebox{0.6}{10.57}}$ & 60.94$_{\scalebox{0.6}{9.70}}$ & 54.52$_{\scalebox{0.6}{13.16}}$ & 55.49$_{\scalebox{0.6}{11.37}}$ & 55.88$_{\scalebox{0.6}{11.36}}$ & \cellcolor{tableblue}57.72$_{\scalebox{0.6}{10.74}}$ & 60.77$_{\scalebox{0.6}{8.08}}$ & 60.94$_{\scalebox{0.6}{8.37}}$ & 59.71$_{\scalebox{0.6}{8.64}}$ & 59.52$_{\scalebox{0.6}{8.53}}$ & 59.70$_{\scalebox{0.6}{9.72}}$ & \cellcolor{tableblue}60.13$_{\scalebox{0.6}{8.65}}$ \\
\rowcolor{tableblue} 
\textbf{AVG} & + \ours & \textbf{62.94}$_{\scalebox{0.6}{9.22}}$ & \textbf{64.38}$_{\scalebox{0.6}{9.67}}$ & \textbf{62.72}$_{\scalebox{0.6}{9.50}}$ & \textbf{59.94}$_{\scalebox{0.6}{11.31}}$ & \textbf{59.78}$_{\scalebox{0.6}{11.22}}$ & \cellcolor{tableblue}\textbf{61.95}$_{\scalebox{0.6}{9.96}}$ & \textbf{61.10}$_{\scalebox{0.6}{6.66}}$ & \textbf{61.51}$_{\scalebox{0.6}{8.19}}$ & \textbf{61.49}$_{\scalebox{0.6}{8.65}}$ & \textbf{61.80}$_{\scalebox{0.6}{8.08}}$ & \textbf{62.03}$_{\scalebox{0.6}{8.55}}$ & \cellcolor{tableblue}\textbf{61.59}$_{\scalebox{0.6}{7.91}}$ \\
\cmidrule(lr){2-14}
\rowcolor{tableblue} 
& SeqKD & 61.77$_{\scalebox{0.6}{9.91}}$ & 61.19$_{\scalebox{0.6}{11.44}}$ & 61.37$_{\scalebox{0.6}{11.27}}$ & 61.15$_{\scalebox{0.6}{10.68}}$ & 62.06$_{\scalebox{0.6}{10.93}}$ & \cellcolor{tableblue}61.51$_{\scalebox{0.6}{10.60}}$ & 59.72$_{\scalebox{0.6}{11.46}}$ & 60.46$_{\scalebox{0.6}{10.92}}$ & 60.96$_{\scalebox{0.6}{9.98}}$ & 60.79$_{\scalebox{0.6}{8.51}}$ & 62.23$_{\scalebox{0.6}{7.69}}$ & \cellcolor{tableblue}60.83$_{\scalebox{0.6}{9.65}}$ \\
\rowcolor{tableblue} 
& + \ours & \textbf{64.81}$_{\scalebox{0.6}{8.90}}$ & \textbf{64.70}$_{\scalebox{0.6}{10.09}}$ & \textbf{64.50}$_{\scalebox{0.6}{10.03}}$ & \textbf{64.54}$_{\scalebox{0.6}{11.13}}$ & \textbf{64.06}$_{\scalebox{0.6}{9.72}}$ & \cellcolor{tableblue}\textbf{64.52}$_{\scalebox{0.6}{9.83}}$ & \textbf{59.98}$_{\scalebox{0.6}{10.51}}$ & \textbf{60.72}$_{\scalebox{0.6}{9.59}}$ & \textbf{62.16}$_{\scalebox{0.6}{9.12}}$ & \textbf{62.21}$_{\scalebox{0.6}{8.66}}$ & \textbf{62.88}$_{\scalebox{0.6}{7.50}}$ & \cellcolor{tableblue}\textbf{61.59}$_{\scalebox{0.6}{8.95}}$ \\
\bottomrule
\end{tabular}%
}
\end{table*}

\newpage
\subsection{In-Domain Agreement Results} \label{app:agr}

\begin{figure*}[htbp]
    \centering
    \includegraphics[width=\textwidth]{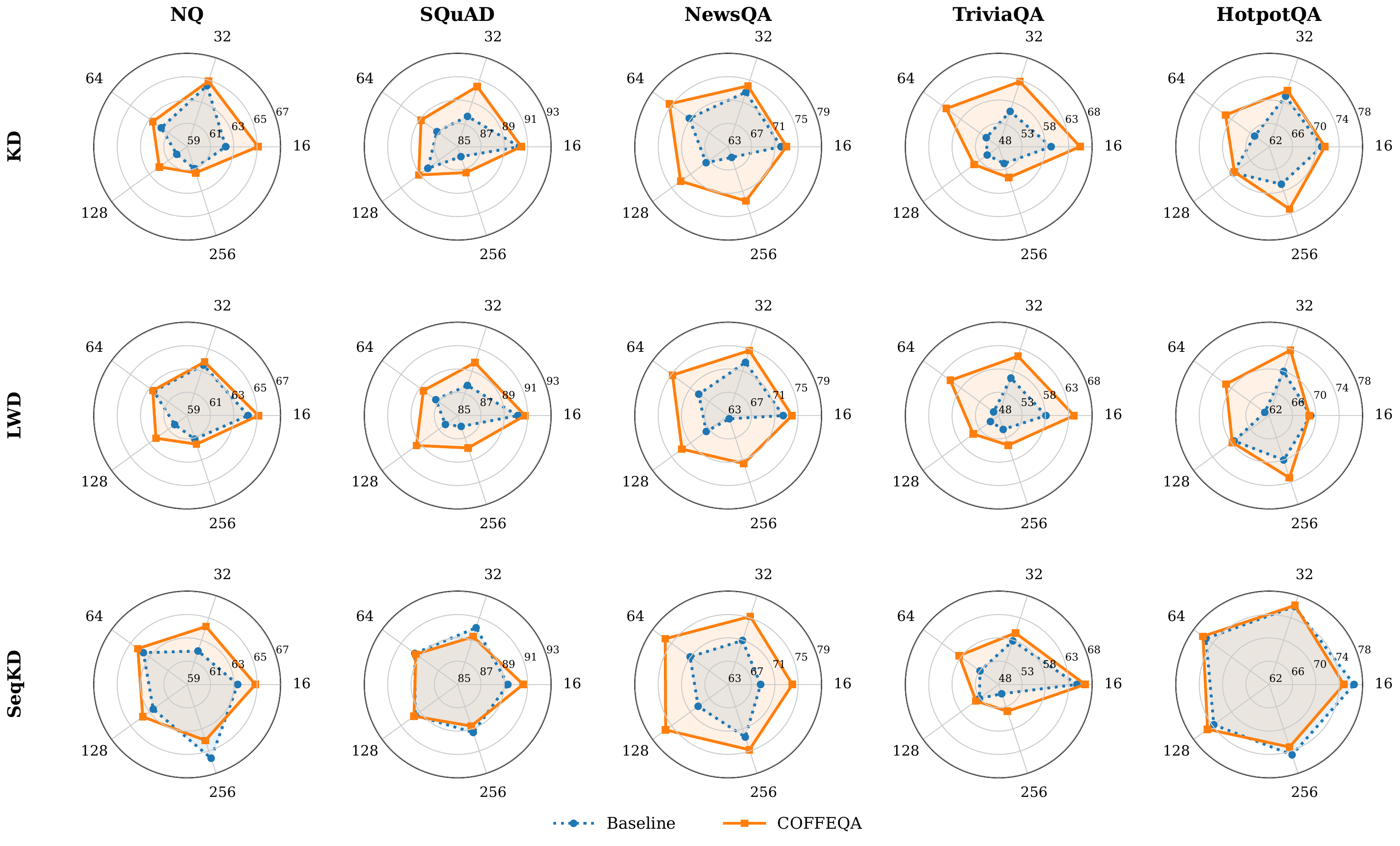}
    \caption{Teacher-student F1 agreement of KD, LWD, and SeqKD with and without \ours across datasets and $k$-shot settings using T5. \ours generally achieves higher agreement across the  distillation baselines. Spokes denote $k$-shot settings, and radial axis F1 score ranges are adjusted independently by dataset for visualization.}
    \label{fig:agreement_f1_radar_t5_full}
\end{figure*}

\begin{figure*}[htbp]
    \centering
    \includegraphics[width=\textwidth]{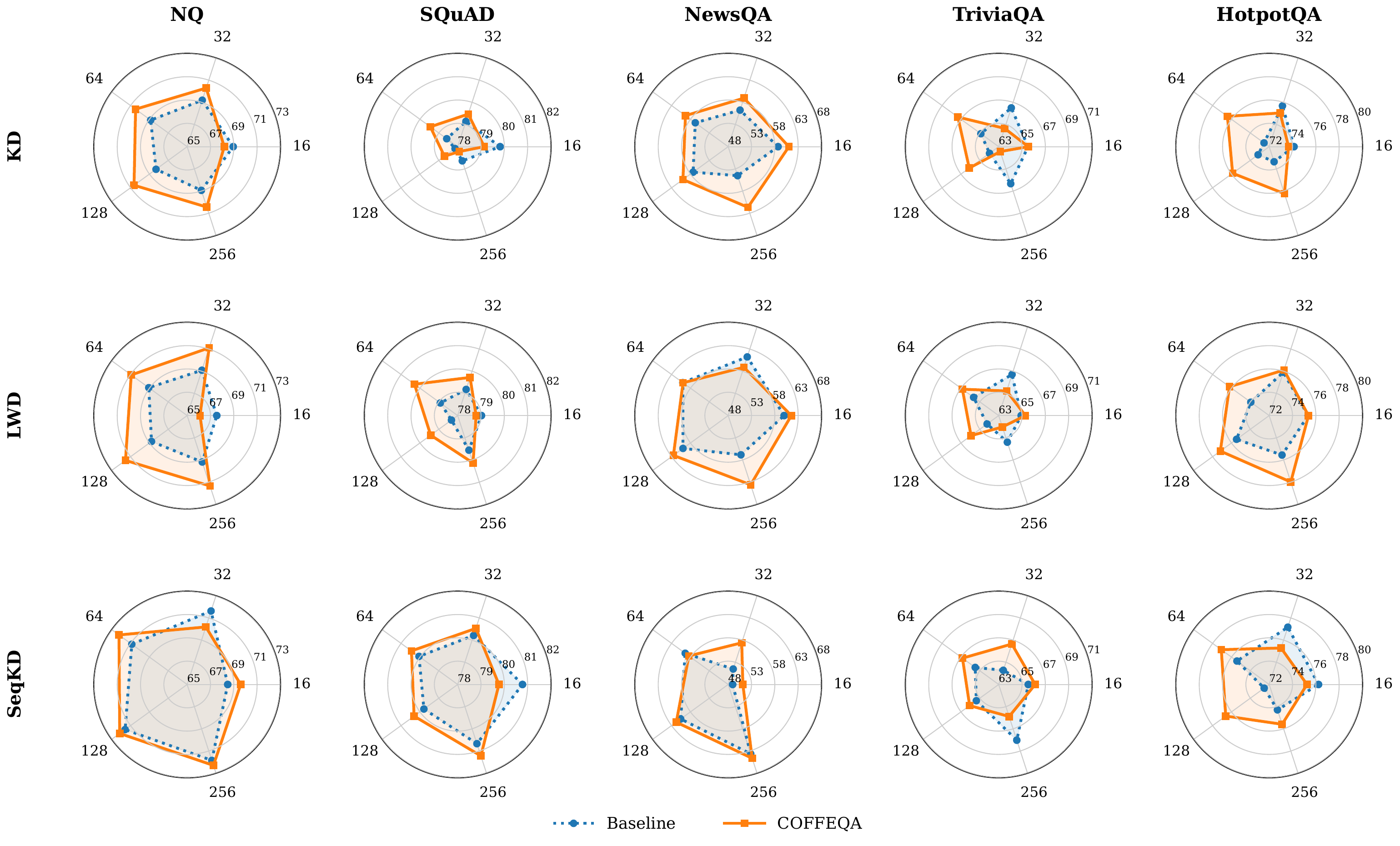}
    \caption{Teacher-student F1 agreement of KD, LWD, and SeqKD with and without \ours across datasets and $k$-shot settings using Qwen2.5. \ours generally achieves higher agreement across the  distillation baselines. Spokes denote $k$-shot settings, and radial axis F1 score ranges are adjusted independently by dataset for visualization.}
    \label{fig:agreement_f1_radar_qwen_full}
\end{figure*}

Tables~\ref{tab:indomain_agreement_em} and~\ref{tab:indomain_agreement_f1} report the in-domain teacher-student agreement results using EM and F1, respectively. For better visualization, Figures~\ref{fig:agreement_f1_radar_t5_full} and~\ref{fig:agreement_f1_radar_qwen_full} illustrate the F1 agreement across datasets and $k$-shot settings for all baselines using T5 and Qwen2.5, respectively.

\begin{table*}[ht]
\caption{In-domain teacher-student agreement EM score comparison across $k$-shots for T5 and Qwen2.5 models using baselines. Mean agreement EM scores are reported with standard deviations as subscripts ($Mean_{std}$). Bold values indicate the better-performing method (\ours vs. baseline). Blue shading highlights aggregate results.}
\label{tab:indomain_agreement_em}
\centering
\fontsize{6.5pt}{8.5pt}\selectfont
\setlength{\tabcolsep}{1.0pt}
\renewcommand{\arraystretch}{0.95}

\definecolor{tableblue}{HTML}{E6F0FF}

\resizebox{\textwidth}{!}{%
\begin{tabular}{llcccccccccccccc}
\toprule
 & & \multicolumn{6}{c}{\textbf{T5}} & \multicolumn{6}{c}{\textbf{Qwen2.5}} \\
\cmidrule(lr){3-8} \cmidrule(lr){9-14}
Dataset & Method & 16 & 32 & 64 & 128 & 256 & \cellcolor{tableblue}Avg & 16 & 32 & 64 & 128 & 256 & \cellcolor{tableblue}Avg \\
\midrule

\multirow{6}{*}{NQ} 
& KD & 56.23$_{\scalebox{0.6}{4.87}}$ & 60.20$_{\scalebox{0.6}{2.87}}$ & 54.47$_{\scalebox{0.6}{6.56}}$ & 50.26$_{\scalebox{0.6}{7.34}}$ & 52.62$_{\scalebox{0.6}{5.00}}$ & \cellcolor{tableblue}54.76$_{\scalebox{0.6}{3.76}}$ & \textbf{61.88}$_{\scalebox{0.6}{0.68}}$ & 62.93$_{\scalebox{0.6}{0.40}}$ & 62.84$_{\scalebox{0.6}{0.62}}$ & 62.53$_{\scalebox{0.6}{0.22}}$ & 63.73$_{\scalebox{0.6}{0.83}}$ & \cellcolor{tableblue}62.78$_{\scalebox{0.6}{0.67}}$ \\
& + \ours & \textbf{61.34}$_{\scalebox{0.6}{0.56}}$ & \textbf{60.64}$_{\scalebox{0.6}{0.49}}$ & \textbf{57.97}$_{\scalebox{0.6}{1.36}}$ & \textbf{52.94}$_{\scalebox{0.6}{6.72}}$ & \textbf{53.46}$_{\scalebox{0.6}{4.92}}$ & \cellcolor{tableblue}\textbf{57.27}$_{\scalebox{0.6}{3.93}}$ & 61.46$_{\scalebox{0.6}{0.53}}$ & \textbf{62.97}$_{\scalebox{0.6}{0.32}}$ & \textbf{64.52}$_{\scalebox{0.6}{0.13}}$ & \textbf{64.94}$_{\scalebox{0.6}{0.11}}$ & \textbf{64.72}$_{\scalebox{0.6}{1.17}}$ & \cellcolor{tableblue}\textbf{63.72}$_{\scalebox{0.6}{1.48}}$ \\
\cmidrule(lr){2-14}
& LWD & 59.86$_{\scalebox{0.6}{0.63}}$ & 58.94$_{\scalebox{0.6}{4.18}}$ & 56.42$_{\scalebox{0.6}{3.20}}$ & 51.35$_{\scalebox{0.6}{4.46}}$ & 52.22$_{\scalebox{0.6}{5.87}}$ & \cellcolor{tableblue}55.76$_{\scalebox{0.6}{3.85}}$ & \textbf{60.08}$_{\scalebox{0.6}{0.64}}$ & 63.05$_{\scalebox{0.6}{0.76}}$ & 62.79$_{\scalebox{0.6}{1.26}}$ & 62.50$_{\scalebox{0.6}{0.77}}$ & 62.66$_{\scalebox{0.6}{1.13}}$ & \cellcolor{tableblue}62.21$_{\scalebox{0.6}{1.21}}$ \\
& + \ours & \textbf{61.31}$_{\scalebox{0.6}{0.40}}$ & \textbf{58.95}$_{\scalebox{0.6}{3.37}}$ & \textbf{56.57}$_{\scalebox{0.6}{4.41}}$ & \textbf{53.98}$_{\scalebox{0.6}{4.65}}$ & \textbf{52.74}$_{\scalebox{0.6}{7.23}}$ & \cellcolor{tableblue}\textbf{56.71}$_{\scalebox{0.6}{3.52}}$ & 58.92$_{\scalebox{0.6}{0.62}}$ & \textbf{63.87}$_{\scalebox{0.6}{0.70}}$ & \textbf{64.87}$_{\scalebox{0.6}{0.51}}$ & \textbf{65.51}$_{\scalebox{0.6}{0.56}}$ & \textbf{65.47}$_{\scalebox{0.6}{1.02}}$ & \cellcolor{tableblue}\textbf{63.73}$_{\scalebox{0.6}{2.77}}$ \\
\cmidrule(lr){2-14}
& SeqKD & 57.52$_{\scalebox{0.6}{1.96}}$ & 57.19$_{\scalebox{0.6}{1.23}}$ & 59.24$_{\scalebox{0.6}{1.76}}$ & 58.92$_{\scalebox{0.6}{1.83}}$ & \textbf{62.41}$_{\scalebox{0.6}{1.97}}$ & \cellcolor{tableblue}59.06$_{\scalebox{0.6}{2.07}}$ & 61.45$_{\scalebox{0.6}{0.52}}$ & \textbf{64.81}$_{\scalebox{0.6}{0.70}}$ & 65.01$_{\scalebox{0.6}{0.38}}$ & 65.86$_{\scalebox{0.6}{0.44}}$ & 66.30$_{\scalebox{0.6}{1.21}}$ & \cellcolor{tableblue}64.69$_{\scalebox{0.6}{1.91}}$ \\
& + \ours & \textbf{60.22}$_{\scalebox{0.6}{0.90}}$ & \textbf{59.69}$_{\scalebox{0.6}{1.20}}$ & \textbf{60.21}$_{\scalebox{0.6}{1.00}}$ & \textbf{60.26}$_{\scalebox{0.6}{1.37}}$ & 60.88$_{\scalebox{0.6}{1.53}}$ & \cellcolor{tableblue}\textbf{60.25}$_{\scalebox{0.6}{0.42}}$ & \textbf{62.45}$_{\scalebox{0.6}{0.47}}$ & 63.09$_{\scalebox{0.6}{0.32}}$ & \textbf{66.36}$_{\scalebox{0.6}{0.48}}$ & \textbf{66.86}$_{\scalebox{0.6}{0.45}}$ & \textbf{67.15}$_{\scalebox{0.6}{0.35}}$ & \cellcolor{tableblue}\textbf{65.18}$_{\scalebox{0.6}{2.23}}$ \\
\cmidrule(lr){1-14}

\multirow{6}{*}{SQuAD} 
& KD & 85.35$_{\scalebox{0.6}{0.75}}$ & 81.94$_{\scalebox{0.6}{1.37}}$ & 80.52$_{\scalebox{0.6}{3.17}}$ & 82.17$_{\scalebox{0.6}{2.32}}$ & 78.67$_{\scalebox{0.6}{6.21}}$ & \cellcolor{tableblue}81.73$_{\scalebox{0.6}{2.46}}$ & \textbf{73.30}$_{\scalebox{0.6}{0.35}}$ & 72.70$_{\scalebox{0.6}{0.26}}$ & 71.99$_{\scalebox{0.6}{0.47}}$ & 69.98$_{\scalebox{0.6}{0.81}}$ & \textbf{71.03}$_{\scalebox{0.6}{0.69}}$ & \cellcolor{tableblue}71.80$_{\scalebox{0.6}{1.32}}$ \\
& + \ours & \textbf{85.73}$_{\scalebox{0.6}{0.39}}$ & \textbf{85.55}$_{\scalebox{0.6}{0.58}}$ & \textbf{83.33}$_{\scalebox{0.6}{1.33}}$ & \textbf{83.96}$_{\scalebox{0.6}{0.45}}$ & \textbf{81.28}$_{\scalebox{0.6}{2.43}}$ & \cellcolor{tableblue}\textbf{83.97}$_{\scalebox{0.6}{1.82}}$ & 72.18$_{\scalebox{0.6}{0.76}}$ & \textbf{73.00}$_{\scalebox{0.6}{0.50}}$ & \textbf{72.55}$_{\scalebox{0.6}{0.76}}$ & \textbf{71.63}$_{\scalebox{0.6}{0.86}}$ & 70.30$_{\scalebox{0.6}{0.67}}$ & \cellcolor{tableblue}\textbf{71.93}$_{\scalebox{0.6}{1.04}}$ \\
\cmidrule(lr){2-14}
& LWD & 85.16$_{\scalebox{0.6}{1.09}}$ & 81.72$_{\scalebox{0.6}{2.03}}$ & 79.94$_{\scalebox{0.6}{4.61}}$ & 78.97$_{\scalebox{0.6}{2.73}}$ & 79.91$_{\scalebox{0.6}{2.29}}$ & \cellcolor{tableblue}81.14$_{\scalebox{0.6}{2.46}}$ & \textbf{72.52}$_{\scalebox{0.6}{0.60}}$ & 72.51$_{\scalebox{0.6}{0.28}}$ & 72.19$_{\scalebox{0.6}{0.52}}$ & 70.53$_{\scalebox{0.6}{1.14}}$ & 72.28$_{\scalebox{0.6}{1.10}}$ & \cellcolor{tableblue}72.01$_{\scalebox{0.6}{0.84}}$ \\
& + \ours & \textbf{85.98}$_{\scalebox{0.6}{0.40}}$ & \textbf{84.37}$_{\scalebox{0.6}{1.68}}$ & \textbf{83.13}$_{\scalebox{0.6}{1.30}}$ & \textbf{84.21}$_{\scalebox{0.6}{0.77}}$ & \textbf{82.19}$_{\scalebox{0.6}{2.26}}$ & \cellcolor{tableblue}\textbf{83.97}$_{\scalebox{0.6}{1.43}}$ & 72.24$_{\scalebox{0.6}{0.63}}$ & \textbf{73.16}$_{\scalebox{0.6}{0.29}}$ & \textbf{73.71}$_{\scalebox{0.6}{0.73}}$ & \textbf{72.37}$_{\scalebox{0.6}{0.18}}$ & \textbf{72.94}$_{\scalebox{0.6}{1.32}}$ & \cellcolor{tableblue}\textbf{72.88}$_{\scalebox{0.6}{0.60}}$ \\
\cmidrule(lr){2-14}
& SeqKD & 83.48$_{\scalebox{0.6}{1.46}}$ & \textbf{84.99}$_{\scalebox{0.6}{1.17}}$ & 83.98$_{\scalebox{0.6}{1.52}}$ & \textbf{84.57}$_{\scalebox{0.6}{0.25}}$ & \textbf{84.59}$_{\scalebox{0.6}{0.35}}$ & \cellcolor{tableblue}84.32$_{\scalebox{0.6}{0.59}}$ & \textbf{74.85}$_{\scalebox{0.6}{0.45}}$ & 74.62$_{\scalebox{0.6}{0.53}}$ & 74.09$_{\scalebox{0.6}{0.78}}$ & 73.90$_{\scalebox{0.6}{0.72}}$ & 74.88$_{\scalebox{0.6}{1.16}}$ & \cellcolor{tableblue}74.47$_{\scalebox{0.6}{0.45}}$ \\
& + \ours & \textbf{85.56}$_{\scalebox{0.6}{0.46}}$ & 84.21$_{\scalebox{0.6}{0.49}}$ & \textbf{84.52}$_{\scalebox{0.6}{0.70}}$ & 84.47$_{\scalebox{0.6}{0.88}}$ & 83.25$_{\scalebox{0.6}{0.72}}$ & \cellcolor{tableblue}\textbf{84.40}$_{\scalebox{0.6}{0.82}}$ & 73.81$_{\scalebox{0.6}{0.65}}$ & \textbf{75.00}$_{\scalebox{0.6}{0.45}}$ & \textbf{74.65}$_{\scalebox{0.6}{0.60}}$ & \textbf{74.40}$_{\scalebox{0.6}{0.43}}$ & \textbf{75.25}$_{\scalebox{0.6}{0.70}}$ & \cellcolor{tableblue}\textbf{74.62}$_{\scalebox{0.6}{0.56}}$ \\
\cmidrule(lr){1-14}

\multirow{6}{*}{NewsQA} 
& KD & 63.28$_{\scalebox{0.6}{2.18}}$ & 63.10$_{\scalebox{0.6}{3.40}}$ & 60.61$_{\scalebox{0.6}{2.80}}$ & 51.19$_{\scalebox{0.6}{4.13}}$ & 46.58$_{\scalebox{0.6}{1.56}}$ & \cellcolor{tableblue}56.95$_{\scalebox{0.6}{7.62}}$ & 47.74$_{\scalebox{0.6}{1.38}}$ & 44.12$_{\scalebox{0.6}{1.68}}$ & 43.65$_{\scalebox{0.6}{1.56}}$ & 46.44$_{\scalebox{0.6}{1.30}}$ & 42.52$_{\scalebox{0.6}{1.92}}$ & \cellcolor{tableblue}44.89$_{\scalebox{0.6}{2.14}}$ \\
& + \ours & \textbf{63.83}$_{\scalebox{0.6}{2.50}}$ & \textbf{66.09}$_{\scalebox{0.6}{0.66}}$ & \textbf{67.13}$_{\scalebox{0.6}{1.61}}$ & \textbf{61.71}$_{\scalebox{0.6}{4.24}}$ & \textbf{58.00}$_{\scalebox{0.6}{3.91}}$ & \cellcolor{tableblue}\textbf{63.35}$_{\scalebox{0.6}{3.65}}$ & \textbf{51.01}$_{\scalebox{0.6}{0.64}}$ & \textbf{48.44}$_{\scalebox{0.6}{1.27}}$ & \textbf{48.17}$_{\scalebox{0.6}{1.06}}$ & \textbf{47.42}$_{\scalebox{0.6}{1.02}}$ & \textbf{50.12}$_{\scalebox{0.6}{1.82}}$ & \cellcolor{tableblue}\textbf{49.03}$_{\scalebox{0.6}{1.48}}$ \\
\cmidrule(lr){2-14}
& LWD & 63.62$_{\scalebox{0.6}{3.09}}$ & 62.81$_{\scalebox{0.6}{3.88}}$ & 55.83$_{\scalebox{0.6}{3.91}}$ & 49.77$_{\scalebox{0.6}{4.38}}$ & 45.94$_{\scalebox{0.6}{1.27}}$ & \cellcolor{tableblue}55.59$_{\scalebox{0.6}{7.81}}$ & 48.49$_{\scalebox{0.6}{1.44}}$ & \textbf{49.33}$_{\scalebox{0.6}{1.12}}$ & 48.20$_{\scalebox{0.6}{1.65}}$ & 47.97$_{\scalebox{0.6}{0.52}}$ & 45.13$_{\scalebox{0.6}{3.12}}$ & \cellcolor{tableblue}47.83$_{\scalebox{0.6}{1.59}}$ \\
& + \ours & \textbf{64.67}$_{\scalebox{0.6}{2.80}}$ & \textbf{66.90}$_{\scalebox{0.6}{0.42}}$ & \textbf{66.38}$_{\scalebox{0.6}{1.74}}$ & \textbf{60.75}$_{\scalebox{0.6}{3.06}}$ & \textbf{56.52}$_{\scalebox{0.6}{4.03}}$ & \cellcolor{tableblue}\textbf{63.04}$_{\scalebox{0.6}{4.37}}$ & \textbf{51.28}$_{\scalebox{0.6}{0.76}}$ & 48.03$_{\scalebox{0.6}{1.76}}$ & \textbf{48.61}$_{\scalebox{0.6}{1.40}}$ & \textbf{52.03}$_{\scalebox{0.6}{1.01}}$ & \textbf{52.78}$_{\scalebox{0.6}{4.17}}$ & \cellcolor{tableblue}\textbf{50.55}$_{\scalebox{0.6}{2.11}}$ \\
\cmidrule(lr){2-14}
& SeqKD & 56.49$_{\scalebox{0.6}{3.01}}$ & 57.16$_{\scalebox{0.6}{2.91}}$ & 59.51$_{\scalebox{0.6}{3.20}}$ & 60.46$_{\scalebox{0.6}{2.70}}$ & 62.09$_{\scalebox{0.6}{1.90}}$ & \cellcolor{tableblue}59.14$_{\scalebox{0.6}{2.32}}$ & 40.23$_{\scalebox{0.6}{1.47}}$ & 41.97$_{\scalebox{0.6}{1.21}}$ & 47.30$_{\scalebox{0.6}{0.72}}$ & 48.78$_{\scalebox{0.6}{1.35}}$ & 53.30$_{\scalebox{0.6}{1.05}}$ & \cellcolor{tableblue}46.32$_{\scalebox{0.6}{5.29}}$ \\
& + \ours & \textbf{64.78}$_{\scalebox{0.6}{1.38}}$ & \textbf{66.29}$_{\scalebox{0.6}{1.75}}$ & \textbf{66.29}$_{\scalebox{0.6}{1.35}}$ & \textbf{68.87}$_{\scalebox{0.6}{1.16}}$ & \textbf{66.03}$_{\scalebox{0.6}{2.30}}$ & \cellcolor{tableblue}\textbf{66.45}$_{\scalebox{0.6}{1.49}}$ & \textbf{41.28}$_{\scalebox{0.6}{1.60}}$ & \textbf{46.49}$_{\scalebox{0.6}{1.11}}$ & \textbf{48.35}$_{\scalebox{0.6}{0.85}}$ & \textbf{51.10}$_{\scalebox{0.6}{1.83}}$ & \textbf{53.94}$_{\scalebox{0.6}{1.74}}$ & \cellcolor{tableblue}\textbf{48.23}$_{\scalebox{0.6}{4.80}}$ \\
\cmidrule(lr){1-14}

\multirow{6}{*}{TriviaQA} 
& KD & 56.44$_{\scalebox{0.6}{3.22}}$ & 53.71$_{\scalebox{0.6}{2.02}}$ & 49.56$_{\scalebox{0.6}{3.44}}$ & 49.46$_{\scalebox{0.6}{2.68}}$ & 50.35$_{\scalebox{0.6}{1.12}}$ & \cellcolor{tableblue}51.90$_{\scalebox{0.6}{3.07}}$ & \textbf{63.26}$_{\scalebox{0.6}{1.05}}$ & \textbf{64.15}$_{\scalebox{0.6}{1.19}}$ & 62.57$_{\scalebox{0.6}{0.60}}$ & 61.25$_{\scalebox{0.6}{1.79}}$ & \textbf{63.39}$_{\scalebox{0.6}{1.28}}$ & \cellcolor{tableblue}\textbf{62.93}$_{\scalebox{0.6}{1.09}}$ \\
& + \ours & \textbf{62.24}$_{\scalebox{0.6}{0.99}}$ & \textbf{59.90}$_{\scalebox{0.6}{0.68}}$ & \textbf{59.31}$_{\scalebox{0.6}{6.57}}$ & \textbf{52.62}$_{\scalebox{0.6}{2.03}}$ & \textbf{53.15}$_{\scalebox{0.6}{2.57}}$ & \cellcolor{tableblue}\textbf{57.44}$_{\scalebox{0.6}{4.31}}$ & 62.93$_{\scalebox{0.6}{0.78}}$ & 61.91$_{\scalebox{0.6}{0.41}}$ & \textbf{64.75}$_{\scalebox{0.6}{0.79}}$ & \textbf{63.76}$_{\scalebox{0.6}{1.02}}$ & 60.76$_{\scalebox{0.6}{2.52}}$ & \cellcolor{tableblue}62.82$_{\scalebox{0.6}{1.56}}$ \\
\cmidrule(lr){2-14}
& LWD & 55.29$_{\scalebox{0.6}{3.79}}$ & 54.20$_{\scalebox{0.6}{2.68}}$ & 47.64$_{\scalebox{0.6}{1.98}}$ & 48.60$_{\scalebox{0.6}{1.94}}$ & 49.75$_{\scalebox{0.6}{1.24}}$ & \cellcolor{tableblue}51.10$_{\scalebox{0.6}{3.43}}$ & 62.11$_{\scalebox{0.6}{1.30}}$ & \textbf{64.38}$_{\scalebox{0.6}{0.96}}$ & 63.30$_{\scalebox{0.6}{1.03}}$ & 61.48$_{\scalebox{0.6}{0.73}}$ & \textbf{62.31}$_{\scalebox{0.6}{2.00}}$ & \cellcolor{tableblue}62.72$_{\scalebox{0.6}{1.14}}$ \\
& + \ours & \textbf{61.05}$_{\scalebox{0.6}{1.30}}$ & \textbf{58.42}$_{\scalebox{0.6}{2.54}}$ & \textbf{58.25}$_{\scalebox{0.6}{7.48}}$ & \textbf{52.78}$_{\scalebox{0.6}{1.70}}$ & \textbf{52.88}$_{\scalebox{0.6}{1.59}}$ & \cellcolor{tableblue}\textbf{56.68}$_{\scalebox{0.6}{3.68}}$ & \textbf{62.80}$_{\scalebox{0.6}{1.00}}$ & 62.67$_{\scalebox{0.6}{0.61}}$ & \textbf{64.25}$_{\scalebox{0.6}{0.69}}$ & \textbf{63.53}$_{\scalebox{0.6}{1.05}}$ & 61.29$_{\scalebox{0.6}{1.53}}$ & \cellcolor{tableblue}\textbf{62.91}$_{\scalebox{0.6}{1.11}}$ \\
\cmidrule(lr){2-14}
& SeqKD & 57.83$_{\scalebox{0.6}{3.70}}$ & 51.83$_{\scalebox{0.6}{2.61}}$ & 46.92$_{\scalebox{0.6}{2.27}}$ & 46.85$_{\scalebox{0.6}{2.66}}$ & 44.48$_{\scalebox{0.6}{1.74}}$ & \cellcolor{tableblue}49.58$_{\scalebox{0.6}{5.33}}$ & 63.30$_{\scalebox{0.6}{1.11}}$ & 61.78$_{\scalebox{0.6}{1.17}}$ & 63.26$_{\scalebox{0.6}{1.49}}$ & 62.47$_{\scalebox{0.6}{0.97}}$ & \textbf{65.31}$_{\scalebox{0.6}{0.86}}$ & \cellcolor{tableblue}63.22$_{\scalebox{0.6}{1.32}}$ \\
& + \ours & \textbf{60.46}$_{\scalebox{0.6}{1.49}}$ & \textbf{53.08}$_{\scalebox{0.6}{2.02}}$ & \textbf{52.32}$_{\scalebox{0.6}{2.40}}$ & \textbf{47.71}$_{\scalebox{0.6}{3.82}}$ & \textbf{48.90}$_{\scalebox{0.6}{1.86}}$ & \cellcolor{tableblue}\textbf{52.49}$_{\scalebox{0.6}{4.99}}$ & \textbf{63.72}$_{\scalebox{0.6}{0.50}}$ & \textbf{64.45}$_{\scalebox{0.6}{0.86}}$ & \textbf{64.55}$_{\scalebox{0.6}{0.80}}$ & \textbf{63.76}$_{\scalebox{0.6}{1.30}}$ & 63.06$_{\scalebox{0.6}{1.45}}$ & \cellcolor{tableblue}\textbf{63.91}$_{\scalebox{0.6}{0.61}}$ \\
\cmidrule(lr){1-14}

\multirow{6}{*}{HotpotQA} 
& KD & 56.97$_{\scalebox{0.6}{4.68}}$ & 58.65$_{\scalebox{0.6}{7.33}}$ & 42.35$_{\scalebox{0.6}{7.99}}$ & 53.01$_{\scalebox{0.6}{4.67}}$ & 54.89$_{\scalebox{0.6}{6.91}}$ & \cellcolor{tableblue}53.17$_{\scalebox{0.6}{6.41}}$ & \textbf{64.74}$_{\scalebox{0.6}{0.39}}$ & \textbf{66.06}$_{\scalebox{0.6}{0.32}}$ & 62.80$_{\scalebox{0.6}{1.31}}$ & 63.13$_{\scalebox{0.6}{0.77}}$ & 63.70$_{\scalebox{0.6}{1.31}}$ & \cellcolor{tableblue}64.09$_{\scalebox{0.6}{1.33}}$ \\
& + \ours & \textbf{60.36}$_{\scalebox{0.6}{4.46}}$ & \textbf{58.95}$_{\scalebox{0.6}{5.76}}$ & \textbf{57.54}$_{\scalebox{0.6}{1.95}}$ & \textbf{53.81}$_{\scalebox{0.6}{4.75}}$ & \textbf{61.99}$_{\scalebox{0.6}{3.86}}$ & \cellcolor{tableblue}\textbf{58.53}$_{\scalebox{0.6}{3.11}}$ & 64.05$_{\scalebox{0.6}{1.24}}$ & 65.77$_{\scalebox{0.6}{0.39}}$ & \textbf{66.67}$_{\scalebox{0.6}{0.79}}$ & \textbf{66.18}$_{\scalebox{0.6}{0.63}}$ & \textbf{66.34}$_{\scalebox{0.6}{0.69}}$ & \cellcolor{tableblue}\textbf{65.80}$_{\scalebox{0.6}{1.03}}$ \\
\cmidrule(lr){2-14}
& LWD & 53.63$_{\scalebox{0.6}{5.91}}$ & 55.49$_{\scalebox{0.6}{8.74}}$ & 38.30$_{\scalebox{0.6}{6.33}}$ & 52.32$_{\scalebox{0.6}{5.27}}$ & 54.96$_{\scalebox{0.6}{5.91}}$ & \cellcolor{tableblue}50.94$_{\scalebox{0.6}{7.17}}$ & \textbf{65.57}$_{\scalebox{0.6}{0.56}}$ & 65.70$_{\scalebox{0.6}{0.78}}$ & 64.05$_{\scalebox{0.6}{0.46}}$ & 65.05$_{\scalebox{0.6}{0.58}}$ & 65.60$_{\scalebox{0.6}{2.06}}$ & \cellcolor{tableblue}65.19$_{\scalebox{0.6}{0.69}}$ \\
& + \ours & \textbf{54.04}$_{\scalebox{0.6}{9.44}}$ & \textbf{62.83}$_{\scalebox{0.6}{4.28}}$ & \textbf{58.02}$_{\scalebox{0.6}{2.00}}$ & \textbf{54.38}$_{\scalebox{0.6}{4.91}}$ & \textbf{60.44}$_{\scalebox{0.6}{4.15}}$ & \cellcolor{tableblue}\textbf{57.94}$_{\scalebox{0.6}{3.81}}$ & 65.49$_{\scalebox{0.6}{1.12}}$ & \textbf{65.99}$_{\scalebox{0.6}{0.60}}$ & \textbf{66.26}$_{\scalebox{0.6}{0.61}}$ & \textbf{66.71}$_{\scalebox{0.6}{1.14}}$ & \textbf{67.86}$_{\scalebox{0.6}{0.48}}$ & \cellcolor{tableblue}\textbf{66.46}$_{\scalebox{0.6}{0.90}}$ \\
\cmidrule(lr){2-14}
& SeqKD & \textbf{67.39}$_{\scalebox{0.6}{0.69}}$ & 66.15$_{\scalebox{0.6}{0.80}}$ & 66.40$_{\scalebox{0.6}{2.12}}$ & 61.39$_{\scalebox{0.6}{5.83}}$ & \textbf{65.40}$_{\scalebox{0.6}{3.45}}$ & \cellcolor{tableblue}65.35$_{\scalebox{0.6}{2.32}}$ & \textbf{68.09}$_{\scalebox{0.6}{1.04}}$ & \textbf{68.43}$_{\scalebox{0.6}{0.41}}$ & 66.76$_{\scalebox{0.6}{0.83}}$ & 63.96$_{\scalebox{0.6}{0.46}}$ & 65.52$_{\scalebox{0.6}{1.40}}$ & \cellcolor{tableblue}66.55$_{\scalebox{0.6}{1.85}}$ \\
& + \ours & 65.51$_{\scalebox{0.6}{1.38}}$ & \textbf{67.13}$_{\scalebox{0.6}{1.28}}$ & \textbf{67.13}$_{\scalebox{0.6}{1.28}}$ & \textbf{65.78}$_{\scalebox{0.6}{2.88}}$ & 63.59$_{\scalebox{0.6}{4.39}}$ & \cellcolor{tableblue}\textbf{65.82}$_{\scalebox{0.6}{1.46}}$ & 67.04$_{\scalebox{0.6}{0.65}}$ & 67.07$_{\scalebox{0.6}{1.30}}$ & \textbf{68.36}$_{\scalebox{0.6}{0.90}}$ & \textbf{67.78}$_{\scalebox{0.6}{0.63}}$ & \textbf{66.81}$_{\scalebox{0.6}{0.96}}$ & \cellcolor{tableblue}\textbf{67.41}$_{\scalebox{0.6}{0.64}}$ \\
\midrule

% OVERALL AVERAGE ROWS
\rowcolor{tableblue} 
& KD & 63.65$_{\scalebox{0.6}{12.48}}$ & 63.52$_{\scalebox{0.6}{10.85}}$ & 57.50$_{\scalebox{0.6}{14.50}}$ & 57.22$_{\scalebox{0.6}{14.01}}$ & 56.62$_{\scalebox{0.6}{12.70}}$ & \cellcolor{tableblue}59.70$_{\scalebox{0.6}{12.46}}$ & 62.18$_{\scalebox{0.6}{9.22}}$ & 61.99$_{\scalebox{0.6}{10.68}}$ & 60.77$_{\scalebox{0.6}{10.37}}$ & 60.67$_{\scalebox{0.6}{8.65}}$ & 60.88$_{\scalebox{0.6}{10.75}}$ & \cellcolor{tableblue}61.30$_{\scalebox{0.6}{9.90}}$ \\
\rowcolor{tableblue} 
& + \ours & \textbf{66.70}$_{\scalebox{0.6}{10.72}}$ & \textbf{66.22}$_{\scalebox{0.6}{11.15}}$ & \textbf{65.06}$_{\scalebox{0.6}{10.93}}$ & \textbf{61.01}$_{\scalebox{0.6}{13.36}}$ & \textbf{61.57}$_{\scalebox{0.6}{11.60}}$ & \cellcolor{tableblue}\textbf{64.11}$_{\scalebox{0.6}{11.37}}$ & \textbf{62.33}$_{\scalebox{0.6}{7.57}}$ & \textbf{62.42}$_{\scalebox{0.6}{8.93}}$ & \textbf{63.33}$_{\scalebox{0.6}{9.07}}$ & \textbf{62.78}$_{\scalebox{0.6}{9.10}}$ & \textbf{62.45}$_{\scalebox{0.6}{7.70}}$ & \cellcolor{tableblue}\textbf{62.66}$_{\scalebox{0.6}{8.41}}$ \\
\cmidrule(lr){2-14}
\rowcolor{tableblue} 
& LWD & 63.51$_{\scalebox{0.6}{12.72}}$ & 62.63$_{\scalebox{0.6}{11.18}}$ & 55.63$_{\scalebox{0.6}{15.46}}$ & 56.20$_{\scalebox{0.6}{12.81}}$ & 56.56$_{\scalebox{0.6}{13.47}}$ & \cellcolor{tableblue}58.91$_{\scalebox{0.6}{12.65}}$ & 61.75$_{\scalebox{0.6}{8.79}}$ & \textbf{63.00}$_{\scalebox{0.6}{8.46}}$ & 62.10$_{\scalebox{0.6}{8.67}}$ & 61.51$_{\scalebox{0.6}{8.34}}$ & 61.59$_{\scalebox{0.6}{10.04}}$ & \cellcolor{tableblue}61.99$_{\scalebox{0.6}{8.83}}$ \\
\rowcolor{tableblue} 
& + \ours & \textbf{65.41}$_{\scalebox{0.6}{12.13}}$ & \textbf{66.29}$_{\scalebox{0.6}{10.66}}$ & \textbf{64.47}$_{\scalebox{0.6}{11.12}}$ & \textbf{61.22}$_{\scalebox{0.6}{13.22}}$ & \textbf{60.95}$_{\scalebox{0.6}{12.28}}$ & \cellcolor{tableblue}\textbf{63.67}$_{\scalebox{0.6}{11.65}}$ & \textbf{62.14}$_{\scalebox{0.6}{7.78}}$ & 62.75$_{\scalebox{0.6}{9.18}}$ & \textbf{63.54}$_{\scalebox{0.6}{9.17}}$ & \textbf{64.03}$_{\scalebox{0.6}{7.47}}$ & \textbf{64.07}$_{\scalebox{0.6}{7.58}}$ & \cellcolor{tableblue}\textbf{63.30}$_{\scalebox{0.6}{8.14}}$ \\
\cmidrule(lr){2-14}
\rowcolor{tableblue} 
& SeqKD & 64.54$_{\scalebox{0.6}{11.47}}$ & 63.46$_{\scalebox{0.6}{13.09}}$ & 63.21$_{\scalebox{0.6}{13.57}}$ & 62.44$_{\scalebox{0.6}{13.69}}$ & 63.79$_{\scalebox{0.6}{14.25}}$ & \cellcolor{tableblue}63.49$_{\scalebox{0.6}{12.94}}$ & 61.58$_{\scalebox{0.6}{13.01}}$ & 62.32$_{\scalebox{0.6}{12.34}}$ & 63.29$_{\scalebox{0.6}{9.84}}$ & 62.99$_{\scalebox{0.6}{9.09}}$ & 65.06$_{\scalebox{0.6}{7.69}}$ & \cellcolor{tableblue}63.05$_{\scalebox{0.6}{10.31}}$ \\
\rowcolor{tableblue} 
\multirow{-6}{*}{\textbf{AVG}} & + \ours & \textbf{67.31}$_{\scalebox{0.6}{10.49}}$ & \textbf{66.08}$_{\scalebox{0.6}{11.61}}$ & \textbf{66.09}$_{\scalebox{0.6}{11.88}}$ & \textbf{65.42}$_{\scalebox{0.6}{13.37}}$ & \textbf{64.53}$_{\scalebox{0.6}{12.36}}$ & \cellcolor{tableblue}\textbf{65.89}$_{\scalebox{0.6}{11.77}}$ & \textbf{61.66}$_{\scalebox{0.6}{12.22}}$ & \textbf{63.22}$_{\scalebox{0.6}{10.43}}$ & \textbf{64.45}$_{\scalebox{0.6}{9.77}}$ & \textbf{64.78}$_{\scalebox{0.6}{8.57}}$ & \textbf{65.24}$_{\scalebox{0.6}{7.72}}$ & \cellcolor{tableblue}\textbf{63.87}$_{\scalebox{0.6}{9.67}}$ \\
\bottomrule
\end{tabular}%
}
\end{table*}

\begin{table*}[htbp]
\caption{In-domain teacher-student agreement F1 score comparison across $k$-shots for T5 and Qwen2.5 models using baselines. Mean agreement F1 scores are reported with standard deviations as subscripts ($Mean_{std}$). Bold values indicate the better-performing method (\ours vs. baseline). Blue shading highlights aggregate results.}
\label{tab:indomain_agreement_f1}
\centering
\fontsize{6.5pt}{8.5pt}\selectfont
\setlength{\tabcolsep}{1.0pt}
\renewcommand{\arraystretch}{0.95}

\definecolor{tableblue}{HTML}{E6F0FF}

\resizebox{\textwidth}{!}{%
\begin{tabular}{llcccccccccccccc}
\toprule
 & & \multicolumn{6}{c}{\textbf{T5}} & \multicolumn{6}{c}{\textbf{Qwen2.5}} \\
\cmidrule(lr){3-8} \cmidrule(lr){9-14}
Dataset & Method & 16 & 32 & 64 & 128 & 256 & \cellcolor{tableblue}Avg & 16 & 32 & 64 & 128 & 256 & \cellcolor{tableblue}Avg \\
\midrule

\multirow{6}{*}{NQ} 
& KD & 62.31$_{\scalebox{0.6}{3.15}}$ & 64.50$_{\scalebox{0.6}{1.70}}$ & 61.75$_{\scalebox{0.6}{2.79}}$ & 60.11$_{\scalebox{0.6}{3.06}}$ & 60.98$_{\scalebox{0.6}{2.11}}$ & \cellcolor{tableblue}61.93$_{\scalebox{0.6}{1.66}}$ & \textbf{68.95}$_{\scalebox{0.6}{0.66}}$ & 69.19$_{\scalebox{0.6}{0.31}}$ & 68.87$_{\scalebox{0.6}{0.59}}$ & 68.30$_{\scalebox{0.6}{0.55}}$ & 68.94$_{\scalebox{0.6}{0.85}}$ & \cellcolor{tableblue}68.85$_{\scalebox{0.6}{0.33}}$ \\
& + \ours & \textbf{65.08}$_{\scalebox{0.6}{0.40}}$ & \textbf{64.91}$_{\scalebox{0.6}{0.42}}$ & \textbf{62.65}$_{\scalebox{0.6}{0.70}}$ & \textbf{61.96}$_{\scalebox{0.6}{2.63}}$ & \textbf{61.36}$_{\scalebox{0.6}{2.37}}$ & \cellcolor{tableblue}\textbf{63.19}$_{\scalebox{0.6}{1.71}}$ & 68.18$_{\scalebox{0.6}{0.32}}$ & \textbf{70.29}$_{\scalebox{0.6}{0.31}}$ & \textbf{70.47}$_{\scalebox{0.6}{0.34}}$ & \textbf{70.63}$_{\scalebox{0.6}{0.28}}$ & \textbf{70.45}$_{\scalebox{0.6}{0.76}}$ & \cellcolor{tableblue}\textbf{70.01}$_{\scalebox{0.6}{1.03}}$ \\
\cmidrule(lr){2-14}
& LWD & 64.22$_{\scalebox{0.6}{0.62}}$ & 63.55$_{\scalebox{0.6}{2.69}}$ & 62.55$_{\scalebox{0.6}{1.45}}$ & 60.32$_{\scalebox{0.6}{2.31}}$ & 61.12$_{\scalebox{0.6}{2.68}}$ & \cellcolor{tableblue}62.35$_{\scalebox{0.6}{1.63}}$ & \textbf{67.54}$_{\scalebox{0.6}{0.52}}$ & 69.09$_{\scalebox{0.6}{1.01}}$ & 69.07$_{\scalebox{0.6}{1.11}}$ & 68.78$_{\scalebox{0.6}{0.64}}$ & 69.19$_{\scalebox{0.6}{0.52}}$ & \cellcolor{tableblue}68.73$_{\scalebox{0.6}{0.68}}$ \\
& + \ours & \textbf{65.11}$_{\scalebox{0.6}{0.37}}$ & \textbf{63.84}$_{\scalebox{0.6}{2.50}}$ & \textbf{62.65}$_{\scalebox{0.6}{2.03}}$ & \textbf{62.31}$_{\scalebox{0.6}{1.45}}$ & \textbf{61.58}$_{\scalebox{0.6}{3.01}}$ & \cellcolor{tableblue}\textbf{63.10}$_{\scalebox{0.6}{1.39}}$ & 66.10$_{\scalebox{0.6}{0.51}}$ & \textbf{71.11}$_{\scalebox{0.6}{0.54}}$ & \textbf{70.93}$_{\scalebox{0.6}{0.65}}$ & \textbf{71.54}$_{\scalebox{0.6}{0.24}}$ & \textbf{71.35}$_{\scalebox{0.6}{0.52}}$ & \cellcolor{tableblue}\textbf{70.21}$_{\scalebox{0.6}{2.30}}$ \\
\cmidrule(lr){2-14}
& SeqKD & 63.33$_{\scalebox{0.6}{1.54}}$ & 62.02$_{\scalebox{0.6}{1.16}}$ & 63.64$_{\scalebox{0.6}{1.50}}$ & 62.60$_{\scalebox{0.6}{1.83}}$ & \textbf{65.66}$_{\scalebox{0.6}{1.78}}$ & \cellcolor{tableblue}63.45$_{\scalebox{0.6}{1.39}}$ & 68.48$_{\scalebox{0.6}{0.35}}$ & \textbf{71.63}$_{\scalebox{0.6}{0.55}}$ & 70.87$_{\scalebox{0.6}{0.41}}$ & 71.58$_{\scalebox{0.6}{0.11}}$ & 71.85$_{\scalebox{0.6}{0.78}}$ & \cellcolor{tableblue}70.88$_{\scalebox{0.6}{1.39}}$ \\
& + \ours & \textbf{64.85}$_{\scalebox{0.6}{0.67}}$ & \textbf{64.23}$_{\scalebox{0.6}{0.93}}$ & \textbf{64.23}$_{\scalebox{0.6}{0.78}}$ & \textbf{63.71}$_{\scalebox{0.6}{1.28}}$ & 64.05$_{\scalebox{0.6}{1.47}}$ & \cellcolor{tableblue}\textbf{64.21}$_{\scalebox{0.6}{0.42}}$ & \textbf{69.59}$_{\scalebox{0.6}{0.42}}$ & 70.19$_{\scalebox{0.6}{0.45}}$ & \textbf{72.23}$_{\scalebox{0.6}{0.57}}$ & \textbf{72.15}$_{\scalebox{0.6}{0.50}}$ & \textbf{72.30}$_{\scalebox{0.6}{0.54}}$ & \cellcolor{tableblue}\textbf{71.29}$_{\scalebox{0.6}{1.30}}$ \\
\cmidrule(lr){1-14}

\multirow{6}{*}{SQuAD} 
& KD & 90.29$_{\scalebox{0.6}{0.71}}$ & 87.73$_{\scalebox{0.6}{0.93}}$ & 87.20$_{\scalebox{0.6}{1.46}}$ & 88.17$_{\scalebox{0.6}{1.55}}$ & 85.90$_{\scalebox{0.6}{3.23}}$ & \cellcolor{tableblue}87.86$_{\scalebox{0.6}{1.60}}$ & \textbf{79.82}$_{\scalebox{0.6}{0.28}}$ & 79.14$_{\scalebox{0.6}{0.08}}$ & 78.58$_{\scalebox{0.6}{0.53}}$ & 78.14$_{\scalebox{0.6}{0.70}}$ & \textbf{78.64}$_{\scalebox{0.6}{0.43}}$ & \cellcolor{tableblue}78.86$_{\scalebox{0.6}{0.64}}$ \\
& + \ours & \textbf{90.45}$_{\scalebox{0.6}{0.25}}$ & \textbf{90.43}$_{\scalebox{0.6}{0.40}}$ & \textbf{88.88}$_{\scalebox{0.6}{0.85}}$ & \textbf{89.13}$_{\scalebox{0.6}{0.39}}$ & \textbf{87.34}$_{\scalebox{0.6}{1.48}}$ & \cellcolor{tableblue}\textbf{89.25}$_{\scalebox{0.6}{1.29}}$ & 79.14$_{\scalebox{0.6}{0.46}}$ & \textbf{79.47}$_{\scalebox{0.6}{0.57}}$ & \textbf{79.45}$_{\scalebox{0.6}{0.58}}$ & \textbf{78.70}$_{\scalebox{0.6}{0.69}}$ & 78.22$_{\scalebox{0.6}{0.45}}$ & \cellcolor{tableblue}\textbf{79.00}$_{\scalebox{0.6}{0.53}}$ \\
\cmidrule(lr){2-14}
& LWD & 90.16$_{\scalebox{0.6}{0.61}}$ & 87.72$_{\scalebox{0.6}{1.35}}$ & 87.32$_{\scalebox{0.6}{2.34}}$ & 86.30$_{\scalebox{0.6}{1.00}}$ & 85.98$_{\scalebox{0.6}{1.50}}$ & \cellcolor{tableblue}87.50$_{\scalebox{0.6}{1.65}}$ & \textbf{79.02}$_{\scalebox{0.6}{0.53}}$ & 79.18$_{\scalebox{0.6}{0.26}}$ & 78.92$_{\scalebox{0.6}{0.30}}$ & 78.33$_{\scalebox{0.6}{0.99}}$ & 79.56$_{\scalebox{0.6}{1.05}}$ & \cellcolor{tableblue}79.00$_{\scalebox{0.6}{0.45}}$ \\
& + \ours & \textbf{90.77}$_{\scalebox{0.6}{0.30}}$ & \textbf{89.78}$_{\scalebox{0.6}{0.79}}$ & \textbf{88.64}$_{\scalebox{0.6}{0.83}}$ & \textbf{89.37}$_{\scalebox{0.6}{0.57}}$ & \textbf{87.92}$_{\scalebox{0.6}{1.87}}$ & \cellcolor{tableblue}\textbf{89.30}$_{\scalebox{0.6}{1.09}}$ & 78.79$_{\scalebox{0.6}{0.21}}$ & \textbf{79.72}$_{\scalebox{0.6}{0.27}}$ & \textbf{80.29}$_{\scalebox{0.6}{0.39}}$ & \textbf{79.43}$_{\scalebox{0.6}{0.28}}$ & \textbf{80.14}$_{\scalebox{0.6}{1.43}}$ & \cellcolor{tableblue}\textbf{79.67}$_{\scalebox{0.6}{0.60}}$ \\
\cmidrule(lr){2-14}
& SeqKD & 89.30$_{\scalebox{0.6}{0.97}}$ & \textbf{90.12}$_{\scalebox{0.6}{0.85}}$ & \textbf{89.54}$_{\scalebox{0.6}{0.67}}$ & 89.51$_{\scalebox{0.6}{0.20}}$ & \textbf{89.33}$_{\scalebox{0.6}{0.28}}$ & \cellcolor{tableblue}\textbf{89.56}$_{\scalebox{0.6}{0.33}}$ & \textbf{80.78}$_{\scalebox{0.6}{0.43}}$ & 80.21$_{\scalebox{0.6}{0.32}}$ & 80.05$_{\scalebox{0.6}{0.73}}$ & 79.79$_{\scalebox{0.6}{0.50}}$ & 80.67$_{\scalebox{0.6}{1.04}}$ & \cellcolor{tableblue}80.30$_{\scalebox{0.6}{0.42}}$ \\
& + \ours & \textbf{90.62}$_{\scalebox{0.6}{0.50}}$ & 89.30$_{\scalebox{0.6}{0.51}}$ & 89.43$_{\scalebox{0.6}{0.66}}$ & \textbf{89.63}$_{\scalebox{0.6}{0.45}}$ & 88.75$_{\scalebox{0.6}{0.71}}$ & \cellcolor{tableblue}89.55$_{\scalebox{0.6}{0.68}}$ & 79.77$_{\scalebox{0.6}{0.44}}$ & \textbf{80.53}$_{\scalebox{0.6}{0.29}}$ & \textbf{80.45}$_{\scalebox{0.6}{0.29}}$ & \textbf{80.31}$_{\scalebox{0.6}{0.42}}$ & \textbf{81.22}$_{\scalebox{0.6}{0.64}}$ & \cellcolor{tableblue}\textbf{80.46}$_{\scalebox{0.6}{0.52}}$ \\
\cmidrule(lr){1-14}

\multirow{6}{*}{NewsQA} 
& KD & 72.06$_{\scalebox{0.6}{1.88}}$ & 72.84$_{\scalebox{0.6}{2.12}}$ & 71.26$_{\scalebox{0.6}{1.39}}$ & 67.68$_{\scalebox{0.6}{2.77}}$ & 64.94$_{\scalebox{0.6}{1.45}}$ & \cellcolor{tableblue}69.76$_{\scalebox{0.6}{3.34}}$ & 58.69$_{\scalebox{0.6}{1.01}}$ & 56.24$_{\scalebox{0.6}{1.74}}$ & 56.72$_{\scalebox{0.6}{1.78}}$ & 57.30$_{\scalebox{0.6}{1.50}}$ & 54.54$_{\scalebox{0.6}{1.80}}$ & \cellcolor{tableblue}56.70$_{\scalebox{0.6}{1.52}}$ \\
& + \ours & \textbf{73.00}$_{\scalebox{0.6}{2.11}}$ & \textbf{73.97}$_{\scalebox{0.6}{0.76}}$ & \textbf{75.48}$_{\scalebox{0.6}{1.49}}$ & \textbf{73.08}$_{\scalebox{0.6}{2.41}}$ & \textbf{72.80}$_{\scalebox{0.6}{1.84}}$ & \cellcolor{tableblue}\textbf{73.67}$_{\scalebox{0.6}{1.11}}$ & \textbf{61.02}$_{\scalebox{0.6}{0.67}}$ & \textbf{58.95}$_{\scalebox{0.6}{1.08}}$ & \textbf{59.33}$_{\scalebox{0.6}{0.52}}$ & \textbf{59.99}$_{\scalebox{0.6}{0.98}}$ & \textbf{61.68}$_{\scalebox{0.6}{2.02}}$ & \cellcolor{tableblue}\textbf{60.19}$_{\scalebox{0.6}{1.14}}$ \\
\cmidrule(lr){2-14}
& LWD & 72.44$_{\scalebox{0.6}{2.55}}$ & 72.60$_{\scalebox{0.6}{2.24}}$ & 69.24$_{\scalebox{0.6}{1.92}}$ & 67.64$_{\scalebox{0.6}{2.49}}$ & 63.58$_{\scalebox{0.6}{1.11}}$ & \cellcolor{tableblue}69.10$_{\scalebox{0.6}{3.74}}$ & 60.00$_{\scalebox{0.6}{1.41}}$ & \textbf{61.23}$_{\scalebox{0.6}{0.59}}$ & \textbf{60.02}$_{\scalebox{0.6}{1.51}}$ & 59.98$_{\scalebox{0.6}{1.16}}$ & 56.85$_{\scalebox{0.6}{2.27}}$ & \cellcolor{tableblue}59.61$_{\scalebox{0.6}{1.64}}$ \\
& + \ours & \textbf{73.94}$_{\scalebox{0.6}{2.16}}$ & \textbf{74.73}$_{\scalebox{0.6}{0.65}}$ & \textbf{74.81}$_{\scalebox{0.6}{1.57}}$ & \textbf{72.80}$_{\scalebox{0.6}{1.39}}$ & \textbf{71.65}$_{\scalebox{0.6}{1.83}}$ & \cellcolor{tableblue}\textbf{73.59}$_{\scalebox{0.6}{1.35}}$ & \textbf{61.54}$_{\scalebox{0.6}{0.86}}$ & 58.89$_{\scalebox{0.6}{1.36}}$ & 59.97$_{\scalebox{0.6}{1.08}}$ & \textbf{62.52}$_{\scalebox{0.6}{1.07}}$ & \textbf{63.61}$_{\scalebox{0.6}{4.34}}$ & \cellcolor{tableblue}\textbf{61.31}$_{\scalebox{0.6}{1.90}}$ \\
\cmidrule(lr){2-14}
& SeqKD & 68.57$_{\scalebox{0.6}{1.99}}$ & 70.94$_{\scalebox{0.6}{1.79}}$ & 71.07$_{\scalebox{0.6}{2.51}}$ & 69.37$_{\scalebox{0.6}{1.92}}$ & 72.49$_{\scalebox{0.6}{1.46}}$ & \cellcolor{tableblue}70.49$_{\scalebox{0.6}{1.54}}$ & 48.96$_{\scalebox{0.6}{1.45}}$ & 51.49$_{\scalebox{0.6}{0.86}}$ & \textbf{59.39}$_{\scalebox{0.6}{0.35}}$ & 60.59$_{\scalebox{0.6}{1.39}}$ & 63.64$_{\scalebox{0.6}{0.99}}$ & \cellcolor{tableblue}56.81$_{\scalebox{0.6}{6.28}}$ \\
& + \ours & \textbf{74.01}$_{\scalebox{0.6}{0.97}}$ & \textbf{75.27}$_{\scalebox{0.6}{1.16}}$ & \textbf{76.35}$_{\scalebox{0.6}{1.83}}$ & \textbf{76.27}$_{\scalebox{0.6}{1.39}}$ & \textbf{74.78}$_{\scalebox{0.6}{1.61}}$ & \cellcolor{tableblue}\textbf{75.33}$_{\scalebox{0.6}{1.00}}$ & \textbf{51.15}$_{\scalebox{0.6}{1.39}}$ & \textbf{57.40}$_{\scalebox{0.6}{0.32}}$ & 58.42$_{\scalebox{0.6}{1.31}}$ & \textbf{61.76}$_{\scalebox{0.6}{1.65}}$ & \textbf{64.65}$_{\scalebox{0.6}{1.69}}$ & \cellcolor{tableblue}\textbf{58.67}$_{\scalebox{0.6}{5.09}}$ \\
\cmidrule(lr){1-14}

\multirow{6}{*}{TriviaQA} 
& KD & 59.24$_{\scalebox{0.6}{3.43}}$ & 55.96$_{\scalebox{0.6}{2.19}}$ & 51.30$_{\scalebox{0.6}{3.58}}$ & 51.04$_{\scalebox{0.6}{2.70}}$ & 51.79$_{\scalebox{0.6}{1.05}}$ & \cellcolor{tableblue}53.87$_{\scalebox{0.6}{3.61}}$ & 65.52$_{\scalebox{0.6}{1.01}}$ & \textbf{66.50}$_{\scalebox{0.6}{1.00}}$ & 64.89$_{\scalebox{0.6}{0.54}}$ & 63.97$_{\scalebox{0.6}{1.56}}$ & \textbf{66.34}$_{\scalebox{0.6}{1.09}}$ & \cellcolor{tableblue}\textbf{65.44}$_{\scalebox{0.6}{1.05}}$ \\
& + \ours & \textbf{65.44}$_{\scalebox{0.6}{0.86}}$ & \textbf{62.68}$_{\scalebox{0.6}{0.75}}$ & \textbf{61.91}$_{\scalebox{0.6}{6.87}}$ & \textbf{54.53}$_{\scalebox{0.6}{1.97}}$ & \textbf{54.98}$_{\scalebox{0.6}{2.30}}$ & \cellcolor{tableblue}\textbf{59.91}$_{\scalebox{0.6}{4.89}}$ & \textbf{65.54}$_{\scalebox{0.6}{0.55}}$ & 64.65$_{\scalebox{0.6}{0.42}}$ & \textbf{67.33}$_{\scalebox{0.6}{0.66}}$ & \textbf{66.12}$_{\scalebox{0.6}{0.89}}$ & 63.44$_{\scalebox{0.6}{2.38}}$ & \cellcolor{tableblue}65.41$_{\scalebox{0.6}{1.47}}$ \\
\cmidrule(lr){2-14}
& LWD & 58.13$_{\scalebox{0.6}{4.26}}$ & 56.47$_{\scalebox{0.6}{2.80}}$ & 49.35$_{\scalebox{0.6}{2.13}}$ & 50.19$_{\scalebox{0.6}{2.02}}$ & 51.13$_{\scalebox{0.6}{1.31}}$ & \cellcolor{tableblue}53.05$_{\scalebox{0.6}{3.97}}$ & 64.93$_{\scalebox{0.6}{1.23}}$ & \textbf{66.71}$_{\scalebox{0.6}{0.93}}$ & 65.66$_{\scalebox{0.6}{0.93}}$ & 64.23$_{\scalebox{0.6}{0.69}}$ & \textbf{65.40}$_{\scalebox{0.6}{1.85}}$ & \cellcolor{tableblue}65.39$_{\scalebox{0.6}{0.92}}$ \\
& + \ours & \textbf{64.15}$_{\scalebox{0.6}{1.26}}$ & \textbf{61.39}$_{\scalebox{0.6}{2.69}}$ & \textbf{60.84}$_{\scalebox{0.6}{7.86}}$ & \textbf{54.74}$_{\scalebox{0.6}{1.74}}$ & \textbf{54.69}$_{\scalebox{0.6}{1.53}}$ & \cellcolor{tableblue}\textbf{59.16}$_{\scalebox{0.6}{4.25}}$ & \textbf{65.29}$_{\scalebox{0.6}{1.07}}$ & 65.22$_{\scalebox{0.6}{0.76}}$ & \textbf{66.85}$_{\scalebox{0.6}{0.73}}$ & \textbf{65.93}$_{\scalebox{0.6}{0.86}}$ & 64.03$_{\scalebox{0.6}{1.46}}$ & \cellcolor{tableblue}\textbf{65.47}$_{\scalebox{0.6}{1.04}}$ \\
\cmidrule(lr){2-14}
& SeqKD & 64.78$_{\scalebox{0.6}{3.94}}$ & 57.82$_{\scalebox{0.6}{2.78}}$ & 52.95$_{\scalebox{0.6}{2.61}}$ & 53.12$_{\scalebox{0.6}{2.81}}$ & 50.10$_{\scalebox{0.6}{1.85}}$ & \cellcolor{tableblue}55.75$_{\scalebox{0.6}{5.76}}$ & 65.54$_{\scalebox{0.6}{1.08}}$ & 64.28$_{\scalebox{0.6}{1.20}}$ & 65.49$_{\scalebox{0.6}{1.09}}$ & 65.37$_{\scalebox{0.6}{1.11}}$ & \textbf{68.03}$_{\scalebox{0.6}{0.88}}$ & \cellcolor{tableblue}65.74$_{\scalebox{0.6}{1.38}}$ \\
& + \ours & \textbf{66.54}$_{\scalebox{0.6}{1.86}}$ & \textbf{59.62}$_{\scalebox{0.6}{2.26}}$ & \textbf{58.50}$_{\scalebox{0.6}{2.34}}$ & \textbf{54.02}$_{\scalebox{0.6}{3.93}}$ & \textbf{54.05}$_{\scalebox{0.6}{1.96}}$ & \cellcolor{tableblue}\textbf{58.55}$_{\scalebox{0.6}{5.14}}$ & \textbf{66.13}$_{\scalebox{0.6}{0.46}}$ & \textbf{66.65}$_{\scalebox{0.6}{0.95}}$ & \textbf{66.84}$_{\scalebox{0.6}{0.68}}$ & \textbf{66.08}$_{\scalebox{0.6}{1.02}}$ & 65.89$_{\scalebox{0.6}{1.23}}$ & \cellcolor{tableblue}\textbf{66.32}$_{\scalebox{0.6}{0.41}}$ \\
\cmidrule(lr){1-14}

\multirow{6}{*}{HotpotQA} 
& KD & 71.01$_{\scalebox{0.6}{2.73}}$ & 71.11$_{\scalebox{0.6}{4.62}}$ & 65.09$_{\scalebox{0.6}{3.52}}$ & \textbf{69.61}$_{\scalebox{0.6}{1.64}}$ & 68.77$_{\scalebox{0.6}{3.88}}$ & \cellcolor{tableblue}69.12$_{\scalebox{0.6}{2.46}}$ & \textbf{74.13}$_{\scalebox{0.6}{0.47}}$ & \textbf{75.68}$_{\scalebox{0.6}{0.37}}$ & 72.55$_{\scalebox{0.6}{1.07}}$ & 73.18$_{\scalebox{0.6}{0.70}}$ & 73.36$_{\scalebox{0.6}{1.14}}$ & \cellcolor{tableblue}73.78$_{\scalebox{0.6}{1.20}}$ \\
& + \ours & \textbf{71.54}$_{\scalebox{0.6}{2.28}}$ & \textbf{72.14}$_{\scalebox{0.6}{2.83}}$ & \textbf{71.25}$_{\scalebox{0.6}{0.72}}$ & 69.38$_{\scalebox{0.6}{2.28}}$ & \textbf{73.26}$_{\scalebox{0.6}{2.65}}$ & \cellcolor{tableblue}\textbf{71.51}$_{\scalebox{0.6}{1.42}}$ & 73.66$_{\scalebox{0.6}{0.94}}$ & 75.04$_{\scalebox{0.6}{0.38}}$ & \textbf{76.43}$_{\scalebox{0.6}{0.66}}$ & \textbf{75.88}$_{\scalebox{0.6}{0.54}}$ & \textbf{76.24}$_{\scalebox{0.6}{1.15}}$ & \cellcolor{tableblue}\textbf{75.45}$_{\scalebox{0.6}{1.13}}$ \\
\cmidrule(lr){2-14}
& LWD & \textbf{69.13}$_{\scalebox{0.6}{2.52}}$ & 69.97$_{\scalebox{0.6}{4.08}}$ & 62.94$_{\scalebox{0.6}{3.25}}$ & 69.41$_{\scalebox{0.6}{1.96}}$ & 70.02$_{\scalebox{0.6}{3.04}}$ & \cellcolor{tableblue}68.29$_{\scalebox{0.6}{3.02}}$ & 75.34$_{\scalebox{0.6}{0.48}}$ & 75.91$_{\scalebox{0.6}{0.89}}$ & 73.95$_{\scalebox{0.6}{0.52}}$ & 75.46$_{\scalebox{0.6}{0.68}}$ & 75.56$_{\scalebox{0.6}{1.81}}$ & \cellcolor{tableblue}75.24$_{\scalebox{0.6}{0.76}}$ \\
& + \ours & 68.89$_{\scalebox{0.6}{3.68}}$ & \textbf{73.80}$_{\scalebox{0.6}{2.02}}$ & \textbf{71.17}$_{\scalebox{0.6}{1.64}}$ & \textbf{69.83}$_{\scalebox{0.6}{2.23}}$ & \textbf{73.23}$_{\scalebox{0.6}{2.24}}$ & \cellcolor{tableblue}\textbf{71.38}$_{\scalebox{0.6}{2.12}}$ & \textbf{75.37}$_{\scalebox{0.6}{1.05}}$ & \textbf{76.11}$_{\scalebox{0.6}{0.59}}$ & \textbf{76.21}$_{\scalebox{0.6}{0.50}}$ & \textbf{77.18}$_{\scalebox{0.6}{0.96}}$ & \textbf{77.99}$_{\scalebox{0.6}{0.62}}$ & \cellcolor{tableblue}\textbf{76.57}$_{\scalebox{0.6}{1.02}}$ \\
\cmidrule(lr){2-14}
& SeqKD & \textbf{76.56}$_{\scalebox{0.6}{0.65}}$ & 76.06$_{\scalebox{0.6}{1.09}}$ & 75.41$_{\scalebox{0.6}{1.83}}$ & 73.75$_{\scalebox{0.6}{3.42}}$ & \textbf{74.70}$_{\scalebox{0.6}{3.20}}$ & \cellcolor{tableblue}\textbf{75.29}$_{\scalebox{0.6}{1.11}}$ & \textbf{76.23}$_{\scalebox{0.6}{0.70}}$ & \textbf{77.17}$_{\scalebox{0.6}{0.47}}$ & 75.42$_{\scalebox{0.6}{0.59}}$ & 72.53$_{\scalebox{0.6}{0.26}}$ & 74.29$_{\scalebox{0.6}{1.49}}$ & \cellcolor{tableblue}75.13$_{\scalebox{0.6}{1.80}}$ \\
& + \ours & 74.79$_{\scalebox{0.6}{1.16}}$ & \textbf{76.27}$_{\scalebox{0.6}{0.53}}$ & \textbf{76.04}$_{\scalebox{0.6}{0.88}}$ & \textbf{75.07}$_{\scalebox{0.6}{2.49}}$ & 73.29$_{\scalebox{0.6}{3.56}}$ & \cellcolor{tableblue}75.09$_{\scalebox{0.6}{1.19}}$ & 75.24$_{\scalebox{0.6}{0.56}}$ & 75.29$_{\scalebox{0.6}{1.52}}$ & \textbf{77.08}$_{\scalebox{0.6}{0.83}}$ & \textbf{76.61}$_{\scalebox{0.6}{0.52}}$ & \textbf{75.59}$_{\scalebox{0.6}{1.35}}$ & \cellcolor{tableblue}\textbf{75.96}$_{\scalebox{0.6}{0.83}}$ \\
\midrule

% OVERALL AVERAGE ROWS
\rowcolor{tableblue} 
& KD & 70.98$_{\scalebox{0.6}{12.11}}$ & 70.43$_{\scalebox{0.6}{11.73}}$ & 67.32$_{\scalebox{0.6}{13.26}}$ & 67.32$_{\scalebox{0.6}{13.76}}$ & 66.48$_{\scalebox{0.6}{12.56}}$ & \cellcolor{tableblue}68.51$_{\scalebox{0.6}{12.59}}$ & 69.42$_{\scalebox{0.6}{8.07}}$ & 69.35$_{\scalebox{0.6}{8.89}}$ & 68.32$_{\scalebox{0.6}{8.21}}$ & 68.18$_{\scalebox{0.6}{8.07}}$ & 68.36$_{\scalebox{0.6}{9.03}}$ & \cellcolor{tableblue}68.73$_{\scalebox{0.6}{8.42}}$ \\
\rowcolor{tableblue} 
& + \ours & \textbf{73.10}$_{\scalebox{0.6}{10.33}}$ & \textbf{72.83}$_{\scalebox{0.6}{10.92}}$ & \textbf{72.03}$_{\scalebox{0.6}{11.03}}$ & \textbf{69.62}$_{\scalebox{0.6}{13.03}}$ & \textbf{69.95}$_{\scalebox{0.6}{12.44}}$ & \cellcolor{tableblue}\textbf{71.50}$_{\scalebox{0.6}{11.44}}$ & \textbf{69.51}$_{\scalebox{0.6}{7.06}}$ & \textbf{69.68}$_{\scalebox{0.6}{8.15}}$ & \textbf{70.60}$_{\scalebox{0.6}{7.91}}$ & \textbf{70.26}$_{\scalebox{0.6}{7.51}}$ & \textbf{70.01}$_{\scalebox{0.6}{7.40}}$ & \cellcolor{tableblue}\textbf{70.01}$_{\scalebox{0.6}{7.55}}$ \\
\cmidrule(lr){2-14}
\rowcolor{tableblue} 
& LWD & 70.82$_{\scalebox{0.6}{12.09}}$ & 70.06$_{\scalebox{0.6}{11.68}}$ & 66.28$_{\scalebox{0.6}{13.81}}$ & 66.77$_{\scalebox{0.6}{13.28}}$ & 66.36$_{\scalebox{0.6}{12.90}}$ & \cellcolor{tableblue}68.06$_{\scalebox{0.6}{12.62}}$ & 69.36$_{\scalebox{0.6}{7.74}}$ & 70.42$_{\scalebox{0.6}{7.19}}$ & 69.52$_{\scalebox{0.6}{7.30}}$ & 69.35$_{\scalebox{0.6}{7.62}}$ & 69.31$_{\scalebox{0.6}{8.87}}$ & \cellcolor{tableblue}69.60$_{\scalebox{0.6}{7.72}}$ \\
\rowcolor{tableblue} 
& + \ours & \textbf{72.57}$_{\scalebox{0.6}{10.88}}$ & \textbf{72.71}$_{\scalebox{0.6}{11.22}}$ & \textbf{71.62}$_{\scalebox{0.6}{11.15}}$ & \textbf{69.81}$_{\scalebox{0.6}{12.99}}$ & \textbf{69.81}$_{\scalebox{0.6}{12.64}}$ & \cellcolor{tableblue}\textbf{71.31}$_{\scalebox{0.6}{11.66}}$ & \textbf{69.42}$_{\scalebox{0.6}{7.30}}$ & \textbf{70.21}$_{\scalebox{0.6}{8.35}}$ & \textbf{70.85}$_{\scalebox{0.6}{7.94}}$ & \textbf{71.32}$_{\scalebox{0.6}{7.18}}$ & \textbf{71.42}$_{\scalebox{0.6}{7.66}}$ & \cellcolor{tableblue}\textbf{70.64}$_{\scalebox{0.6}{7.60}}$ \\
\cmidrule(lr){2-14}
\rowcolor{tableblue} 
& SeqKD & 72.51$_{\scalebox{0.6}{10.70}}$ & 71.39$_{\scalebox{0.6}{12.69}}$ & 70.52$_{\scalebox{0.6}{13.62}}$ & 69.67$_{\scalebox{0.6}{13.55}}$ & 70.46$_{\scalebox{0.6}{14.28}}$ & \cellcolor{tableblue}70.91$_{\scalebox{0.6}{12.77}}$ & 68.00$_{\scalebox{0.6}{12.24}}$ & 68.96$_{\scalebox{0.6}{11.49}}$ & 70.24$_{\scalebox{0.6}{8.12}}$ & 69.97$_{\scalebox{0.6}{7.33}}$ & 71.70$_{\scalebox{0.6}{6.44}}$ & \cellcolor{tableblue}69.77$_{\scalebox{0.6}{9.01}}$ \\
\rowcolor{tableblue} 
\multirow{-6}{*}{\textbf{AVG}} & + \ours & \textbf{74.16}$_{\scalebox{0.6}{10.20}}$ & \textbf{72.94}$_{\scalebox{0.6}{11.59}}$ & \textbf{72.91}$_{\scalebox{0.6}{12.02}}$ & \textbf{71.74}$_{\scalebox{0.6}{13.51}}$ & \textbf{70.98}$_{\scalebox{0.6}{12.94}}$ & \cellcolor{tableblue}\textbf{72.55}$_{\scalebox{0.6}{11.92}}$ & \textbf{68.38}$_{\scalebox{0.6}{10.96}}$ & \textbf{70.01}$_{\scalebox{0.6}{8.79}}$ & \textbf{71.00}$_{\scalebox{0.6}{8.71}}$ & \textbf{71.38}$_{\scalebox{0.6}{7.55}}$ & \textbf{71.93}$_{\scalebox{0.6}{6.88}}$ & \cellcolor{tableblue}\textbf{70.54}$_{\scalebox{0.6}{8.47}}$ \\
\bottomrule
\end{tabular}%
}
\end{table*}

\newpage
\subsection{Out-of-Domain EM Results} \label{app:ood}

Table~\ref{tab:ood_em_t5_qwen} reports the out-of-domain EM results for T5 and Qwen2.5.

\begin{table*}[htbp]
\caption{Out-of-domain EM score comparison across $k$-shots for T5 and Qwen2.5 models using KD. Mean EM scores are reported with standard deviations as subscripts ($Mean_{std}$). Bold values indicate the better-performing method (\ours vs. KD). Blue shading highlights aggregate results.}
\label{tab:ood_em_t5_qwen}
\centering
\fontsize{7pt}{9pt}\selectfont
\setlength{\tabcolsep}{1.2pt}
\renewcommand{\arraystretch}{0.95}

\definecolor{tableblue}{HTML}{E6F0FF}

\resizebox{\textwidth}{!}{%
\begin{tabular}{llcccccccccccccc}
\toprule
 & & \multicolumn{6}{c}{\textbf{T5}} & \multicolumn{6}{c}{\textbf{Qwen2.5}} \\
\cmidrule(lr){3-8} \cmidrule(lr){9-14}
Dataset & Method & 16 & 32 & 64 & 128 & 256 & \cellcolor{tableblue}Avg & 16 & 32 & 64 & 128 & 256 & \cellcolor{tableblue}Avg \\
\midrule

\multirow{2}{*}{DROP} 
& KD & 34.75$_{\scalebox{0.6}{1.78}}$ & 32.00$_{\scalebox{0.6}{4.64}}$ & 27.56$_{\scalebox{0.6}{3.50}}$ & 22.63$_{\scalebox{0.6}{4.29}}$ & 23.80$_{\scalebox{0.6}{1.97}}$ & \cellcolor{tableblue}28.15$_{\scalebox{0.6}{5.20}}$ & \textbf{44.73}$_{\scalebox{0.6}{0.78}}$ & 39.19$_{\scalebox{0.6}{1.71}}$ & 36.93$_{\scalebox{0.6}{0.33}}$ & 35.03$_{\scalebox{0.6}{0.90}}$ & \textbf{36.57}$_{\scalebox{0.6}{1.04}}$ & \cellcolor{tableblue}38.49$_{\scalebox{0.6}{3.79}}$ \\
& + \ours & \textbf{35.68}$_{\scalebox{0.6}{1.01}}$ & \textbf{33.86}$_{\scalebox{0.6}{1.47}}$ & \textbf{28.36}$_{\scalebox{0.6}{3.54}}$ & \textbf{25.98}$_{\scalebox{0.6}{3.56}}$ & \textbf{24.65}$_{\scalebox{0.6}{2.56}}$ & \cellcolor{tableblue}\textbf{29.71}$_{\scalebox{0.6}{4.85}}$ & 44.04$_{\scalebox{0.6}{0.73}}$ & \textbf{44.32}$_{\scalebox{0.6}{1.35}}$ & \textbf{39.56}$_{\scalebox{0.6}{1.51}}$ & \textbf{37.58}$_{\scalebox{0.6}{1.03}}$ & 35.19$_{\scalebox{0.6}{1.10}}$ & \cellcolor{tableblue}\textbf{40.14}$_{\scalebox{0.6}{4.00}}$ \\
\cmidrule(lr){1-14}

\multirow{2}{*}{DuoRC} 
& KD & 44.63$_{\scalebox{0.6}{1.56}}$ & 43.12$_{\scalebox{0.6}{3.14}}$ & \textbf{41.47}$_{\scalebox{0.6}{1.91}}$ & 38.17$_{\scalebox{0.6}{1.43}}$ & 39.33$_{\scalebox{0.6}{1.84}}$ & \cellcolor{tableblue}41.35$_{\scalebox{0.6}{2.65}}$ & \textbf{33.99}$_{\scalebox{0.6}{0.80}}$ & \textbf{38.53}$_{\scalebox{0.6}{1.06}}$ & 36.66$_{\scalebox{0.6}{1.60}}$ & 36.62$_{\scalebox{0.6}{1.10}}$ & 36.62$_{\scalebox{0.6}{1.80}}$ & \cellcolor{tableblue}36.48$_{\scalebox{0.6}{1.62}}$ \\
& + \ours & \textbf{46.01}$_{\scalebox{0.6}{0.84}}$ & \textbf{44.63}$_{\scalebox{0.6}{0.87}}$ & 39.02$_{\scalebox{0.6}{2.37}}$ & \textbf{40.94}$_{\scalebox{0.6}{1.17}}$ & \textbf{40.04}$_{\scalebox{0.6}{1.08}}$ & \cellcolor{tableblue}\textbf{42.13}$_{\scalebox{0.6}{3.03}}$ & 33.68$_{\scalebox{0.6}{0.84}}$ & 37.55$_{\scalebox{0.6}{1.39}}$ & \textbf{36.88}$_{\scalebox{0.6}{0.78}}$ & \textbf{36.70}$_{\scalebox{0.6}{0.87}}$ & \textbf{38.58}$_{\scalebox{0.6}{1.48}}$ & \cellcolor{tableblue}\textbf{36.68}$_{\scalebox{0.6}{1.83}}$ \\
\cmidrule(lr){1-14}

\multirow{2}{*}{RelationExtraction} 
& KD & \textbf{74.71}$_{\scalebox{0.6}{1.74}}$ & \textbf{75.13}$_{\scalebox{0.6}{0.68}}$ & 74.49$_{\scalebox{0.6}{1.19}}$ & 66.70$_{\scalebox{0.6}{5.61}}$ & 66.65$_{\scalebox{0.6}{3.71}}$ & \cellcolor{tableblue}71.53$_{\scalebox{0.6}{4.44}}$ & \textbf{78.38}$_{\scalebox{0.6}{0.36}}$ & \textbf{78.95}$_{\scalebox{0.6}{0.41}}$ & 78.00$_{\scalebox{0.6}{0.17}}$ & 76.38$_{\scalebox{0.6}{1.13}}$ & \textbf{76.88}$_{\scalebox{0.6}{0.52}}$ & \cellcolor{tableblue}77.72$_{\scalebox{0.6}{1.06}}$ \\
& + \ours & 74.62$_{\scalebox{0.6}{0.58}}$ & 74.34$_{\scalebox{0.6}{1.86}}$ & \textbf{75.32}$_{\scalebox{0.6}{1.82}}$ & \textbf{69.84}$_{\scalebox{0.6}{4.12}}$ & \textbf{69.22}$_{\scalebox{0.6}{2.65}}$ & \cellcolor{tableblue}\textbf{72.67}$_{\scalebox{0.6}{2.90}}$ & 78.29$_{\scalebox{0.6}{0.87}}$ & 78.73$_{\scalebox{0.6}{0.49}}$ & \textbf{79.03}$_{\scalebox{0.6}{0.38}}$ & \textbf{77.06}$_{\scalebox{0.6}{0.89}}$ & 75.90$_{\scalebox{0.6}{1.59}}$ & \cellcolor{tableblue}\textbf{77.80}$_{\scalebox{0.6}{1.30}}$ \\
\midrule

% OVERALL AVERAGE ROWS
\rowcolor{tableblue} 
& KD & 51.36$_{\scalebox{0.6}{20.81}}$ & 50.08$_{\scalebox{0.6}{22.39}}$ & \textbf{47.84}$_{\scalebox{0.6}{24.11}}$ & 42.50$_{\scalebox{0.6}{22.35}}$ & 43.26$_{\scalebox{0.6}{21.70}}$ & \cellcolor{tableblue}47.01$_{\scalebox{0.6}{22.24}}$ & \textbf{52.36}$_{\scalebox{0.6}{23.16}}$ & 52.22$_{\scalebox{0.6}{23.15}}$ & 50.53$_{\scalebox{0.6}{23.79}}$ & 49.34$_{\scalebox{0.6}{23.43}}$ & \textbf{50.02}$_{\scalebox{0.6}{23.26}}$ & \cellcolor{tableblue}50.90$_{\scalebox{0.6}{23.25}}$ \\
\rowcolor{tableblue} 
\textbf{AVG} & + \ours & \textbf{52.10}$_{\scalebox{0.6}{20.17}}$ & \textbf{50.94}$_{\scalebox{0.6}{20.96}}$ & 47.57$_{\scalebox{0.6}{24.62}}$ & \textbf{45.58}$_{\scalebox{0.6}{22.29}}$ & \textbf{44.64}$_{\scalebox{0.6}{22.64}}$ & \cellcolor{tableblue}\textbf{48.17}$_{\scalebox{0.6}{22.11}}$ & 52.00$_{\scalebox{0.6}{23.35}}$ & \textbf{53.53}$_{\scalebox{0.6}{22.08}}$ & \textbf{51.82}$_{\scalebox{0.6}{23.60}}$ & \textbf{50.45}$_{\scalebox{0.6}{23.05}}$ & 49.89$_{\scalebox{0.6}{22.59}}$ & \cellcolor{tableblue}\textbf{51.54}$_{\scalebox{0.6}{22.81}}$ \\
\bottomrule
\end{tabular}%
}
\end{table*}

\subsection{Performance under Modified Contextual Evidence}
\label{app:cfperf}

Tables~\ref{tab:counterfactual_em} and~\ref{tab:counterfactual_f1} report the EM and F1 scores on modified-context test sets, respectively. Tables~\ref{tab:agreement_em} and~\ref{tab:agreement_f1} report the corresponding teacher-student agreement scores using EM and F1, respectively.

\begin{table*}[ht]
\caption{EM score comparison on modified-context test sets across $k$-shot settings for T5 and Qwen2.5 models using KD. Mean EM scores are reported with standard deviations as subscripts ($Mean_{std}$). Bold values indicate the better-performing method (\ours vs. KD). Blue shading highlights aggregate results.}
\label{tab:counterfactual_em}
\centering
\fontsize{6.5pt}{8.5pt}\selectfont
\setlength{\tabcolsep}{1.0pt}
\renewcommand{\arraystretch}{0.95}

\definecolor{tableblue}{HTML}{E6F0FF}

\resizebox{\textwidth}{!}{%
\begin{tabular}{llcccccccccccccc}
\toprule
 & & \multicolumn{6}{c}{\textbf{T5}} & \multicolumn{6}{c}{\textbf{Qwen2.5}} \\
\cmidrule(lr){3-8} \cmidrule(lr){9-14}
Dataset & Method & 16 & 32 & 64 & 128 & 256 & \cellcolor{tableblue}Avg & 16 & 32 & 64 & 128 & 256 & \cellcolor{tableblue}Avg \\
\midrule

\multirow{2}{*}{NQ} 
& KD & 54.90$_{\scalebox{0.6}{4.68}}$ & 58.16$_{\scalebox{0.6}{2.79}}$ & 53.18$_{\scalebox{0.6}{5.88}}$ & 49.00$_{\scalebox{0.6}{6.84}}$ & 51.18$_{\scalebox{0.6}{4.85}}$ & \cellcolor{tableblue}53.28$_{\scalebox{0.6}{3.51}}$ & \textbf{52.10}$_{\scalebox{0.6}{0.64}}$ & 52.98$_{\scalebox{0.6}{0.66}}$ & 52.90$_{\scalebox{0.6}{0.36}}$ & 51.53$_{\scalebox{0.6}{0.57}}$ & 53.04$_{\scalebox{0.6}{0.68}}$ & \cellcolor{tableblue}52.51$_{\scalebox{0.6}{0.67}}$ \\
& + \ours & \textbf{58.70}$_{\scalebox{0.6}{0.78}}$ & \textbf{59.33}$_{\scalebox{0.6}{0.62}}$ & \textbf{56.20}$_{\scalebox{0.6}{1.23}}$ & \textbf{51.83}$_{\scalebox{0.6}{6.47}}$ & \textbf{52.34}$_{\scalebox{0.6}{4.46}}$ & \cellcolor{tableblue}\textbf{55.68}$_{\scalebox{0.6}{3.49}}$ & 51.65$_{\scalebox{0.6}{0.51}}$ & \textbf{53.90}$_{\scalebox{0.6}{1.04}}$ & \textbf{56.10}$_{\scalebox{0.6}{0.50}}$ & \textbf{56.73}$_{\scalebox{0.6}{0.33}}$ & \textbf{56.93}$_{\scalebox{0.6}{1.20}}$ & \cellcolor{tableblue}\textbf{55.06}$_{\scalebox{0.6}{2.26}}$ \\
\cmidrule(lr){1-14}

\multirow{2}{*}{SQuAD} 
& KD & 78.14$_{\scalebox{0.6}{0.84}}$ & 75.71$_{\scalebox{0.6}{0.88}}$ & 74.52$_{\scalebox{0.6}{2.49}}$ & 75.63$_{\scalebox{0.6}{1.17}}$ & 73.43$_{\scalebox{0.6}{5.21}}$ & \cellcolor{tableblue}75.49$_{\scalebox{0.6}{1.75}}$ & \textbf{60.37}$_{\scalebox{0.6}{0.35}}$ & 59.74$_{\scalebox{0.6}{0.66}}$ & 59.18$_{\scalebox{0.6}{0.35}}$ & 60.34$_{\scalebox{0.6}{0.78}}$ & 60.98$_{\scalebox{0.6}{0.41}}$ & \cellcolor{tableblue}60.12$_{\scalebox{0.6}{0.69}}$ \\
& + \ours & \textbf{78.42}$_{\scalebox{0.6}{0.30}}$ & \textbf{78.49}$_{\scalebox{0.6}{0.37}}$ & \textbf{77.05}$_{\scalebox{0.6}{1.20}}$ & \textbf{76.98}$_{\scalebox{0.6}{0.61}}$ & \textbf{75.77}$_{\scalebox{0.6}{1.62}}$ & \cellcolor{tableblue}\textbf{77.34}$_{\scalebox{0.6}{1.14}}$ & 58.48$_{\scalebox{0.6}{0.40}}$ & \textbf{60.80}$_{\scalebox{0.6}{0.38}}$ & \textbf{61.08}$_{\scalebox{0.6}{0.74}}$ & \textbf{62.57}$_{\scalebox{0.6}{1.35}}$ & \textbf{64.77}$_{\scalebox{0.6}{1.63}}$ & \cellcolor{tableblue}\textbf{61.54}$_{\scalebox{0.6}{2.32}}$ \\
\cmidrule(lr){1-14}

\multirow{2}{*}{NewsQA} 
& KD & \textbf{64.29}$_{\scalebox{0.6}{2.04}}$ & 64.58$_{\scalebox{0.6}{3.03}}$ & 63.19$_{\scalebox{0.6}{2.52}}$ & 54.84$_{\scalebox{0.6}{3.97}}$ & 50.49$_{\scalebox{0.6}{0.94}}$ & \cellcolor{tableblue}59.48$_{\scalebox{0.6}{6.43}}$ & 39.54$_{\scalebox{0.6}{1.29}}$ & 34.99$_{\scalebox{0.6}{1.04}}$ & 34.99$_{\scalebox{0.6}{1.38}}$ & 33.42$_{\scalebox{0.6}{1.22}}$ & 32.90$_{\scalebox{0.6}{2.06}}$ & \cellcolor{tableblue}35.17$_{\scalebox{0.6}{2.61}}$ \\
& + \ours & 64.09$_{\scalebox{0.6}{3.66}}$ & \textbf{67.25}$_{\scalebox{0.6}{1.36}}$ & \textbf{67.97}$_{\scalebox{0.6}{1.77}}$ & \textbf{64.20}$_{\scalebox{0.6}{3.27}}$ & \textbf{61.91}$_{\scalebox{0.6}{3.52}}$ & \cellcolor{tableblue}\textbf{65.08}$_{\scalebox{0.6}{2.49}}$ & \textbf{46.09}$_{\scalebox{0.6}{0.98}}$ & \textbf{43.25}$_{\scalebox{0.6}{1.35}}$ & \textbf{39.42}$_{\scalebox{0.6}{1.14}}$ & \textbf{40.12}$_{\scalebox{0.6}{2.49}}$ & \textbf{40.84}$_{\scalebox{0.6}{1.85}}$ & \cellcolor{tableblue}\textbf{41.94}$_{\scalebox{0.6}{2.73}}$ \\
\cmidrule(lr){1-14}

\multirow{2}{*}{TriviaQA} 
& KD & 51.60$_{\scalebox{0.6}{3.07}}$ & 48.43$_{\scalebox{0.6}{2.15}}$ & 44.68$_{\scalebox{0.6}{3.20}}$ & 44.38$_{\scalebox{0.6}{2.50}}$ & 45.50$_{\scalebox{0.6}{1.21}}$ & \cellcolor{tableblue}46.92$_{\scalebox{0.6}{3.07}}$ & \textbf{38.25}$_{\scalebox{0.6}{0.63}}$ & 32.95$_{\scalebox{0.6}{0.74}}$ & 34.33$_{\scalebox{0.6}{1.89}}$ & 33.38$_{\scalebox{0.6}{2.00}}$ & 34.30$_{\scalebox{0.6}{2.85}}$ & \cellcolor{tableblue}34.64$_{\scalebox{0.6}{2.11}}$ \\
& + \ours & \textbf{57.03}$_{\scalebox{0.6}{1.10}}$ & \textbf{54.60}$_{\scalebox{0.6}{0.90}}$ & \textbf{53.71}$_{\scalebox{0.6}{6.19}}$ & \textbf{48.11}$_{\scalebox{0.6}{1.60}}$ & \textbf{48.14}$_{\scalebox{0.6}{2.06}}$ & \cellcolor{tableblue}\textbf{52.32}$_{\scalebox{0.6}{4.02}}$ & 38.06$_{\scalebox{0.6}{2.03}}$ & \textbf{43.46}$_{\scalebox{0.6}{0.73}}$ & \textbf{39.51}$_{\scalebox{0.6}{1.46}}$ & \textbf{45.44}$_{\scalebox{0.6}{1.55}}$ & \textbf{44.71}$_{\scalebox{0.6}{2.20}}$ & \cellcolor{tableblue}\textbf{42.23}$_{\scalebox{0.6}{3.27}}$ \\
\cmidrule(lr){1-14}

\multirow{2}{*}{HotpotQA} 
& KD & 60.63$_{\scalebox{0.6}{4.26}}$ & 62.12$_{\scalebox{0.6}{5.92}}$ & 46.31$_{\scalebox{0.6}{8.29}}$ & 56.83$_{\scalebox{0.6}{4.49}}$ & 58.26$_{\scalebox{0.6}{6.67}}$ & \cellcolor{tableblue}56.83$_{\scalebox{0.6}{6.23}}$ & 43.39$_{\scalebox{0.6}{0.55}}$ & 44.14$_{\scalebox{0.6}{0.88}}$ & 40.87$_{\scalebox{0.6}{1.19}}$ & 45.36$_{\scalebox{0.6}{0.93}}$ & 47.45$_{\scalebox{0.6}{1.66}}$ & \cellcolor{tableblue}44.24$_{\scalebox{0.6}{2.43}}$ \\
& + \ours & \textbf{63.17}$_{\scalebox{0.6}{4.19}}$ & \textbf{62.56}$_{\scalebox{0.6}{5.51}}$ & \textbf{61.81}$_{\scalebox{0.6}{2.05}}$ & \textbf{57.24}$_{\scalebox{0.6}{4.83}}$ & \textbf{64.81}$_{\scalebox{0.6}{3.95}}$ & \cellcolor{tableblue}\textbf{61.92}$_{\scalebox{0.6}{2.84}}$ & \textbf{45.22}$_{\scalebox{0.6}{0.40}}$ & \textbf{45.85}$_{\scalebox{0.6}{0.49}}$ & \textbf{46.25}$_{\scalebox{0.6}{0.62}}$ & \textbf{46.41}$_{\scalebox{0.6}{0.65}}$ & \textbf{54.05}$_{\scalebox{0.6}{1.30}}$ & \cellcolor{tableblue}\textbf{47.56}$_{\scalebox{0.6}{3.66}}$ \\
\midrule

% OVERALL AVERAGE ROWS
\rowcolor{tableblue} 
& KD & 61.91$_{\scalebox{0.6}{10.32}}$ & 61.80$_{\scalebox{0.6}{9.92}}$ & 56.37$_{\scalebox{0.6}{12.49}}$ & 56.14$_{\scalebox{0.6}{11.95}}$ & 55.77$_{\scalebox{0.6}{10.87}}$ & \cellcolor{tableblue}58.40$_{\scalebox{0.6}{10.65}}$ & 46.73$_{\scalebox{0.6}{9.35}}$ & 44.96$_{\scalebox{0.6}{11.48}}$ & 44.45$_{\scalebox{0.6}{11.10}}$ & 44.80$_{\scalebox{0.6}{11.69}}$ & 45.73$_{\scalebox{0.6}{12.09}}$ & \cellcolor{tableblue}45.34$_{\scalebox{0.6}{11.06}}$ \\
\rowcolor{tableblue} 
\multirow{-2}{*}{\textbf{AVG}} & + \ours & \textbf{64.28}$_{\scalebox{0.6}{8.44}}$ & \textbf{64.44}$_{\scalebox{0.6}{9.11}}$ & \textbf{63.35}$_{\scalebox{0.6}{9.43}}$ & \textbf{59.67}$_{\scalebox{0.6}{11.42}}$ & \textbf{60.59}$_{\scalebox{0.6}{10.87}}$ & \cellcolor{tableblue}\textbf{62.47}$_{\scalebox{0.6}{9.71}}$ & \textbf{47.90}$_{\scalebox{0.6}{7.64}}$ & \textbf{49.45}$_{\scalebox{0.6}{7.68}}$ & \textbf{48.47}$_{\scalebox{0.6}{9.80}}$ & \textbf{50.25}$_{\scalebox{0.6}{9.14}}$ & \textbf{52.26}$_{\scalebox{0.6}{9.60}}$ & \cellcolor{tableblue}\textbf{49.67}$_{\scalebox{0.6}{8.51}}$ \\
\bottomrule
\end{tabular}%
}
\end{table*}

\begin{table*}[htbp]
\caption{F1 score comparison on modified-context test sets across $k$-shot settings for T5 and Qwen2.5 models using KD. Mean F1 scores are reported with standard deviations as subscripts ($Mean_{std}$). Bold values indicate the better-performing method (\ours vs. KD). Blue shading highlights aggregate results.}
\label{tab:counterfactual_f1}
\centering
\fontsize{6.5pt}{8.5pt}\selectfont
\setlength{\tabcolsep}{1.0pt}
\renewcommand{\arraystretch}{0.95}

\definecolor{tableblue}{HTML}{E6F0FF}

\resizebox{\textwidth}{!}{%
\begin{tabular}{llcccccccccccccc}
\toprule
 & & \multicolumn{6}{c}{\textbf{T5}} & \multicolumn{6}{c}{\textbf{Qwen2.5}} \\
\cmidrule(lr){3-8} \cmidrule(lr){9-14}
Dataset & Method & 16 & 32 & 64 & 128 & 256 & \cellcolor{tableblue}Avg & 16 & 32 & 64 & 128 & 256 & \cellcolor{tableblue}Avg \\
\midrule

\multirow{2}{*}{NQ} 
& KD & 60.03$_{\scalebox{0.6}{3.26}}$ & 61.70$_{\scalebox{0.6}{1.69}}$ & 59.73$_{\scalebox{0.6}{2.45}}$ & 57.84$_{\scalebox{0.6}{2.95}}$ & 58.49$_{\scalebox{0.6}{2.03}}$ & \cellcolor{tableblue}59.56$_{\scalebox{0.6}{1.49}}$ & \textbf{54.52}$_{\scalebox{0.6}{0.43}}$ & 55.36$_{\scalebox{0.6}{0.65}}$ & 55.40$_{\scalebox{0.6}{0.54}}$ & 54.07$_{\scalebox{0.6}{0.58}}$ & 55.29$_{\scalebox{0.6}{0.50}}$ & \cellcolor{tableblue}54.93$_{\scalebox{0.6}{0.60}}$ \\
& + \ours & \textbf{62.05}$_{\scalebox{0.6}{0.53}}$ & \textbf{62.65}$_{\scalebox{0.6}{0.65}}$ & \textbf{60.04}$_{\scalebox{0.6}{0.70}}$ & \textbf{59.79}$_{\scalebox{0.6}{2.52}}$ & \textbf{59.30}$_{\scalebox{0.6}{1.90}}$ & \cellcolor{tableblue}\textbf{60.76}$_{\scalebox{0.6}{1.49}}$ & 53.74$_{\scalebox{0.6}{0.50}}$ & \textbf{56.26}$_{\scalebox{0.6}{0.96}}$ & \textbf{58.43}$_{\scalebox{0.6}{0.45}}$ & \textbf{59.07}$_{\scalebox{0.6}{0.30}}$ & \textbf{59.51}$_{\scalebox{0.6}{1.25}}$ & \cellcolor{tableblue}\textbf{57.40}$_{\scalebox{0.6}{2.40}}$ \\
\cmidrule(lr){1-14}

\multirow{2}{*}{SQuAD} 
& KD & 86.28$_{\scalebox{0.6}{0.78}}$ & 84.15$_{\scalebox{0.6}{0.74}}$ & 83.72$_{\scalebox{0.6}{1.19}}$ & 84.55$_{\scalebox{0.6}{1.04}}$ & 82.90$_{\scalebox{0.6}{2.76}}$ & \cellcolor{tableblue}84.32$_{\scalebox{0.6}{1.26}}$ & \textbf{65.26}$_{\scalebox{0.6}{0.33}}$ & 64.49$_{\scalebox{0.6}{0.59}}$ & 64.19$_{\scalebox{0.6}{0.38}}$ & 65.67$_{\scalebox{0.6}{0.74}}$ & 66.22$_{\scalebox{0.6}{0.60}}$ & \cellcolor{tableblue}65.17$_{\scalebox{0.6}{0.83}}$ \\
& + \ours & \textbf{86.52}$_{\scalebox{0.6}{0.19}}$ & \textbf{86.51}$_{\scalebox{0.6}{0.35}}$ & \textbf{85.28}$_{\scalebox{0.6}{0.98}}$ & \textbf{85.34}$_{\scalebox{0.6}{0.62}}$ & \textbf{84.31}$_{\scalebox{0.6}{1.31}}$ & \cellcolor{tableblue}\textbf{85.59}$_{\scalebox{0.6}{0.94}}$ & 63.82$_{\scalebox{0.6}{0.44}}$ & \textbf{65.57}$_{\scalebox{0.6}{0.28}}$ & \textbf{66.12}$_{\scalebox{0.6}{0.75}}$ & \textbf{67.46}$_{\scalebox{0.6}{1.36}}$ & \textbf{69.79}$_{\scalebox{0.6}{1.74}}$ & \cellcolor{tableblue}\textbf{66.55}$_{\scalebox{0.6}{2.23}}$ \\
\cmidrule(lr){1-14}

\multirow{2}{*}{NewsQA} 
& KD & 71.99$_{\scalebox{0.6}{2.01}}$ & 73.56$_{\scalebox{0.6}{1.97}}$ & 72.39$_{\scalebox{0.6}{1.33}}$ & 69.03$_{\scalebox{0.6}{3.06}}$ & 66.17$_{\scalebox{0.6}{0.96}}$ & \cellcolor{tableblue}70.63$_{\scalebox{0.6}{3.00}}$ & 46.19$_{\scalebox{0.6}{1.28}}$ & 42.10$_{\scalebox{0.6}{1.00}}$ & 42.91$_{\scalebox{0.6}{1.56}}$ & 39.78$_{\scalebox{0.6}{1.20}}$ & 39.19$_{\scalebox{0.6}{2.34}}$ & \cellcolor{tableblue}42.03$_{\scalebox{0.6}{2.80}}$ \\
& + \ours & \textbf{72.81}$_{\scalebox{0.6}{3.00}}$ & \textbf{74.10}$_{\scalebox{0.6}{1.22}}$ & \textbf{75.66}$_{\scalebox{0.6}{1.73}}$ & \textbf{73.92}$_{\scalebox{0.6}{1.69}}$ & \textbf{74.23}$_{\scalebox{0.6}{1.75}}$ & \cellcolor{tableblue}\textbf{74.15}$_{\scalebox{0.6}{1.02}}$ & \textbf{52.16}$_{\scalebox{0.6}{0.74}}$ & \textbf{49.46}$_{\scalebox{0.6}{1.12}}$ & \textbf{46.00}$_{\scalebox{0.6}{1.24}}$ & \textbf{47.37}$_{\scalebox{0.6}{1.99}}$ & \textbf{47.49}$_{\scalebox{0.6}{1.25}}$ & \cellcolor{tableblue}\textbf{48.50}$_{\scalebox{0.6}{2.39}}$ \\
\cmidrule(lr){1-14}

\multirow{2}{*}{TriviaQA} 
& KD & 54.79$_{\scalebox{0.6}{3.30}}$ & 51.31$_{\scalebox{0.6}{2.29}}$ & 46.77$_{\scalebox{0.6}{3.37}}$ & 46.18$_{\scalebox{0.6}{2.44}}$ & 47.35$_{\scalebox{0.6}{1.26}}$ & \cellcolor{tableblue}49.28$_{\scalebox{0.6}{3.68}}$ & \textbf{40.22}$_{\scalebox{0.6}{0.69}}$ & 34.68$_{\scalebox{0.6}{0.70}}$ & 35.75$_{\scalebox{0.6}{1.90}}$ & 34.98$_{\scalebox{0.6}{2.06}}$ & 36.23$_{\scalebox{0.6}{3.00}}$ & \cellcolor{tableblue}36.37$_{\scalebox{0.6}{2.24}}$ \\
& + \ours & \textbf{60.72}$_{\scalebox{0.6}{1.02}}$ & \textbf{57.89}$_{\scalebox{0.6}{0.89}}$ & \textbf{57.05}$_{\scalebox{0.6}{6.86}}$ & \textbf{50.47}$_{\scalebox{0.6}{1.55}}$ & \textbf{50.22}$_{\scalebox{0.6}{2.09}}$ & \cellcolor{tableblue}\textbf{55.27}$_{\scalebox{0.6}{4.70}}$ & 39.77$_{\scalebox{0.6}{2.14}}$ & \textbf{45.27}$_{\scalebox{0.6}{0.70}}$ & \textbf{41.43}$_{\scalebox{0.6}{1.40}}$ & \textbf{46.84}$_{\scalebox{0.6}{1.40}}$ & \textbf{46.57}$_{\scalebox{0.6}{2.29}}$ & \cellcolor{tableblue}\textbf{43.98}$_{\scalebox{0.6}{3.19}}$ \\
\cmidrule(lr){1-14}

\multirow{2}{*}{HotpotQA} 
& KD & \textbf{71.87}$_{\scalebox{0.6}{2.37}}$ & 71.75$_{\scalebox{0.6}{3.96}}$ & 66.22$_{\scalebox{0.6}{3.20}}$ & \textbf{70.06}$_{\scalebox{0.6}{1.28}}$ & 69.10$_{\scalebox{0.6}{3.78}}$ & \cellcolor{tableblue}69.80$_{\scalebox{0.6}{2.32}}$ & 47.08$_{\scalebox{0.6}{0.52}}$ & 47.94$_{\scalebox{0.6}{0.70}}$ & 44.26$_{\scalebox{0.6}{1.25}}$ & 48.55$_{\scalebox{0.6}{1.05}}$ & 50.53$_{\scalebox{0.6}{1.57}}$ & \cellcolor{tableblue}47.67$_{\scalebox{0.6}{2.29}}$ \\
& + \ours & 71.73$_{\scalebox{0.6}{1.89}}$ & \textbf{72.79}$_{\scalebox{0.6}{2.70}}$ & \textbf{72.24}$_{\scalebox{0.6}{0.81}}$ & 69.88$_{\scalebox{0.6}{2.26}}$ & \textbf{73.22}$_{\scalebox{0.6}{2.78}}$ & \cellcolor{tableblue}\textbf{71.97}$_{\scalebox{0.6}{1.30}}$ & \textbf{49.56}$_{\scalebox{0.6}{0.43}}$ & \textbf{49.78}$_{\scalebox{0.6}{0.31}}$ & \textbf{49.92}$_{\scalebox{0.6}{0.57}}$ & \textbf{50.01}$_{\scalebox{0.6}{0.54}}$ & \textbf{57.38}$_{\scalebox{0.6}{1.43}}$ & \cellcolor{tableblue}\textbf{51.33}$_{\scalebox{0.6}{3.39}}$ \\
\midrule

% OVERALL AVERAGE ROWS
\rowcolor{tableblue} 
& KD & 68.99$_{\scalebox{0.6}{12.23}}$ & 68.49$_{\scalebox{0.6}{12.48}}$ & 65.76$_{\scalebox{0.6}{13.82}}$ & 65.53$_{\scalebox{0.6}{14.39}}$ & 64.80$_{\scalebox{0.6}{13.15}}$ & \cellcolor{tableblue}66.72$_{\scalebox{0.6}{13.13}}$ & 50.66$_{\scalebox{0.6}{9.62}}$ & 48.92$_{\scalebox{0.6}{11.56}}$ & 48.50$_{\scalebox{0.6}{11.24}}$ & 48.61$_{\scalebox{0.6}{12.09}}$ & 49.49$_{\scalebox{0.6}{12.21}}$ & \cellcolor{tableblue}49.23$_{\scalebox{0.6}{11.25}}$ \\
\rowcolor{tableblue} 
\multirow{-2}{*}{\textbf{AVG}} & + \ours & \textbf{70.76}$_{\scalebox{0.6}{10.37}}$ & \textbf{70.79}$_{\scalebox{0.6}{11.12}}$ & \textbf{70.06}$_{\scalebox{0.6}{11.59}}$ & \textbf{67.88}$_{\scalebox{0.6}{13.36}}$ & \textbf{68.26}$_{\scalebox{0.6}{13.45}}$ & \cellcolor{tableblue}\textbf{69.55}$_{\scalebox{0.6}{11.89}}$ & \textbf{51.81}$_{\scalebox{0.6}{8.63}}$ & \textbf{53.27}$_{\scalebox{0.6}{7.92}}$ & \textbf{52.38}$_{\scalebox{0.6}{9.90}}$ & \textbf{54.15}$_{\scalebox{0.6}{8.91}}$ & \textbf{56.15}$_{\scalebox{0.6}{9.56}}$ & \cellcolor{tableblue}\textbf{53.55}$_{\scalebox{0.6}{8.75}}$ \\
\bottomrule
\end{tabular}%
}
\end{table*}

\begin{table*}[t]
\caption{Teacher-student agreement EM score comparison on modified-context test sets across $k$-shot settings for T5 and Qwen2.5 models using KD. Mean agreement EM scores are reported with standard deviations as subscripts ($Mean_{std}$). Bold values indicate the better-performing method (\ours vs. KD). Blue shading highlights aggregate results.}
\label{tab:agreement_em}
\centering
\fontsize{6.5pt}{8.5pt}\selectfont
\setlength{\tabcolsep}{1.0pt}
\renewcommand{\arraystretch}{0.95}

\definecolor{tableblue}{HTML}{E6F0FF}

\resizebox{\textwidth}{!}{%
\begin{tabular}{llcccccccccccccc}
\toprule
 & & \multicolumn{6}{c}{\textbf{T5}} & \multicolumn{6}{c}{\textbf{Qwen2.5}} \\
\cmidrule(lr){3-8} \cmidrule(lr){9-14}
Dataset & Method & 16 & 32 & 64 & 128 & 256 & \cellcolor{tableblue}Avg & 16 & 32 & 64 & 128 & 256 & \cellcolor{tableblue}Avg \\
\midrule

\multirow{2}{*}{NQ} 
& KD & 56.23$_{\scalebox{0.6}{4.87}}$ & 60.20$_{\scalebox{0.6}{2.87}}$ & 54.47$_{\scalebox{0.6}{6.56}}$ & 50.26$_{\scalebox{0.6}{7.34}}$ & 52.62$_{\scalebox{0.6}{5.00}}$ & \cellcolor{tableblue}54.76$_{\scalebox{0.6}{3.76}}$ & \textbf{53.69}$_{\scalebox{0.6}{0.61}}$ & 55.75$_{\scalebox{0.6}{0.87}}$ & 55.44$_{\scalebox{0.6}{0.53}}$ & 54.31$_{\scalebox{0.6}{0.82}}$ & 55.25$_{\scalebox{0.6}{0.27}}$ & \cellcolor{tableblue}54.89$_{\scalebox{0.6}{0.86}}$ \\
& + \ours & \textbf{61.34}$_{\scalebox{0.6}{0.56}}$ & \textbf{60.64}$_{\scalebox{0.6}{0.49}}$ & \textbf{57.97}$_{\scalebox{0.6}{1.36}}$ & \textbf{52.94}$_{\scalebox{0.6}{6.72}}$ & \textbf{53.46}$_{\scalebox{0.6}{4.92}}$ & \cellcolor{tableblue}\textbf{57.27}$_{\scalebox{0.6}{3.93}}$ & 52.62$_{\scalebox{0.6}{0.58}}$ & \textbf{56.01}$_{\scalebox{0.6}{1.03}}$ & \textbf{57.34}$_{\scalebox{0.6}{0.62}}$ & \textbf{58.57}$_{\scalebox{0.6}{0.68}}$ & \textbf{58.50}$_{\scalebox{0.6}{0.62}}$ & \cellcolor{tableblue}\textbf{56.61}$_{\scalebox{0.6}{2.46}}$ \\
\cmidrule(lr){1-14}

\multirow{2}{*}{SQuAD} 
& KD & 85.35$_{\scalebox{0.6}{0.75}}$ & 81.94$_{\scalebox{0.6}{1.37}}$ & 80.52$_{\scalebox{0.6}{3.17}}$ & 82.17$_{\scalebox{0.6}{2.32}}$ & 78.67$_{\scalebox{0.6}{6.21}}$ & \cellcolor{tableblue}81.73$_{\scalebox{0.6}{2.46}}$ & \textbf{61.86}$_{\scalebox{0.6}{0.51}}$ & 61.42$_{\scalebox{0.6}{0.73}}$ & 61.10$_{\scalebox{0.6}{0.59}}$ & 61.37$_{\scalebox{0.6}{1.08}}$ & 61.69$_{\scalebox{0.6}{0.77}}$ & \cellcolor{tableblue}61.49$_{\scalebox{0.6}{0.29}}$ \\
& + \ours & \textbf{85.73}$_{\scalebox{0.6}{0.39}}$ & \textbf{85.55}$_{\scalebox{0.6}{0.58}}$ & \textbf{83.33}$_{\scalebox{0.6}{1.33}}$ & \textbf{83.96}$_{\scalebox{0.6}{0.45}}$ & \textbf{81.28}$_{\scalebox{0.6}{2.43}}$ & \cellcolor{tableblue}\textbf{83.97}$_{\scalebox{0.6}{1.82}}$ & 60.45$_{\scalebox{0.6}{0.43}}$ & \textbf{62.22}$_{\scalebox{0.6}{0.32}}$ & \textbf{62.22}$_{\scalebox{0.6}{1.01}}$ & \textbf{63.67}$_{\scalebox{0.6}{0.99}}$ & \textbf{64.48}$_{\scalebox{0.6}{0.86}}$ & \cellcolor{tableblue}\textbf{62.61}$_{\scalebox{0.6}{1.55}}$ \\
\cmidrule(lr){1-14}

\multirow{2}{*}{NewsQA} 
& KD & 63.28$_{\scalebox{0.6}{2.18}}$ & 63.10$_{\scalebox{0.6}{3.40}}$ & 60.61$_{\scalebox{0.6}{2.80}}$ & 51.19$_{\scalebox{0.6}{4.13}}$ & 46.58$_{\scalebox{0.6}{1.56}}$ & \cellcolor{tableblue}56.95$_{\scalebox{0.6}{7.62}}$ & 41.80$_{\scalebox{0.6}{1.25}}$ & 37.86$_{\scalebox{0.6}{0.82}}$ & 37.51$_{\scalebox{0.6}{0.50}}$ & 36.87$_{\scalebox{0.6}{1.29}}$ & 35.97$_{\scalebox{0.6}{1.75}}$ & \cellcolor{tableblue}38.00$_{\scalebox{0.6}{2.24}}$ \\
& + \ours & \textbf{63.83}$_{\scalebox{0.6}{2.50}}$ & \textbf{66.09}$_{\scalebox{0.6}{0.66}}$ & \textbf{67.13}$_{\scalebox{0.6}{1.61}}$ & \textbf{61.71}$_{\scalebox{0.6}{4.24}}$ & \textbf{58.00}$_{\scalebox{0.6}{3.91}}$ & \cellcolor{tableblue}\textbf{63.35}$_{\scalebox{0.6}{3.65}}$ & \textbf{46.12}$_{\scalebox{0.6}{1.78}}$ & \textbf{43.59}$_{\scalebox{0.6}{0.93}}$ & \textbf{40.84}$_{\scalebox{0.6}{0.80}}$ & \textbf{41.91}$_{\scalebox{0.6}{1.62}}$ & \textbf{43.30}$_{\scalebox{0.6}{1.75}}$ & \cellcolor{tableblue}\textbf{43.15}$_{\scalebox{0.6}{1.99}}$ \\
\cmidrule(lr){1-14}

\multirow{2}{*}{TriviaQA} 
& KD & 56.44$_{\scalebox{0.6}{3.22}}$ & 53.71$_{\scalebox{0.6}{2.02}}$ & 49.56$_{\scalebox{0.6}{3.44}}$ & 49.46$_{\scalebox{0.6}{2.68}}$ & 50.35$_{\scalebox{0.6}{1.12}}$ & \cellcolor{tableblue}51.90$_{\scalebox{0.6}{3.07}}$ & \textbf{39.04}$_{\scalebox{0.6}{1.14}}$ & \textbf{40.13}$_{\scalebox{0.6}{1.66}}$ & 38.25$_{\scalebox{0.6}{0.98}}$ & \textbf{38.16}$_{\scalebox{0.6}{0.98}}$ & \textbf{38.06}$_{\scalebox{0.6}{1.68}}$ & \cellcolor{tableblue}\textbf{38.73}$_{\scalebox{0.6}{0.88}}$ \\
& + \ours & \textbf{62.24}$_{\scalebox{0.6}{0.99}}$ & \textbf{59.90}$_{\scalebox{0.6}{0.68}}$ & \textbf{59.31}$_{\scalebox{0.6}{6.57}}$ & \textbf{52.62}$_{\scalebox{0.6}{2.03}}$ & \textbf{53.15}$_{\scalebox{0.6}{2.57}}$ & \cellcolor{tableblue}\textbf{57.44}$_{\scalebox{0.6}{4.31}}$ & 37.46$_{\scalebox{0.6}{1.34}}$ & 37.69$_{\scalebox{0.6}{0.63}}$ & \textbf{39.67}$_{\scalebox{0.6}{1.35}}$ & 35.42$_{\scalebox{0.6}{0.26}}$ & 34.53$_{\scalebox{0.6}{1.08}}$ & \cellcolor{tableblue}36.96$_{\scalebox{0.6}{2.03}}$ \\
\cmidrule(lr){1-14}

\multirow{2}{*}{HotpotQA} 
& KD & 56.97$_{\scalebox{0.6}{4.68}}$ & 58.65$_{\scalebox{0.6}{7.33}}$ & 42.35$_{\scalebox{0.6}{7.99}}$ & 53.01$_{\scalebox{0.6}{4.67}}$ & 54.89$_{\scalebox{0.6}{6.91}}$ & \cellcolor{tableblue}53.17$_{\scalebox{0.6}{6.41}}$ & 50.38$_{\scalebox{0.6}{0.70}}$ & 50.92$_{\scalebox{0.6}{0.64}}$ & 46.89$_{\scalebox{0.6}{1.09}}$ & 50.65$_{\scalebox{0.6}{1.52}}$ & 51.49$_{\scalebox{0.6}{1.81}}$ & \cellcolor{tableblue}50.07$_{\scalebox{0.6}{1.82}}$ \\
& + \ours & \textbf{60.36}$_{\scalebox{0.6}{4.46}}$ & \textbf{58.95}$_{\scalebox{0.6}{5.76}}$ & \textbf{57.54}$_{\scalebox{0.6}{1.95}}$ & \textbf{53.81}$_{\scalebox{0.6}{4.75}}$ & \textbf{61.99}$_{\scalebox{0.6}{3.86}}$ & \cellcolor{tableblue}\textbf{58.53}$_{\scalebox{0.6}{3.11}}$ & \textbf{51.86}$_{\scalebox{0.6}{0.89}}$ & \textbf{52.23}$_{\scalebox{0.6}{0.58}}$ & \textbf{52.81}$_{\scalebox{0.6}{0.67}}$ & \textbf{52.18}$_{\scalebox{0.6}{0.93}}$ & \textbf{56.61}$_{\scalebox{0.6}{1.45}}$ & \cellcolor{tableblue}\textbf{53.14}$_{\scalebox{0.6}{1.97}}$ \\
\midrule

% OVERALL AVERAGE ROWS
\rowcolor{tableblue} 
& KD & 63.65$_{\scalebox{0.6}{12.48}}$ & 63.52$_{\scalebox{0.6}{10.85}}$ & 57.50$_{\scalebox{0.6}{14.50}}$ & 57.22$_{\scalebox{0.6}{14.01}}$ & 56.62$_{\scalebox{0.6}{12.70}}$ & \cellcolor{tableblue}59.70$_{\scalebox{0.6}{12.46}}$ & 49.35$_{\scalebox{0.6}{9.21}}$ & 49.21$_{\scalebox{0.6}{10.08}}$ & 47.84$_{\scalebox{0.6}{10.41}}$ & 48.27$_{\scalebox{0.6}{10.56}}$ & 48.49$_{\scalebox{0.6}{11.12}}$ & \cellcolor{tableblue}48.63$_{\scalebox{0.6}{10.22}}$ \\
\rowcolor{tableblue} 
\multirow{-2}{*}{\textbf{AVG}} & + \ours & \textbf{66.70}$_{\scalebox{0.6}{10.72}}$ & \textbf{66.22}$_{\scalebox{0.6}{11.15}}$ & \textbf{65.06}$_{\scalebox{0.6}{10.93}}$ & \textbf{61.01}$_{\scalebox{0.6}{13.36}}$ & \textbf{61.57}$_{\scalebox{0.6}{11.60}}$ & \cellcolor{tableblue}\textbf{64.11}$_{\scalebox{0.6}{11.37}}$ & \textbf{49.70}$_{\scalebox{0.6}{8.54}}$ & \textbf{50.35}$_{\scalebox{0.6}{9.78}}$ & \textbf{50.57}$_{\scalebox{0.6}{10.00}}$ & \textbf{50.35}$_{\scalebox{0.6}{11.65}}$ & \textbf{51.48}$_{\scalebox{0.6}{12.23}}$ & \cellcolor{tableblue}\textbf{50.49}$_{\scalebox{0.6}{10.35}}$ \\
\bottomrule
\end{tabular}%
}
\end{table*}

\begin{table*}[t]
\caption{Teacher-student agreement F1 score comparison on modified-context test sets across $k$-shot settings for T5 and Qwen2.5 models using KD. Mean agreement F1 scores are reported with standard deviations as subscripts ($Mean_{std}$). Bold values indicate the better-performing method (\ours vs. KD). Blue shading highlights aggregate results.}
\label{tab:agreement_f1}
\centering
\fontsize{6.5pt}{8.5pt}\selectfont
\setlength{\tabcolsep}{1.0pt}
\renewcommand{\arraystretch}{0.95}

\definecolor{tableblue}{HTML}{E6F0FF}

\resizebox{\textwidth}{!}{%
\begin{tabular}{llcccccccccccccc}
\toprule
 & & \multicolumn{6}{c}{\textbf{T5}} & \multicolumn{6}{c}{\textbf{Qwen2.5}} \\
\cmidrule(lr){3-8} \cmidrule(lr){9-14}
Dataset & Method & 16 & 32 & 64 & 128 & 256 & \cellcolor{tableblue}Avg & 16 & 32 & 64 & 128 & 256 & \cellcolor{tableblue}Avg \\
\midrule

\multirow{2}{*}{NQ} 
& KD & 62.31$_{\scalebox{0.6}{3.15}}$ & 64.50$_{\scalebox{0.6}{1.70}}$ & 61.75$_{\scalebox{0.6}{2.79}}$ & 60.11$_{\scalebox{0.6}{3.06}}$ & 60.98$_{\scalebox{0.6}{2.11}}$ & \cellcolor{tableblue}61.93$_{\scalebox{0.6}{1.66}}$ & \textbf{59.30}$_{\scalebox{0.6}{0.46}}$ & 61.21$_{\scalebox{0.6}{0.76}}$ & 60.87$_{\scalebox{0.6}{0.35}}$ & 59.53$_{\scalebox{0.6}{0.70}}$ & 60.45$_{\scalebox{0.6}{0.37}}$ & \cellcolor{tableblue}60.27$_{\scalebox{0.6}{0.83}}$ \\
& + \ours & \textbf{65.08}$_{\scalebox{0.6}{0.40}}$ & \textbf{64.91}$_{\scalebox{0.6}{0.42}}$ & \textbf{62.65}$_{\scalebox{0.6}{0.70}}$ & \textbf{61.96}$_{\scalebox{0.6}{2.63}}$ & \textbf{61.36}$_{\scalebox{0.6}{2.37}}$ & \cellcolor{tableblue}\textbf{63.19}$_{\scalebox{0.6}{1.71}}$ & 58.04$_{\scalebox{0.6}{0.48}}$ & \textbf{61.55}$_{\scalebox{0.6}{1.04}}$ & \textbf{62.83}$_{\scalebox{0.6}{0.51}}$ & \textbf{63.71}$_{\scalebox{0.6}{0.58}}$ & \textbf{64.03}$_{\scalebox{0.6}{0.59}}$ & \cellcolor{tableblue}\textbf{62.03}$_{\scalebox{0.6}{2.43}}$ \\
\cmidrule(lr){1-14}

\multirow{2}{*}{SQuAD} 
& KD & 90.29$_{\scalebox{0.6}{0.71}}$ & 87.73$_{\scalebox{0.6}{0.93}}$ & 87.20$_{\scalebox{0.6}{1.46}}$ & 88.17$_{\scalebox{0.6}{1.55}}$ & 85.90$_{\scalebox{0.6}{3.23}}$ & \cellcolor{tableblue}87.86$_{\scalebox{0.6}{1.60}}$ & \textbf{67.69}$_{\scalebox{0.6}{0.47}}$ & 67.24$_{\scalebox{0.6}{0.39}}$ & 66.98$_{\scalebox{0.6}{0.45}}$ & 68.04$_{\scalebox{0.6}{0.86}}$ & 68.34$_{\scalebox{0.6}{0.57}}$ & \cellcolor{tableblue}67.66$_{\scalebox{0.6}{0.56}}$ \\
& + \ours & \textbf{90.45}$_{\scalebox{0.6}{0.25}}$ & \textbf{90.43}$_{\scalebox{0.6}{0.40}}$ & \textbf{88.88}$_{\scalebox{0.6}{0.85}}$ & \textbf{89.13}$_{\scalebox{0.6}{0.39}}$ & \textbf{87.34}$_{\scalebox{0.6}{1.48}}$ & \cellcolor{tableblue}\textbf{89.25}$_{\scalebox{0.6}{1.29}}$ & 66.49$_{\scalebox{0.6}{0.57}}$ & \textbf{67.98}$_{\scalebox{0.6}{0.16}}$ & \textbf{68.37}$_{\scalebox{0.6}{0.80}}$ & \textbf{69.49}$_{\scalebox{0.6}{0.87}}$ & \textbf{71.00}$_{\scalebox{0.6}{1.13}}$ & \cellcolor{tableblue}\textbf{68.66}$_{\scalebox{0.6}{1.69}}$ \\
\cmidrule(lr){1-14}

\multirow{2}{*}{NewsQA} 
& KD & 72.06$_{\scalebox{0.6}{1.88}}$ & 72.84$_{\scalebox{0.6}{2.12}}$ & 71.26$_{\scalebox{0.6}{1.39}}$ & 67.68$_{\scalebox{0.6}{2.77}}$ & 64.94$_{\scalebox{0.6}{1.45}}$ & \cellcolor{tableblue}69.76$_{\scalebox{0.6}{3.34}}$ & 50.55$_{\scalebox{0.6}{1.41}}$ & 47.13$_{\scalebox{0.6}{1.02}}$ & 47.65$_{\scalebox{0.6}{1.17}}$ & 45.17$_{\scalebox{0.6}{1.52}}$ & 44.41$_{\scalebox{0.6}{2.23}}$ & \cellcolor{tableblue}46.98$_{\scalebox{0.6}{2.40}}$ \\
& + \ours & \textbf{73.00}$_{\scalebox{0.6}{2.11}}$ & \textbf{73.97}$_{\scalebox{0.6}{0.76}}$ & \textbf{75.48}$_{\scalebox{0.6}{1.49}}$ & \textbf{73.08}$_{\scalebox{0.6}{2.41}}$ & \textbf{72.80}$_{\scalebox{0.6}{1.84}}$ & \cellcolor{tableblue}\textbf{73.67}$_{\scalebox{0.6}{1.11}}$ & \textbf{54.76}$_{\scalebox{0.6}{1.54}}$ & \textbf{52.54}$_{\scalebox{0.6}{1.06}}$ & \textbf{50.42}$_{\scalebox{0.6}{1.16}}$ & \textbf{51.80}$_{\scalebox{0.6}{1.19}}$ & \textbf{51.95}$_{\scalebox{0.6}{1.57}}$ & \cellcolor{tableblue}\textbf{52.29}$_{\scalebox{0.6}{1.58}}$ \\
\cmidrule(lr){1-14}

\multirow{2}{*}{TriviaQA} 
& KD & 59.24$_{\scalebox{0.6}{3.43}}$ & 55.96$_{\scalebox{0.6}{2.19}}$ & 51.30$_{\scalebox{0.6}{3.58}}$ & 51.04$_{\scalebox{0.6}{2.70}}$ & 51.79$_{\scalebox{0.6}{1.05}}$ & \cellcolor{tableblue}53.87$_{\scalebox{0.6}{3.61}}$ & \textbf{41.48}$_{\scalebox{0.6}{1.15}}$ & \textbf{42.62}$_{\scalebox{0.6}{1.27}}$ & 40.53$_{\scalebox{0.6}{0.74}}$ & \textbf{40.39}$_{\scalebox{0.6}{1.21}}$ & \textbf{40.48}$_{\scalebox{0.6}{1.58}}$ & \cellcolor{tableblue}\textbf{41.10}$_{\scalebox{0.6}{0.96}}$ \\
& + \ours & \textbf{65.44}$_{\scalebox{0.6}{0.86}}$ & \textbf{62.68}$_{\scalebox{0.6}{0.75}}$ & \textbf{61.91}$_{\scalebox{0.6}{6.87}}$ & \textbf{54.53}$_{\scalebox{0.6}{1.97}}$ & \textbf{54.98}$_{\scalebox{0.6}{2.30}}$ & \cellcolor{tableblue}\textbf{59.91}$_{\scalebox{0.6}{4.89}}$ & 39.75$_{\scalebox{0.6}{1.33}}$ & 40.07$_{\scalebox{0.6}{0.55}}$ & \textbf{42.01}$_{\scalebox{0.6}{1.32}}$ & 37.77$_{\scalebox{0.6}{0.19}}$ & 37.38$_{\scalebox{0.6}{1.09}}$ & \cellcolor{tableblue}39.40$_{\scalebox{0.6}{1.88}}$ \\
\cmidrule(lr){1-14}

\multirow{2}{*}{HotpotQA} 
& KD & 71.01$_{\scalebox{0.6}{2.73}}$ & 71.11$_{\scalebox{0.6}{4.62}}$ & 65.09$_{\scalebox{0.6}{3.52}}$ & \textbf{69.61}$_{\scalebox{0.6}{1.64}}$ & 68.77$_{\scalebox{0.6}{3.88}}$ & \cellcolor{tableblue}69.12$_{\scalebox{0.6}{2.46}}$ & 56.90$_{\scalebox{0.6}{0.71}}$ & 57.62$_{\scalebox{0.6}{0.58}}$ & 53.65$_{\scalebox{0.6}{1.08}}$ & 57.20$_{\scalebox{0.6}{1.38}}$ & 58.01$_{\scalebox{0.6}{1.66}}$ & \cellcolor{tableblue}56.68$_{\scalebox{0.6}{1.74}}$ \\
& + \ours & \textbf{71.54}$_{\scalebox{0.6}{2.28}}$ & \textbf{72.14}$_{\scalebox{0.6}{2.83}}$ & \textbf{71.25}$_{\scalebox{0.6}{0.72}}$ & 69.38$_{\scalebox{0.6}{2.28}}$ & \textbf{73.26}$_{\scalebox{0.6}{2.65}}$ & \cellcolor{tableblue}\textbf{71.51}$_{\scalebox{0.6}{1.42}}$ & \textbf{58.74}$_{\scalebox{0.6}{0.70}}$ & \textbf{59.12}$_{\scalebox{0.6}{0.62}}$ & \textbf{59.70}$_{\scalebox{0.6}{0.64}}$ & \textbf{58.93}$_{\scalebox{0.6}{0.87}}$ & \textbf{63.02}$_{\scalebox{0.6}{1.12}}$ & \cellcolor{tableblue}\textbf{59.90}$_{\scalebox{0.6}{1.78}}$ \\
\midrule

% OVERALL AVERAGE ROWS
\rowcolor{tableblue} 
& KD & 70.98$_{\scalebox{0.6}{12.11}}$ & 70.43$_{\scalebox{0.6}{11.73}}$ & 67.32$_{\scalebox{0.6}{13.26}}$ & 67.32$_{\scalebox{0.6}{13.76}}$ & 66.48$_{\scalebox{0.6}{12.56}}$ & \cellcolor{tableblue}68.51$_{\scalebox{0.6}{12.59}}$ & 55.18$_{\scalebox{0.6}{9.82}}$ & 55.16$_{\scalebox{0.6}{10.13}}$ & 53.94$_{\scalebox{0.6}{10.46}}$ & 54.07$_{\scalebox{0.6}{11.19}}$ & 54.34$_{\scalebox{0.6}{11.59}}$ & \cellcolor{tableblue}54.54$_{\scalebox{0.6}{10.58}}$ \\
\rowcolor{tableblue} 
\multirow{-2}{*}{\textbf{AVG}} & + \ours & \textbf{73.10}$_{\scalebox{0.6}{10.33}}$ & \textbf{72.83}$_{\scalebox{0.6}{10.92}}$ & \textbf{72.03}$_{\scalebox{0.6}{11.03}}$ & \textbf{69.62}$_{\scalebox{0.6}{13.03}}$ & \textbf{69.95}$_{\scalebox{0.6}{12.44}}$ & \cellcolor{tableblue}\textbf{71.50}$_{\scalebox{0.6}{11.44}}$ & \textbf{55.56}$_{\scalebox{0.6}{9.83}}$ & \textbf{56.25}$_{\scalebox{0.6}{10.60}}$ & \textbf{56.67}$_{\scalebox{0.6}{10.46}}$ & \textbf{56.34}$_{\scalebox{0.6}{12.24}}$ & \textbf{57.47}$_{\scalebox{0.6}{13.14}}$ & \cellcolor{tableblue}\textbf{56.46}$_{\scalebox{0.6}{11.18}}$ \\
\bottomrule
\end{tabular}%
}
\end{table*}

\end{document}